\documentclass[11pt]{article}

\usepackage[final]{acl}

\usepackage{times}
\usepackage{latexsym}
\usepackage{soul}
\usepackage[T1]{fontenc}
\usepackage[utf8]{inputenc}
\usepackage{multicol}
\usepackage{microtype}

\usepackage{inconsolata}

\usepackage{graphicx}

\usepackage{mdframed}
\usepackage{microtype}
\usepackage{graphicx}
\usepackage{fontawesome5}
\usepackage{subcaption}
\usepackage{booktabs} 
\usepackage{xcolor}
\usepackage{pifont}
\usepackage{multirow}
\usepackage{array}
\usepackage{hyperref}              
\usepackage{url}                   
\usepackage{booktabs}              
\usepackage{boxedminipage}

\newlength{\kvqaboxsep}
\newlength{\kvqaboxrule}
\usepackage{amsfonts, amssymb, amsmath} 
\usepackage{nicefrac}              
\usepackage{microtype}             
\usepackage{xcolor}                
\usepackage{colortbl}              
\usepackage{multirow}              
\usepackage{algorithm}
\usepackage{algpseudocode}
\usepackage[export]{adjustbox}     
\usepackage{threeparttable}        
\usepackage{enumitem}              
\usepackage{newfloat}              
\usepackage{listings}              
\usepackage{caption}               
\usepackage{wrapfig}               
\usepackage{pifont}                
\usepackage{float}                 
\usepackage{stmaryrd}
\usepackage{bm}                    
\usepackage{placeins}              
\usepackage{graphicx}              
\definecolor{cadetblue}{rgb}{0.372, 0.620, 0.627}
\usepackage{array}
\usepackage{booktabs}
\usepackage{multirow}
\usepackage{minitoc}

\definecolor{mygray}{gray}{.9}
\definecolor{mygreen}{RGB}{93,173,85}
\definecolor{mywarning}{RGB}{233,144,61}

\definecolor{InvBlue}{RGB}{57,84,148}  
\definecolor{FwdGreen}{RGB}{96,129,63}
\definecolor{MutedPurple}{HTML}{703D8A}
\definecolor{MutedRust}{HTML}{8A543D}
\definecolor{DarkBlue}{RGB}{64,101,149}
\definecolor{azure}{rgb}{0.0, 0.5, 1.0}
\definecolor{gray}{rgb}{0.3, 0.3, 0.3}
\definecolor{DarkGreen}{RGB}{42,110,63}
\definecolor{Green}{rgb}{0.0, 0.5, 0.0}       
\definecolor{RedOrange}{rgb}{1.0, 0.27, 0.0}  
\definecolor{sage1}{RGB}{217,229,243}
\definecolor{sage2}{RGB}{229,237,248}
\definecolor{sage3}{RGB}{240,245,252}
\newcommand{\blue}[1]{$_{\color{Green}\downarrow #1}$}
\newcommand{\myred}[1]{$_{\color{RedOrange}\uparrow #1}$}

\newcolumntype{x}[1]{>{\centering\arraybackslash}p{#1pt}}
\newcolumntype{I}{!{\vrule width 1pt}} 

\definecolor{darkgreen}{RGB}{0,100,0}
\newcommand{\diff}[1]{{\scriptsize\textbf{\textcolor{darkgreen}{#1}}}}
\usepackage[most]{tcolorbox}
\tcbuselibrary{theorems}
\definecolor{lightgray}{gray}{.9}
\definecolor{deepgray}{gray}{.8}
\definecolor{HdrA}{HTML}{D6E2F0} 
\definecolor{HdrB}{HTML}{E3ECF7} 
\definecolor{HdrC}{HTML}{EFF4FB} 
\definecolor{HdrD}{HTML}{E9EEF5} 
\definecolor{HdrE}{HTML}{F4F7FB} 
\newcommand{\vp}[1]{\textcolor{teal!70!black}{#1}}      
\newcommand{\vr}[1]{\textcolor{orange!85!black}{#1}}    
\newcommand{\tgr}[1]{\textcolor{purple!80!black}{#1}}   
\tcbset{highlight math/.append style={left=0mm,right=0mm,top=0mm,bottom=0mm, colframe=white}}
\definecolor{keywordcolor}{RGB}{178,34,34} 
\newtcbox{\predwrong}{
  on line,
  colback=red!10,
  colframe=red!60!black,
  boxrule=0.4pt,
  arc=1mm,
  left=1pt,
  right=1pt,
  top=1pt,
  bottom=1pt
}

\newtcbox{\predcorrect}{
  on line,
  colback=green!10,
  colframe=green!60!black,
  boxrule=0.4pt,
  arc=1mm,
  left=1pt,
  right=1pt,
  top=1pt,
  bottom=1pt
}
\definecolor{TagBlue}{HTML}{1F4E79}
\newtcbox{\searchtag}{%
  on line,
  box align=base,
  colframe=TagBlue,
  colback=TagBlue!2!white, 
  coltext=TagBlue,  
  boxrule=0.5pt,
  arc=1.2pt,
  boxsep=0.3pt,
  left=1.0pt,right=1.0pt,top=0.2pt,bottom=0.2pt,
  fontupper=\ttfamily\small
}
\definecolor{bestblue}{rgb}{0.908, 0.961, 0.908}
\definecolor{bestyellow}{rgb}{0.960, 0.866, 0.721}
\newcommand{\ab}[1]{\colorbox{bestblue}{\strut #1}}
\newcommand{\vb}[1]{\colorbox{bestyellow}{\strut #1}}

\definecolor{TagTeal}{HTML}{1B6B6F}
\definecolor{TagPurple}{HTML}{5B3FA6}
\definecolor{TagRose}{HTML}{8B2E5A}

\newtcbox{\tealtag}{%
  on line,
  box align=base,
  colframe=TagTeal,
  colback=TagTeal!2!white,
  coltext=TagTeal,
  boxrule=0.5pt,
  arc=1.2pt,
  boxsep=0.3pt,
  left=1.0pt,right=1.0pt,top=0.2pt,bottom=0.2pt,
  fontupper=\ttfamily\small
}

\newtcbox{\purpletag}{%
  on line,
  box align=base,
  colframe=TagPurple,
  colback=TagPurple!2!white,
  coltext=TagPurple,
  boxrule=0.5pt,
  arc=1.2pt,
  boxsep=0.3pt,
  left=1.0pt,right=1.0pt,top=0.2pt,bottom=0.2pt,
  fontupper=\ttfamily\small
}

\newtcbox{\rosetag}{%
  on line,
  box align=base,
  colframe=TagRose,
  colback=TagRose!2!white,
  coltext=TagRose,
  boxrule=0.5pt,
  arc=1.2pt,
  boxsep=0.3pt,
  left=1.0pt,right=1.0pt,top=0.2pt,bottom=0.2pt,
  fontupper=\ttfamily\small 
}

\definecolor{TagOrange}{HTML}{D97706}

\newtcbox{\orangetag}{%
  on line,
  breakable,
  tcbox width=auto limited, 
  box align=base,
  colframe=TagOrange,
  colback=TagOrange!2!white,
  coltext=TagOrange,
  boxrule=0.5pt,
  arc=1.2pt,
  boxsep=0.3pt,
  left=1.0pt,right=1.0pt,top=0.2pt,bottom=0.2pt,
  fontupper=\ttfamily\small
}

\newtcolorbox{casebox}[1]{
  enhanced,
  colback=black!2,
  colframe=black!60,
  boxrule=0.6pt,
  arc=2mm,
  outer arc=2mm,
  top=10mm,
  bottom=4mm,
  left=8mm,
  right=8mm,
  fontupper=\ttfamily\small,
  width=\textwidth,
  title={#1},
  overlay unbroken={
    \path[fill=black!75, draw=black!75]
      (frame.north west) rectangle ([yshift=-18pt]frame.north east);
    \node[anchor=west, font=\bfseries\ttfamily\large, text=white]
      at ([xshift=10pt, yshift=-9pt]frame.north west)
      {\tcbtitle};
  }
}

\newtcolorbox{caseboxpurple}[1]{%
  enhanced,
  colback=violet!5,          
  colframe=violet!50!black,  
  boxrule=0.6pt,
  arc=2mm,
  outer arc=2mm,
  top=12mm,
  bottom=4mm,
  left=8mm,
  right=8mm,
  fontupper=\ttfamily\small, 
  width=\textwidth,
  title={#1},
  overlay unbroken={
    \path[fill=violet!70!black, draw=violet!70!black]
      (frame.north west) rectangle ([yshift=-18pt]frame.north east);
    \node[anchor=west, font=\bfseries\ttfamily\large, text=white]
      at ([xshift=10pt, yshift=-9pt]frame.north west)
      {\tcbtitle};
  }
}

\makeatletter
\newcommand{\thickhline}{%
    \noalign {\ifnum 0=`}\fi \hrule height 1pt
    \futurelet \reserved@a \@xhline
}
\makeatother



\usepackage[capitalize]{cleveref}
\crefname{proposition}{Prop.}{Props.}
\crefname{section}{Sec.}{Secs.}
\crefname{table}{Tab.}{Tabs.}

\usepackage{xspace}
\makeatletter
\DeclareRobustCommand\onedot{\futurelet\@let@token\@onedot}
\def\@onedot{\ifx\@let@token.\else.\null\fi\xspace}

\definecolor{CadetBlue}{RGB}{95,158,160} 
\makeatother

\usepackage{amsmath}

\usepackage{amssymb}
\usepackage{mathtools}
\usepackage{amsthm}

\theoremstyle{plain}

\theoremstyle{definition}

\theoremstyle{remark}

\crefname{definition}{Definition}{Definitions}
\Crefname{definition}{Definition}{Definitions}

\title{\texorpdfstring{%
  \raisebox{-0.2\height}{\includegraphics[height=1.2em]{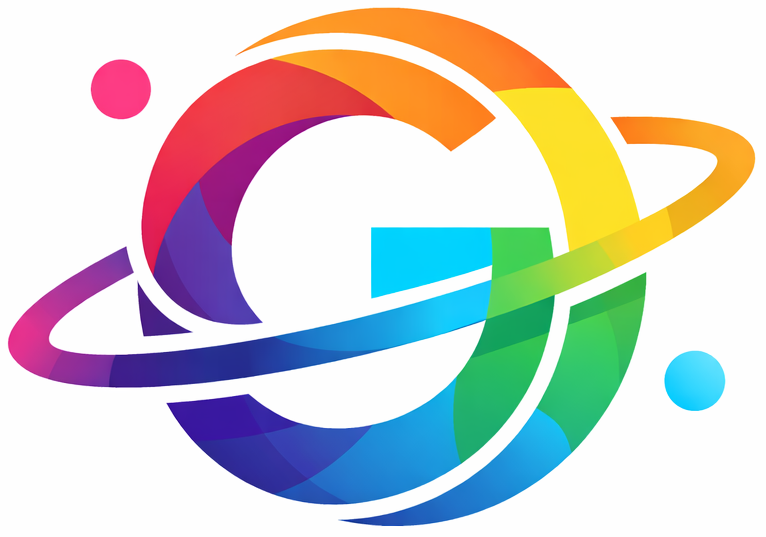}}%
  \hspace{-0.2em} 
  GraphVerse: A Comprehensive Visual Graph Reasoning Benchmark \\for Multimodal Large Language Models
}{GraphVerse: A Visual Graph Understanding and Reasoning Benchmark for Multimodal Large Language Models}}

\author{
  \textbf{Yuanfu Sun}$^{1,2,3}$\footnotemark[1],
  \textbf{Yuanhang Ren}$^{2}$\footnotemark[1],
  \textbf{Kang Li}$^{4}$\footnotemark[1],
  \\
  \textbf{Chuanhao Ji}$^{1}$,
  \textbf{Jiaxi Li}$^{5}$,
  \textbf{Jiajin Liu}$^{3}$,
  \textbf{Ninghao Liu}$^{6}$,
  \textbf{Qiaoyu Tan}$^{1}$\footnotemark[2]
  \\[0.3em]
  $^{1}$New York University Shanghai
  $^{2}$Sensetime Research
  $^{3}$New York University
  \\
  $^{4}$Tsinghua University
  $^{5}$University of Georgia
  \\
  $^{6}$The Hong Kong Polytechnic University
  \\
  \texttt{\{yuanfu.sun, qiaoyu.tan\}@nyu.edu, lik24@mails.tsinghua.edu.cn}\\
}

\begin{document}
\doparttoc
\faketableofcontents

\maketitle
\renewcommand{\thefootnote}{\fnsymbol{footnote}}
\footnotetext[1]{Equal Contribution. Yuanfu Sun led the project.} \footnotetext[2]{Corresponding author.}
\begin{abstract}

Recent Multimodal Large Language Models (MLLMs) have achieved remarkable progress across diverse vision-language tasks, creating an urgent need for more challenging benchmarks. Yet existing evaluations still provide limited insight into whether these models can truly reason over structured visual information. Visual Graph Reasoning (VGR) offers a compelling testbed for this challenge, requiring models to integrate perception, structural understanding, and multi-step reasoning over graph-based visual inputs. However, prior VGR benchmarks often reduce the task to visual perception followed by text-based reasoning, restrict evaluation to single-image settings, rely on answer-only metrics, and underrepresent realistic graph-centric scenarios. To bridge the gap, we introduce \textbf{GraphVerse}, a unified benchmark that jointly evaluates perception, visual reasoning, and text-based graph reasoning in MLLMs under both single-image and paired-image settings. At its core is a suite of Graph-centric Image Editing (GIE) strategies that modify graph images while preserving their semantics, turning them into active tests of visual reasoning. We further propose VGR-Score, a process-sensitive metric that evaluates reasoning quality beyond final-answer accuracy. Extensive experiments reveal several key limitations of current MLLMs in VGR, while also validating the effectiveness of GIE strategies and the transferability of GraphVerse to broader multimodal reasoning capabilities. The code is available at \url{https://github.com/sunyuanfu/GraphVerse}.

\end{abstract}

\section{Introduction}

The recent advancement of Multimodal Large Language Models (MLLMs) \cite{team2025kimi,bai2025qwen2} has been accompanied by a surge of benchmarks designed to evaluate multimodal understanding and reasoning at scale, from broad expert-domain exams \cite{yue2024mmmu} to capability-focused stress tests \cite{yu2023mm}. Among these, multimodal mathematical reasoning has become a key paradigm, requiring both visual grounding and multi-step inference, while often enabling automatic answer verification \cite{zhang2024mathverse}. However, this paradigm still faces several limitations: high-quality data curation is costly, many benchmarks emphasize basic perception or language-based reasoning rather than genuine visual cognition \cite{xu2025visulogic}, and recurring formats increasingly blur genuine reasoning with learned priors \cite{TangZLCL25}.

To address these limitations, we seek an evaluation substrate that is scalable, easily verifiable, resistant to shortcut, and capable of rewarding reasoning that remains faithful to the underlying structure.
\textbf{Graphs provide a natural foundation}: they serve as a universal representation of relational systems, such as E-commerce \cite{shchur2018pitfalls}, molecules \cite{hu2020open} and social interactions \cite{hamilton2017inductive}.
Importantly, these systems often contain rich visual signals rather than symbolic edge lists.
This motivates the task of visual graph understanding and reasoning: an MLLM must perceive nodes, edges, and attributes from the diagram, internalize the latent topology, and solve graph-based problems via structure-consistent inference \cite{zhu2025benchmarking}.

However, existing benchmarks for visual graph reasoning (VGR) exhibit several key limitations.
\textbf{First, they reduce visual graph reasoning to visual perception plus text-based graph reasoning}: models transcribe diagrams into symbolic graphs, and then solve the task in the text domain \cite{wei2024gita,li2024visiongraph,zhu2025benchmarking,babaiee2025visual}.
This ``vision-to-text'' design weakens the very challenge that visual graph reasoning is meant to capture---it allows the problem to collapse into text-based graph QA, and it fails to stress whether a model can truly reason in the vision domain to perform multi-step, relational/logical inference grounded in different or noisy visual evidence.
\textbf{Second, they focus only on single-image reasoning} and overlook a more realistic setting in which evidence is distributed across multiple images, requiring models to perform cross-image reasoning.
\textbf{Third, evaluation is often answer-only}: these benchmarks typically judge a single final output and ignore intermediate reasoning behaviors, making failures non-diagnostic, and making successes hard to interpret.
\textbf{Finally, questions are often posed as context-free algorithmic templates rather than situated problems}, even when real-world graphs are used \cite{zhu2025benchmarking}, weakening ecological validity and reducing our ability to study how MLLMs reason about graphs as they appear in practice.
These limitations motivate the following research question:
\vspace{-0.6cm}
\begin{tcolorbox}[
    colback=blue!5,      
    colframe=black,      
    arc=8pt,             
    boxrule=0.8pt,       
    left=8pt,right=8pt,  
    top=6pt,bottom=6pt   
]
\textit{\textcolor{red}{\textbf{(RQ)}} Can we build a solver-verifiable benchmark that evaluates visual graph understanding and reasoning end-to-end, moving from visual perception to \textbf{both} vision-based and text-based reasoning, while providing diagnostic signals beyond final accuracy and spanning difficulty from tractable to combinatorial-hard problems in realistic scenarios?}
\end{tcolorbox}
\vspace{-0.2cm}

We take a step toward this goal with GraphVerse, a benchmark for comprehensively evaluating MLLMs’ robust and in-depth visual graph understanding and reasoning. Grounded in real-world graph-centric scenarios, GraphVerse spans diverse multimodal graph problems across a wide range of complexity. We organize these tasks by visual context cardinality into single-image VGR, which reasons over one rendered graph, and paired-image VGR, which requires cross-image structural alignment. To place greater emphasis on vision-based reasoning, we further introduce a suite of graph-centric image editing strategies that push models beyond one-shot perception toward genuine reasoning. In addition, we propose VGR-Score, an evaluation protocol that goes beyond final accuracy and provides more diagnostic and fine-grained assessment of graph understanding and reasoning. Extensive experiments systematically reveal the limitations and failure modes of current MLLMs on VGR, offer practical insights for improving their capabilities, validate the effectiveness of our editing strategies, and show that reasoning skills learned on GraphVerse can transfer beyond the benchmark to improve performance on other domains. Our key contributions can be summarized as follows.

\begin{itemize}
    \item We introduce GraphVerse, the first benchmark to jointly evaluate perception, visual reasoning, and text-based graph reasoning in MLLMs under a unified visual graph reasoning setting, together with VGR-Score, a task-specific process-sensitive metric that measures reasoning quality beyond final answers.
    \item We propose graph-centric image editing strategies, a new structure-aware editing framework that transforms graph images into active tests of visual reasoning for MLLMs. Our strategy is broadly applicable across diverse, realistic visual graph reasoning tasks.
    \item We systematically evaluate several state-of-the-art MLLMs on GraphVerse across diverse tasks and reasoning patterns, establishing strong baselines and highlighting substantial opportunities for future research.
\end{itemize}

\section{Related Work}

\noindent \textbf{Visual Reasoning.}
Recent works increasingly view visual reasoning as structured inference over visually grounded entities and relations, rather than isolated cue recognition \cite{yue2025mmmu,xu2025visulogic,yuan2025mme,yue2024mmmu}. 
A common formulation separates it into perception and reasoning: perception extracts task-relevant entities, attributes, and relations from images, while reasoning composes them into multi-step inference such as comparison, relational chaining, and rule-based deduction to answer a query~\cite{johnson2017clevr,hudson2019gqa,zhang2019raven,huang2021seeing,mao2022clevrer,li2025mits}. Crucially, both intermediate conclusions and final answers must be grounded in visual reasoning rather than textual paraphrases or language shortcuts~\cite{xu2025visulogic, liu2025mitigating}.

\begin{figure*}[t]
    \centering
    \includegraphics[width=1\linewidth]{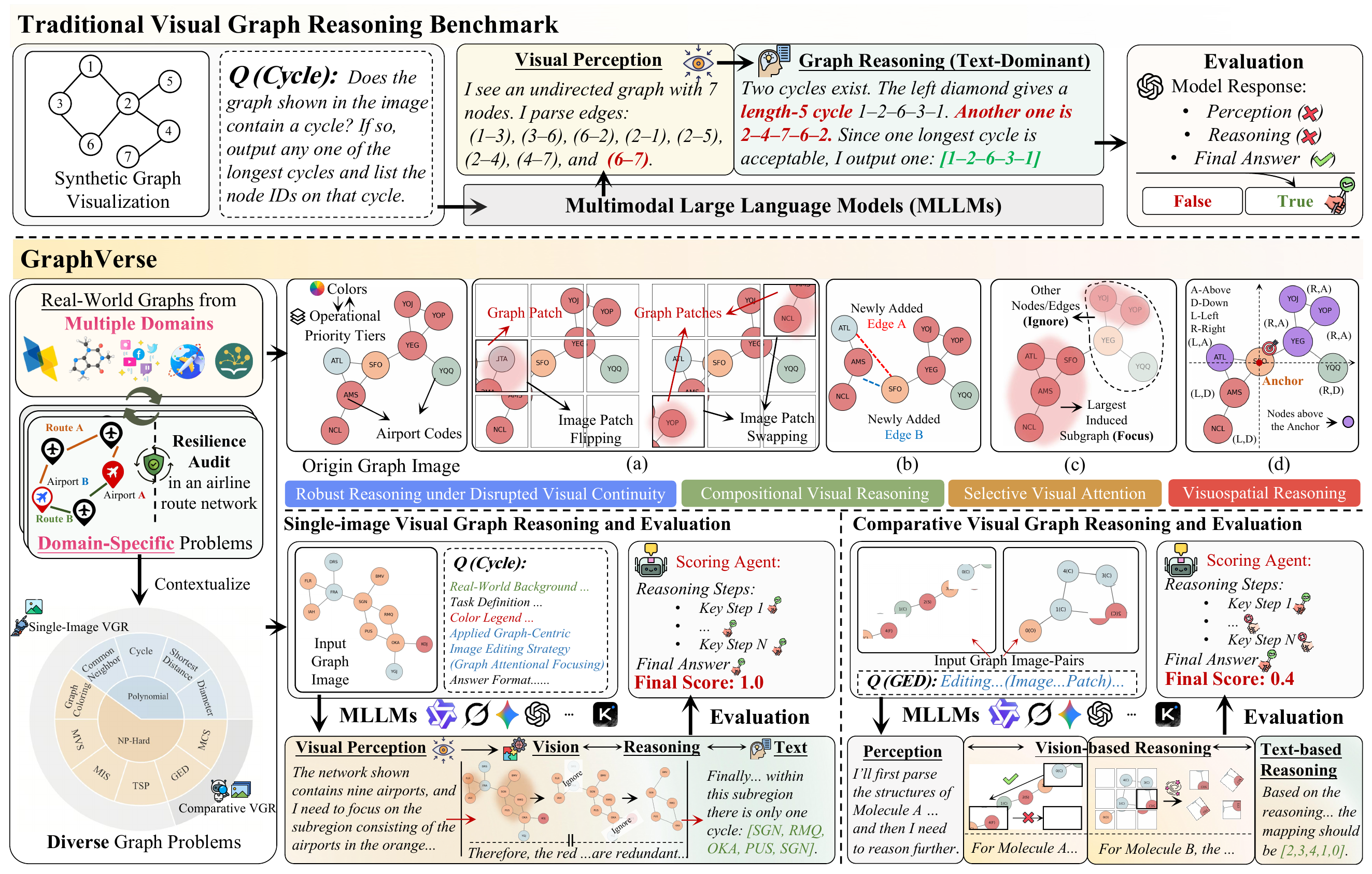}
    \vspace{-0.6cm}
    \caption{Design overview of GraphVerse. 
    All tasks are grounded in real-world scenarios and span a broad range of complexity (bottom left).
    Unlike traditional VGR benchmarks (top), which collapse visual graph reasoning into perception followed by text-based inference, GraphVerse promotes vision-based reasoning through four graph-centric image editing strategies: (a) Image-Graph Patch Perturbation, (b) Cross-Graph Composition, (c) Graph Attentional Focusing, and (d) Spatially-Conditioned Recoloring. These strategies are applied across two settings, single-image and paired-image (bottom right). A scoring agent evaluates both reasoning process and final answer.
    }
    \label{fig:placeholder}
    \vspace{-0.4cm}
\end{figure*}

\noindent \textbf{Benchmarking Algorithmic Reasoning on Graphs.}
Beyond traditional graph learning tasks \cite{sun2025graphicl,sun2026agentgl,sun2026mario, yang2026one,zhang2025trustglm}, graph computational problems have increasingly emerged as an important research focus.
Recent work has positioned graph computational problems as a crucial testbed for LLM and MLLM reasoning, posing greater challenges than standard graph learning and conventional reasoning problems \cite{chen2025medfact,fan2026mlaga,liu2026graphvlm} due to their demands for structural understanding and multi-step, long-range inference. Early text-only benchmarks evaluate graph problem solving from symbolic graph descriptions \cite{wang2023can,fatemi2023talk,wu2025grapheval36k}. More recent efforts \cite{li2024visiongraph,wei2024gita,zhu2025benchmarking} extend this line to visual graphs. However, these benchmarks still fall short in evaluating broad and robust visual graph reasoning. Some work focuses on layout-induced perception challenges and answer-only evaluation, overlooking richer visually grounded reasoning processes ~\cite{babaiee2025visual}.

\section{Preliminaries}

\textbf{Notations.}
Real-world structured data can often be modeled as graphs. We denote a graph by
$G = (V, E, \mathcal{A}_V, \mathcal{A}_E)$, where $V$ is the node set, $E \subseteq V \times V$ is the edge set, and $\mathcal{A}_V$ / $\mathcal{A}_E$ are optional node/edge attributes. Here, we only consider textual attributes.
The adjacency matrix $A$ encodes connectivity with $A_{ij}=1$ iff $(v_i,v_j)\in E$. We provide the problem definition as follows:

\noindent \textbf{Problem 3.1} (\emph{Visual Graph Reasoning with MLLMs}). 
Let $I=\mathrm{Render}(G)$ be a graph image that depicts a graph $G=(V,E,\mathcal{A}_V,\mathcal{A}_E)$, including its connectivity and (optional) node/edge attributes. 
Given $I$ and a text-specified graph reasoning task $T$ grounded in $I$, an MLLM $f_\theta$ produces a reasoning trace $R$ and a final answer $\hat{y}$. 
Evaluation uses metrics computed from $R$ and $\hat{y}$. 
The ground-truth $y$ is produced by a pre-defined programmatic  solver $\mathcal{S}$ executed on $G$, i.e., $y=\mathcal{S}(G)$.

\section{GraphVerse Benchmark}
In this section, we present GraphVerse, a benchmark for comprehensively evaluating visual graph reasoning in MLLMs. We first describe the data curation process (Sec.~\ref{sec:data_collection}), then present the task taxonomy (Sec.~\ref{sec:task_definition}) and four proposed Graph-Centric Image Editing strategies (Sec.~\ref{sec:image editing}), and finally outline the evaluation protocol (Sec.~\ref{sec:evaluation}). For additional design rationale, please refer to Sec.~\ref{sec:design_ration}.

\subsection{Graph Data Collection and Visualization}
\label{sec:data_collection}
\textbf{Principle of Real-World Grounding.}
To ensure that GraphVerse reflects real-world problem settings, we curate each datapoint according to three principles:
\textcolor{MutedPurple}{(a) the underlying graph structure is derived from naturally occurring graphs}; \textcolor{MutedRust}{(b) the visualization is information-complete with respect to the graph}, i.e., graph-relevant cues are conveyed primarily through the image rather than being disclosed in the accompanying text and \textcolor{DarkBlue}{(c) each question is contextualized within a realistic scenario}.

\noindent \textbf{Graph Sources \textcolor{MutedPurple}{(Principle a)}.}
Following prior work \cite{TangZLCL25}, we construct GraphVerse from diverse real-world graph sources, including citation networks (DBLP \cite{Ley02}), social networks (Digg, Flixster, Lastfm, Pokec \cite{RossiA15}), knowledge graphs (DBpedia1M \cite{BizerLKABCH09}), transportation graphs (OpenFlights \cite{TangZLCL25}), and molecular graphs (PubChemQC \cite{DBLP:journals/jcisd/NakataS17}, PCQM4Mv2 \cite{DBLP:conf/nips/HuFRNDL21}). Table~\ref{tab:graph_task_desc} details the semantics of nodes and edges for each graph type. Except for the thousand-scale OpenFlights graph, all sources are million-scale, enabling sufficient per-instance sampling. We further use random walk with restart \cite{TangZLCL25} to extract compact, well-connected local subgraphs. 
 
\noindent \textbf{Graph Visualization \textcolor{MutedRust}{(Principle b)}.}
We visualize each sampled subgraph using GraphViz \cite{DBLP:conf/gd/EllsonGKNW00}. To encourage evidence-seeking from the image, we disentangle modalities between evidence and query: all graph-relevant information required to solve the task (e.g., structure, node/edge attributes, when available) is rendered directly in the visualization, while the question itself is provided only in the text prompt. To reduce shortcuts, we follow a consistent and well-suited visualization protocol (layout, styling, and legends) and avoid duplicating any graph evidence in the prompt.

\noindent \textbf{Question Design \textcolor{DarkBlue}{(Principle c)}.}
Given each sampled subgraph, we assign a suitable graph-theoretic task and apply various graph-centric image editing strategies (Section~\ref{sec:image editing}) to its visualization. In addition, we curate a large set of graph problems grounded in the corresponding real-world scenarios and make minimal adaptations (e.g., adding constraints and standardizing answer formats) to fit the task or image editing strategies and to facilitate automatic answer extraction. We solve each instantiated problem with rule-based algorithms to obtain verifiable ground-truth. The detailed prompts can be found in Appendix~\ref{sec:g} (hereafter §~\ref{sec:g}).

\subsection{Task Definition}
\label{sec:task_definition}

We instantiate GraphVerse with classic graph theory problems because they offer well-defined, verifiable solutions and naturally elicit relational, multi-hop reasoning that transfers across domains.
We organize tasks according to the number of input graph images: (1) \textbf{Single-image VGR}, where the model solves a graph problem from a single graph image, and (2) \textbf{Paired-image comparative VGR}, where the model jointly analyzes two graph images to infer their correspondences and/or differences.
We place greater emphasis on the single-image setting, as most existing MLLM evaluations assume a single-visual-context input.
Overall, our tasks span polynomial-time \textbf{\tealtag{\texttt{Poly}}} and NP-hard \textbf{\rosetag{\texttt{NPH}}} problems. For NP-hard tasks, we use optimization variants that require models to produce optimal solutions rather than binary yes/no answers, enabling a more thorough evaluation of MLLM reasoning.

\noindent \paragraph{Single-image VGR Tasks.} Given a graph image $I$, an MLLM must first obtain the underlying graph structure from $I$ through visual understanding and reasoning and then solve one of the tasks:

\noindent \textbf{\tealtag{\texttt{Poly}} Shortest Distance (SD).} Given $u,v\in V$, compute the shortest-path distance $d_G(u,v)$, and output one shortest path if required.

\noindent \textbf{\tealtag{\texttt{Poly}} Diameter.} Compute the maximum shortest-path distance over all node pairs in the graph $\mathrm{diam}(G)=\max_{u,v\in V} d_G(u,v)$, $\forall u,v\in V$.

\noindent \textbf{\tealtag{\texttt{Poly}} Common Neighbors (CN).} Given $u,v\in V$, return the common neighbor set $\Gamma(u)\cap\Gamma(v)$, explicitly listing all shared neighbors.

\noindent \textbf{\rosetag{\texttt{NPH}} Traveling Salesman (TSP).} Given a weighted graph with edge weights $w$, find a Hamiltonian tour of minimum total cost that visits each node exactly once and returns to the starting node.

\noindent \textbf{\rosetag{\texttt{NPH}} Maximum Clique Problem (MCP).} Find a maximum-size clique $C\subseteq V$ such that every pair of distinct nodes in $C$ is connected by an edge, i.e., $\forall u,v\in C,\ u\neq v \Rightarrow (u,v)\in E$.

\noindent \textbf{\rosetag{\texttt{NPH}} Minimum Vertex Cover (MVC):} find a minimum-size set $S\subseteq V$ covering all edges,
i.e., $\forall (u,v)\in E$, $u\in S$ or $v\in S$.
  
\noindent \textbf{\rosetag{\texttt{NPH}} Maximum Independent Set (MIS).} Find a maximum-size set $S\subseteq V$ with no internal edges, i.e., $\forall u\neq v\in S$, $(u,v)\notin E$.

\noindent \textbf{\rosetag{\texttt{NPH}} Cycle.} Decide whether $G$ contains a cycle; if so, return any longest cycle $C'$. Otherwise, report that no cycle exists in $G$.

\noindent \textbf{\rosetag{\texttt{NPH}} Graph Coloring.} Find a proper vertex coloring $c:V\rightarrow [k]$ that minimizes $k$, where adjacent vertices receive different colors.

\noindent \paragraph{Paired-image Comparative VGR Tasks.} 
Given a paired input $(I_1,I_2)$ that renders two graphs $G_1$ and $G_2$, an MLLM is required to establish cross-image correspondences and perform pairwise structural reasoning to solve one of the following tasks:

\noindent \textbf{\rosetag{\texttt{NPH}} Graph Edit Distance (GED).} Compute the minimum number (or cost) of edit operations required to transform $G_1$ into $G_2$, including node and edge insertions, deletions, and substitutions.

\noindent \textbf{\rosetag{\texttt{NPH}} Maximum Common Subgraph (MCS).} Find a largest subgraph $H$ that is isomorphic to subgraphs of both $G_1$ and $G_2$, and output the corresponding node mapping.

\subsection{Graph-Centric Image Editing}
\label{sec:image editing}
\noindent We introduce four Graph-centric Image Editing (GIE) strategies, each targeting a core facet of visual reasoning. By altering the visual realization of the underlying graph structure, these strategies discourage shortcut exploitation and push models to perform vision-based reasoning rather than rely on shallow perception.
See §~\ref{sec:d} for detailed GIE algorithms in both single- and paired-image settings.

\noindent \textbf{Image-Graph Patch Perturbation.}
This edit manipulates image patches to disrupt the visual realization of graph patches (local edges/nodes), thereby probing robust reasoning under disrupted visual continuity, staying faithful to the underlying structure despite local corruption \cite{shen2025assessing}.
Specifically, we partition an input image $I$ into an $r \times r$ grid of patches $\{P_{ij}\}$, and apply one of two perturbations: (i) a random patch flip $P'_{ij}\leftarrow \mathrm{Flip}(P_{ij})$, or (ii) a random patch swap $(P_{ij},P_{kl})\leftarrow (P_{kl},P_{ij})$, producing $I'=\mathrm{Edit}_{\mathrm{patch}}(I)$.
We operate only on non-empty patches.
Solving tasks on $I'$ requires the model to reconstruct the intended visual structure in the image space, recover the correct graph topology, and then carry out visually grounded reasoning on the recovered structure.

\noindent\textbf{Cross-Graph Composition.}
To mirror compositional visual reasoning \cite{johnson2017clevr} over multiple objects, where the model must combine evidence across separate entities and their relations, we treat each “object” as a subgraph rendered in its own image. Given a sampled graph $G=(V,E)$, we partition $V$ into two disjoint, balanced subsets $V^{(1)}$ and $V^{(2)}$, inducing $G^{(1)}=(V^{(1)},E^{(1)})$ and $G^{(2)}=(V^{(2)},E^{(2)})$. We render them as two views $I_1$ and $I_2$. 
The prompt provides cross-graph links $E^{(\times)}\subseteq V^{(1)}\times V^{(2)}$, which specify how the two “objects” should be connected, forming $\tilde{G}=(V^{(1)}\cup V^{(2)},\,E^{(1)}\cup E^{(2)}\cup E^{(\times)})$. The model must bind nodes across $(I_1,I_2)$, apply the cross-links to compose $\tilde{G}$, enabling compositional reasoning by integrating multiple visual evidence.

\begin{table*}[t]\footnotesize
\centering

\caption{Overall performance of each MLLM across all tasks. Two evaluation metrics are reported: Acc (\%) and VGR-Score (VGR-S). \myred{} indicates that VGR-S is higher than Acc, \blue{} indicates that VGR-S is lower than Acc, and the absence of an arrow indicates equality. \ab{-} and \vb{-} highlight the best score in each task.}

\vspace{-0.4cm}

\scriptsize{
\resizebox{\linewidth}{!}{
\setlength\tabcolsep{2pt}
\renewcommand\arraystretch{1.0}
\setlength{\arrayrulewidth}{0.3mm}
\begin{tabular}{c||cc|cc|cc|cc|cc|cc|cc|cc|cc||cc|cc}
\hline

\rowcolor{HdrA}
\textbf{Settings}
& \multicolumn{18}{c||}{\textbf{Single-image}}
& \multicolumn{4}{c}{\textbf{Paired-image}}\\
\hline

\rowcolor{HdrB}
\textbf{Tasks}
& \multicolumn{2}{c|}{\textbf{Diameter}}
& \multicolumn{2}{c|}{\textbf{SD}}
& \multicolumn{2}{c|}{\textbf{CN}}
& \multicolumn{2}{c|}{\textbf{Cycle}}
& \multicolumn{2}{c|}{\textbf{TSP}}
& \multicolumn{2}{c|}{\textbf{Coloring}}
& \multicolumn{2}{c|}{\textbf{MVC}}
& \multicolumn{2}{c|}{\textbf{MCP}}
& \multicolumn{2}{c||}{\textbf{MIS}}
& \multicolumn{2}{c|}{\textbf{GED}}
& \multicolumn{2}{c}{\textbf{MCS}}\\
\hline

\rowcolor{HdrC}
\textbf{Metrics}
& \textbf{Acc} & \textbf{VGR-S}
& \textbf{Acc} & \textbf{VGR-S}
& \textbf{Acc} & \textbf{VGR-S}
& \textbf{Acc} & \textbf{VGR-S}
& \textbf{Acc} & \textbf{VGR-S}
& \textbf{Acc} & \textbf{VGR-S}
& \textbf{Acc} & \textbf{VGR-S}
& \textbf{Acc} & \textbf{VGR-S}
& \textbf{Acc} & \textbf{VGR-S}
& \textbf{Acc} & \textbf{VGR-S}
& \textbf{Acc} & \textbf{VGR-S}\\
\hline
\hline

\rowcolor{HdrD}
\multicolumn{23}{c}{\textbf{Reference}}\\
\hline
\hline
Human
& 100.0 & - & 100.0 & - & 100.0 & - & 100.0 & - & 83.2 & - & 73.5 & -
& 78.0 & - & 76.0 & - & 85.0 & - & 64.0 & - & 84.0 & - \\
\hline
\hline
\rowcolor{HdrD}
\multicolumn{23}{c}{\textbf{MLLM Description $\boldsymbol{\Rightarrow}$ LLMs}}\\
\hline
\hline
DeepSeek-V4-Pro
& 29.0 & 46.4\myred{} & 43.0 & 44.1\myred{} & 42.0 & 39.8\blue{} & 43.0 & 67.8\myred{} & 3.9 & 34.2\myred{} & 21.0 & 37.8\myred{}
& 27.0 & 31.2\myred{} & 30.0 & 38.8\myred{} & 28.0 & 30.2\myred{} & 4.0 & 32.6\myred{} & 0.0 & 34.8\myred{} \\
Qwen2.5-72B-Instruct
& 4.0 & 20.9\myred{} & 14.0 & 41.2\myred{} & 33.0 & 40.8\myred{} & 28.0 & 34.6\myred{} & 0.0 & 31.5\myred{} & 3.0 & 19.4\myred{}
& 21.0 & 23.9\myred{} & 12.0 & 23.8\myred{} & 9.0 & 40.4\myred{} & 0.0 & 8.0\myred{} & 0.0 & 19.9\myred{} \\
QwQ-32B
& 1.0 & 16.5\myred{} & 12.0 & 32.6\myred{} & 16.0 & 37.1\myred{} & 14.0 & 28.9\myred{} & 0.0 & 15.8\myred{} & 1.2 & 16.1\myred{}
& 4.0 & 18.8\myred{} & 3.0 & 10.9\myred{} & 1.0 & 30.4\myred{} & 1.0 & 22.1\myred{} & 0.0 & 12.6\myred{} \\
\hline
\hline
\rowcolor{HdrD}
\multicolumn{23}{c}{\textbf{Close Source MLLMs}}\\
\hline
\hline

Gemini-3-Pro
& \ab{64.0} & \vb{63.4}\blue{} & 65.0 & 55.7\blue{} & 86.0 & 76.2\blue{} & \ab{94.0} & 89.7\blue{}
& 56.2 & 42.1\blue{} & 40.0 & \vb{72.9}\myred{} & \ab{51.0} & \vb{59.6}\myred{} & \ab{55.0} & \vb{55.9}\myred{}
& \ab{55.0} & \vb{56.7}\myred{} & \ab{30.0} & 35.1\myred{} & 6.0 & 34.3\myred{} \\

Gemini-3-flash
& 60.0 & 58.8\blue{} & \ab{69.0} & 55.5\blue{} & \ab{89.0} & 78.8\blue{} & 77.0 & 79.2\myred{}
& \ab{58.8} & \vb{48.5}\blue{} & 37.5 & 66.1\myred{} & 44.0 & 44.3\myred{} & 41.0 & 40.6\blue{}
& 44.0 & 44.5\myred{} & 27.0 & 32.2\myred{} & 4.0 & 23.0\myred{} \\

GPT-5.2
& 57.0 & 60.6\myred{} & 58.0 & 55.9\blue{} & 87.0 & \vb{88.1}\myred{} & 93.0 & \vb{91.8}\blue{}
& 32.5 & 33.0\myred{} & 36.2 & 43.5\myred{} & 40.0 & 45.0\myred{} & 39.0 & 43.4\myred{}
& 39.0 & 44.9\myred{} & 23.0 & 30.5\myred{} & 2.0 & 20.3\myred{} \\

GPT-o1
& 41.0 & 50.4\myred{} & 54.0 & \vb{57.2}\myred{} & 64.0 & 66.1\myred{} & 67.0 & 70.2\myred{}
& 0.0 & 9.8\myred{} & 33.8 & 47.9\myred{} & 25.0 & 34.5\myred{} & 24.0 & 33.5\myred{}
& 27.0 & 36.0\myred{} & 24.0 & 25.2\myred{} & 1.0 & 13.3\myred{} \\

Qwen-VL-Max
& 38.0 & 45.3\myred{} & 53.0 & 48.8\blue{} & 76.0 & 70.8\blue{} & 76.0 & 74.5\blue{}
& 6.2 & 30.0\myred{} & 41.2 & 58.6\myred{} & 35.0 & 37.9\myred{} & 29.0 & 33.6\myred{}
& 37.0 & 38.7\myred{} & 28.0 & \vb{35.8}\myred{} & 0.0 & 30.3\myred{} \\

Qwen3-VL-Plus
& 45.0 & 49.2\myred{} & 61.0 & 54.2\blue{} & 75.0 & 75.3\myred{} & 81.0 & 78.5\blue{}
& 12.5 & 45.4\myred{} & 41.2 & 58.7\myred{} & 32.0 & 32.9\myred{} & 29.0 & 36.8\myred{}
& 30.0 & 31.1\myred{} & 44.0 & 43.3\blue{} & \ab{7.0} & 37.4\myred{} \\

Qwen3-VL-Flash
& 15.0 & 34.6\myred{} & 45.0 & 46.5\myred{} & 64.0 & 62.6\blue{} & 71.0 & 67.5\blue{}
& 0.0 & 22.1\myred{} & \ab{50.0} & 60.6\myred{} & 21.0 & 28.4\myred{} & 21.0 & 30.8\myred{}
& 25.0 & 31.3\myred{} & 21.0 & 29.1\myred{} & 3.0 & 29.8\myred{} \\

Grok4-Fast
& 15.0 & 26.9\myred{} & 33.0 & 39.2\myred{} & 47.0 & 47.2\myred{} & 31.0 & 39.9\myred{}
& 0.0 & 12.9\myred{} & 27.5 & 40.8\myred{} & 26.0 & 34.4\myred{} & 32.0 & 42.6\myred{}
& 27.0 & 37.6\myred{} & 25.0 & 24.5\blue{} & 0.0 & 14.1\myred{} \\

\hline
\hline

\rowcolor{HdrD}
\multicolumn{23}{c}{\textbf{Open Source MLLMs}}\\
\hline
\hline

Kimi-K2.5
& 30.0 & 48.1\myred{} & 46.0 & 52.5\myred{} & 66.0 & 70.3\myred{} & 41.0 & 67.7\myred{}
& 8.8 & 41.7\myred{} & 37.5 & 62.1\myred{} & 26.0 & 30.9\myred{} & 28.0 & 35.1\myred{}
& 31.0 & 31.8\myred{} & 10.0 & 34.5\myred{} & 1.0 & 39.5\myred{} \\

GLM-4.6V
& 24.0 & 36.3\myred{} & 52.0 & 49.3\blue{} & 66.0 & 68.4\myred{} & 72.0 & 72.7\myred{}
& 6.2 & 33.8\myred{} & 41.2 & 57.5\myred{} & 29.0 & 29.6\myred{} & 22.0 & 33.1\myred{}
& 21.0 & 28.5\myred{} & 18.0 & 26.2\myred{} & 2.0 & 26.0\myred{} \\

GLM-4.6-flashx
& 19.0 & 42.1\myred{} & 45.0 & 53.2\myred{} & 57.0 & 66.4\myred{} & 10.0 & 49.0\myred{}
& 0.0 & 25.9\myred{} & 18.8 & 48.7\myred{} & 9.0 & 41.9\myred{} & 12.0 & 47.6\myred{}
& 7.0 & 43.9\myred{} & 1.0 & 26.1\myred{} & 0.0 & 25.8\myred{} \\

Phi-3.5-Vision-Instruct
& 0.0 & 13.2\myred{} & 4.0 & 26.3\myred{} & 0.0 & 8.0\myred{} & 5.0 & 18.3\myred{}
& 0.0 & 5.5\myred{} & 26.2 & 26.2 & 1.0 & 26.4\myred{} & 1.0 & 20.0\myred{}
& 1.0 & 21.6\myred{} & 0.0 & 6.0\myred{} & 0.0 & 7.6\myred{} \\

QvQ-72B-Preview
& 3.0 & 41.7\myred{} & 10.0 & 36.8\myred{} & 18.0 & 54.2\myred{} & 31.0 & 44.4\myred{}
& 0.0 & 37.3\myred{} & 1.2 & 44.2\myred{} & 2.0 & 42.8\myred{} & 0.0 & 47.1\myred{}
& 0.0 & 45.3\myred{} & 0.0 & 32.7\myred{} & 0.0 & \vb{39.9}\myred{} \\

Qwen3.5-35B-A3B
& 9.0 & 19.0\myred{} & 27.0 & 23.0\blue{} & 19.0 & 14.3\blue{} & 18.0 & 20.7\myred{}
& 5.0 & 8.5\myred{} & 10.0 & 16.4\myred{} & 18.0 & 25.3\myred{} & 8.0 & 11.7\myred{}
& 6.0 & 32.1\myred{} & 4.0 & 38.6\myred{} & 0.0 & 29.6\myred{} \\

InternVL2.5-8B
& 0.0 & 15.2\myred{} & 5.0 & 27.2\myred{} & 17.0 & 31.6\myred{} & 20.0 & 30.7\myred{}
& 0.0 & 27.6\myred{} & 20.0 & 36.2\myred{} & 1.0 & 24.0\myred{} & 6.0 & 28.2\myred{}
& 6.0 & 27.3\myred{} & 5.0 & 10.8\myred{} & 1.0 & 16.6\myred{} \\

InternVL3-8B
& 0.0 & 17.0\myred{} & 13.0 & 40.6\myred{} & 14.0 & 45.1\myred{} & 18.0 & 31.7\myred{}
& 0.0 & 16.5\myred{} & 10.0 & 47.2\myred{} & 7.0 & 32.0\myred{} & 9.0 & 40.8\myred{}
& 6.0 & 28.4\myred{} & 1.0 & 6.8\myred{} & 0.0 & 11.7\myred{} \\

InternVL3.5-8B
& 10.0 & 30.6\myred{} & 15.0 & 37.3\myred{} & 19.0 & 35.9\myred{} & 14.0 & 33.5\myred{}
& 0.0 & 29.1\myred{} & 11.2 & 26.5\myred{} & 4.0 & 20.7\myred{} & 1.0 & 16.9\myred{}
& 7.0 & 35.1\myred{} & 6.0 & 17.0\myred{} & 1.0 & 22.9\myred{} \\

InternVL3-38B
& 0.0 & 44.1\myred{} & 11.0 & 30.4\myred{} & 7.0 & 40.6\myred{} & 1.0 & 29.7\myred{}
& 0.0 & 28.4\myred{} & 25.0 & 41.4\myred{} & 1.0 & 29.1\myred{} & 5.0 & 45.9\myred{}
& 0.0 & 39.8\myred{} & 0.0 & 28.6\myred{} & 0.0 & 35.9\myred{} \\

Qwen3.5-4B
& 11.0 & 39.9\myred{} & 23.0 & 48.4\myred{} & 15.0 & 52.0\myred{} & 33.0 & 44.8\myred{}
& 0.0 & 31.5\myred{} & 10.0 & 14.8\myred{} & 14.0 & 20.8\myred{} & 4.0 & 14.3\myred{}
& 6.0 & 27.1\myred{} & 2.0 & 25.1\myred{} & 0.0 & 14.7\myred{} \\

\hline
\end{tabular}}}

\label{tab:main_results}
\vspace{-0.5cm}
\end{table*}

\noindent\textbf{Graph Attentional Focusing.}
This edit targets selective visual attention \cite{anderson2018bottom} in the presence of distractors.
We assign node colors via $\mathrm{col}:V\rightarrow\mathcal{C}$ and choose a target connected subgraph $M=(V_M,E_M)$, setting $\mathrm{col}(v)=c^\star$ for all $v\in V_M$.
For any $u\notin V_M$ adjacent to $V_M$ (i.e., $\exists v\in V_M:(u,v)\in E$), we enforce $\mathrm{col}(u)\neq c^\star$; other nodes may also take $c^\star$, creating additional monochromatic components.
Let $G[c^\star]$ be the induced subgraph on $\{v\in V:\mathrm{col}(v)=c^\star\}$ with connected components $\mathcal{C}(G[c^\star])$.
We guarantee that $M$ is the largest component in $\mathcal{C}(G[c^\star])$; the model is instructed to find and recover $M$ and solve the downstream task on $M$ (discarding edges outside $E_M$ and any other smaller induced subgraphs).

\noindent\textbf{Spatially-Conditioned Recoloring.}
Unlike naive color swapping which can be resolved by text-based reasoning, we enforce recoloring by spatial relations to trigger the model to re-examine the image and perform visual spatial reasoning~\cite{liu2023visual}.
Formally, let each node $v$ have a 2D center $p(v)=(x_v, y_v)$ in the image. Given a sampled anchor $a\in V$ and direction $\delta\in\{\mathrm{left},\mathrm{right},\mathrm{above},\mathrm{below}\}$, we define the set of spatially selected nodes as:
\[
S_\delta(a) = \{v \in V \mid \text{coord}(v) \gtrless \text{coord}(a) \text{ w.r.t. } \delta\}.
\]
The model is then instructed to recolor all nodes in $S_\delta(a)$ to a target color $c^\dagger$, producing a modified image $I'$, and must perform this spatially grounded update before solving the task. The question further requires identifying the colors of the nodes along the answer path to ensure the update is assessed.

\subsection{Evaluation}
\label{sec:evaluation}

To obtain finer-grained evaluation signals, a natural choice, as in many existing benchmarks \cite{zhang2024mathverse}, is to use a stronger MLLM given the ground-truth answer to assess intermediate reasoning steps. However, for VGR, this is both costly and unnecessary. Because the underlying graph structure is fully preserved prior to visualization, the graph evidence needed to verify reasoning can be programmatically recovered and verbalized exactly, removing the need for a vision-capable evaluator. We therefore verbalize the structure-aware evidence required for correct reasoning and use an LLM for verification. This design mitigates hallucinated reasoning and inconsistencies between the reasoning process and the final answer, while substantially reducing overall evaluation cost.

\noindent \textbf{Graph Evidence Verbalization.} For each instance, we retain the original adjacency matrix of the source graph and instantiate predefined templates to verbalize its relational structure in natural language. In addition, we programmatically extract the structural changes associated with each GIE operation, such as the set of nodes recolored in spatially-conditioned recoloring. These editing-derived signals are likewise verbalized through dedicated templates and, together with the original relational structure, form the graph evidence $E_G$ used as the basis for subsequent scoring (details in §~\ref{sec:d}).

\noindent \textbf{LLM-based Scoring Agent.} We use GPT-5.1 as the scoring agent. For each response, the agent is given the model's full output, the graph evidence $E_G$, the ground-truth answer set, and task-specific scoring instructions. It is first asked to extract the key reasoning steps from the response, then evaluate each step against $E_G$ to determine whether the perception and reasoning reflected in that step are correct, and finally assess whether the predicted final answer matches any valid gold answer. To avoid introducing additional bias, we do not provide gold reasoning traces during scoring.

Formally, let $\mathcal{S}=\{s_1,\dots,s_m\}$ denote the extracted key steps, and define the step-level score:
\[
R_{\text{step}}=\frac{1}{m}\sum_{i=1}^{m}\mathbb{I}[s_i \text{ is consistent with } E_G].
\]
The overall \textit{Visual Graph Reasoning Score} (VGR-Score) is then formulated as:
\[
\mathrm{Score}=\lambda R_{\text{step}}+(1-\lambda)\mathbb{I}[\hat{y}\in\mathcal{A}^\ast],
\]
where $\hat{y}$ is the predicted final answer, $\mathcal{A}^\ast$ is the set of valid gold answers, and $\lambda\in[0,1]$ controls the trade-off between process correctness and answer correctness. Following prior work \cite{zhang2024mathverse} that places greater emphasis on the reasoning process, we set $\lambda=0.7$ in our experiments.
\section{Experiments}

\subsection{Experimental Setup}

\noindent \textbf{Dataset Split.} 
GraphVerse contains 11,000 samples. For more efficient evaluation, we constructed a smaller subset, \textit{testmini}, by sampling from the full dataset for our experiments and it contains 1,060 samples in total. The samples are drawn to closely match the distribution of the full test set, so as to preserve distributional consistency. This subset covers all tasks and all GIE strategies. Please refer to §~\ref{sec:b} for dataset statistics and additional details.

\noindent \textbf{Evaluation Models.} 
We evaluate a total of 19 MLLMs, including 11 open-source models—Kimi-K2.5 (thinking) (\citeyear{team2026kimi}), GLM-4.6V (\citeyear{zeng2025glm}), GLM-4.6-flashx (\citeyear{zeng2025glm}), Phi-3.5-Vision-Instruct (\citeyear{abdin2024phi3}), QvQ-72B-Preview (\citeyear{wang2024qwen2}), Qwen3.5-35B-A3B (\citeyear{qwen3.5}), 
InternVL2.5-8B (\citeyear{chen2024expanding}), InternVL3-8B (\citeyear{zhu2025internvl3}), InternVL3.5-8B (\citeyear{wang2025internvl3}), InternVL3-38B (\citeyear{chen2024expanding}), and Qwen3.5-4B (\citeyear{qwen3.5})—and 8 closed-source models—Gemini (\citeyear{team2023gemini}) (3-flash, 3-pro), GPT (\citeyear{achiam2023gpt})(o1, 5.2), Qwen (\citeyear{bai2025qwen3}) (VL-Max, 3-VL-Plus, 3-VL-Flash), and Grok4-Fast. Moreover, we add a text-only baseline: An MLLM (GPT-5.2) first captions the edited image, and a text-only LLM then solves the task. We test three LLMs under this setting: DeepSeek-V4-Pro (\citeyear{deepseekai2026deepseekv4}), Qwen2.5-72B-Instruct (\citeyear{DBLP:qwen2.5}), and QwQ-32B (\citeyear{qwq32b}). We further analyze the robustness of VGR-S in §~\ref{sec:vgr_robustness}.

\begin{figure}[h]
    \centering
    \includegraphics[width=0.98\linewidth]{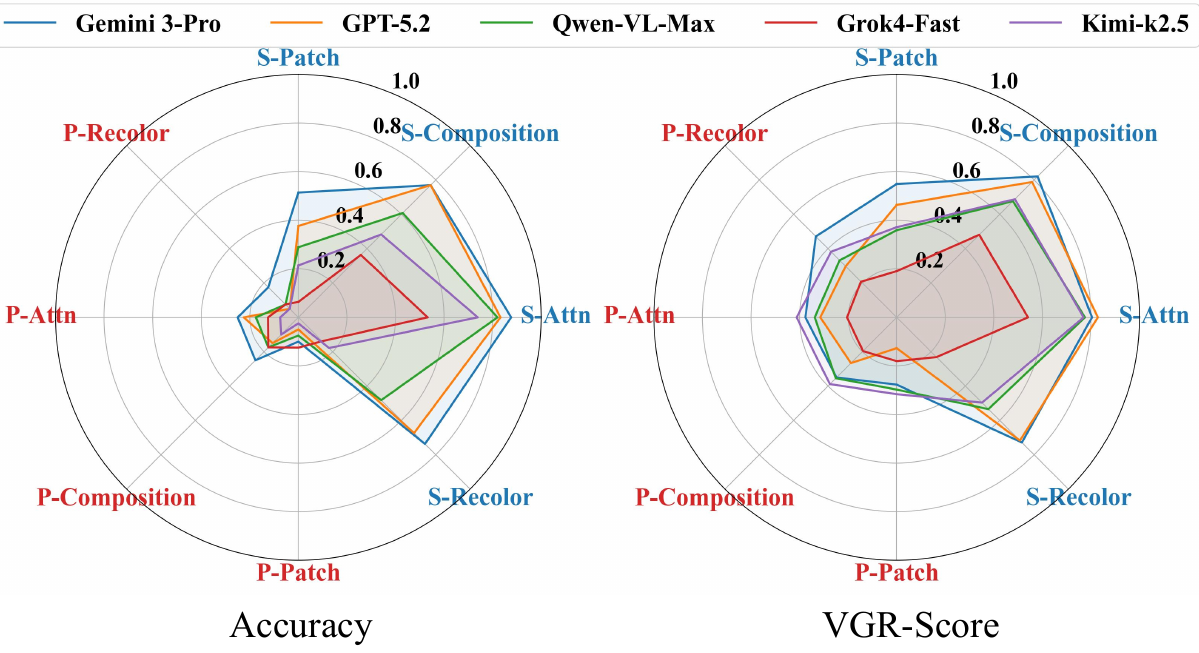}
    \caption{Radar chart of model performance on VGR under different GIE operations. P- and S- denote Paired-Image and Single-Image VGR, respectively.}
    \label{fig:radar_chart}
    \vspace{-0.6cm}
\end{figure}

\subsection{Overall Results}

\noindent $\blacktriangleright$ \textbf{Visual graph reasoning remains a major challenge for MLLMs.}  
Even the strongest current models, including Gemini-3-Pro and GPT-5.2, still leave substantial room for improvement on many VGR tasks. For example, on MCS, their accuracy remains below 10\%, and they trail the human baseline by an average of 35\% across all tasks. Moreover, the text-only pipeline performs poorly, especially on hard single- and paired-image tasks, showing that direct visual access is crucial for GraphVerse. Meanwhile, open-source models perform markedly worse than closed-source ones, likely reflecting differences in model scale and design.

\noindent $\blacktriangleright$ \textbf{Different MLLMs specialize in different VGR tasks, yet all struggle much more with paired-image reasoning.} As shown in Figure~\ref{fig:radar_chart}, all models perform substantially worse on paired-image VGR than in the single-image setting. For instance, Kimi-K2.5 trails by 34\% on Acc and about 20\% on VGR-S on average, \emph{while the gap becomes even larger for stronger models}: Gemini-3-Pro drops by more than 50\% in Acc and over 30\% in VGR-S. A closer look at the specific GIE settings, which reflect different types of VGR demands, further shows that models exhibit distinct strengths. For example, Grok4 remains behind on nearly all other dimensions in VGR-S, yet still outperforms GPT-5.2 on paired-image patch perturbation setting.

\noindent $\blacktriangleright$ \textbf{Visual reasoning complexity matters more than theoretical task complexity.} In text-only graph problem benchmarks~\cite{TangZLCL25}, a common finding is that \textbf{\rosetag{\texttt{NPH}}} tasks are generally more challenging than \textbf{\tealtag{\texttt{Poly}}} tasks because of their greater intrinsic complexity. After introducing visual reasoning, this trend still holds on average: as shown in Table~\ref{tab:poly_nphard}, MLLMs struggle more with \textbf{\rosetag{\texttt{NPH}}} tasks overall. That said, an important exception emerges. For \textbf{\rosetag{\texttt{NPH}}} tasks that require limited visual reasoning, such as Cycle, nearly 95\% of models achieve higher accuracy than on at least one \textbf{\tealtag{\texttt{Poly}}} task. For closed-source models, the average accuracy on Cycle even exceeds that on \textbf{\tealtag{\texttt{Poly}}} tasks by 17.04\% in Acc and 16.94\% in VGR-S. This suggests that in VGR, task difficulty is shaped not only by theoretical complexity, but even more strongly by the complexity of the required visual reasoning. 

\noindent $\blacktriangleright$ \textbf{Final-answer accuracy can mask deficiencies in the reasoning process.} Table~\ref{tab:main_results} clearly reveals a substantial mismatch between reasoning quality and accuracy. On harder tasks, models may achieve near-zero accuracy while still obtaining much higher VGR-S, suggesting partially correct reasoning despite incorrect final answers. In many cases, VGR-S exceeds accuracy by 30\%–40\%, implying that errors may arise from only a few critical reasoning steps. Conversely, for stronger models like Gemini-3-Pro, VGR-S can also fall more than 10\% below accuracy on some tasks, indicating that even powerful models do not always reason in a stable or faithful manner.

\begin{table}[t]\scriptsize
\vspace{-5pt}
\centering

\caption{Results of different MLLM families on Polynomial and NP-Hard tasks.
Each family’s score is averaged over its corresponding models in Table~\ref{tab:main_results}, and \blue{} denotes the relative drop from Polynomial to NP-Hard tasks.}
\vspace{-6pt}

\resizebox{\linewidth}{!}{%
    \setlength\tabcolsep{8pt}
    \renewcommand\arraystretch{1.2}
    \setlength{\arrayrulewidth}{0.3mm}
    \begin{tabular}{c||cc|cc}
        \hline \thickhline
        \rowcolor{CadetBlue!20}
        Models
        & \multicolumn{2}{c|}{\textbf{Polynomial}}
        & \multicolumn{2}{c}{\textbf{NP-Hard}} \\
        \rowcolor{CadetBlue!20}
        & \textbf{Acc}
        & \textbf{VGR-S}
        & \textbf{Acc}
        & \textbf{VGR-S} \\
        \hline\hline

        \textbf{GPT Family}      & 60.17 & 63.05 & 31.66 \blue{47.38} & 38.92 \blue{38.27} \\
        \textbf{Qwen Family}     & 33.72 & 45.37 & 19.03 \blue{43.56} & 35.68 \blue{21.36} \\
        \textbf{Gemini Family}   & 72.17 & 64.73 & 45.03 \blue{37.61} & 51.54 \blue{20.38} \\
        \textbf{GLM Family}      & 43.83 & 52.62 & 16.82 \blue{61.62} & 38.52 \blue{26.80} \\

        \hline
    \end{tabular}%
}

\label{tab:poly_nphard}
\vspace{-14pt}
\end{table}

\begin{table*}[t]
\centering
\footnotesize
\caption{Performance comparison under training-free and training-based settings across four task categories. \myred{} indicates absolute improvements over direct inference. The table reports acc only; VGR-S results are deferred to §~\ref{sec:e}. For Gemini-3-Pro, we apply PoT and denote the variant as Coder. For GPT-5.2, we use Codex with PoT.}
\vspace{-0.2cm}

\setlength{\tabcolsep}{5pt}
\renewcommand{\arraystretch}{1.35}
\arrayrulecolor{black}
\resizebox{\linewidth}{!}{%
\begin{tabular}{c||ccccc||c||ccccc}
\hline\hline

\multicolumn{12}{c}{\textbf{Training-Free}} \\
\hline

\multirow{2}{*}{\textbf{Setting}}
& \multicolumn{5}{c||}{\textbf{GPT-5.2}} 
& \multirow{2}{*}{\textbf{Setting}}
& \multicolumn{5}{c}{\textbf{Gemini-3-Pro}} \\
\cline{2-6} \cline{8-12}

& \textbf{Single-Image} & \textbf{Paired Image} & \textbf{Polynomial} & \textbf{NP-Hard} & \textbf{Average}
& & \textbf{Single-Image} & \textbf{Paired Image} & \textbf{Polynomial} & \textbf{NP-Hard} & \textbf{Average} \\
\hline\hline

\textbf{Direct Inference} 
& 53.5 & 12.5 & 67.3 & 38.3 & 46.1
& \textbf{Direct Inference} 
& 62.9 & 18.0 & 71.6 & 48.9 & 54.7 \\

\hline
\rowcolor[HTML]{D7F6FF}
\textbf{GPT-5.2-Codex} 
& 59.9 {\color{orange}$\uparrow$6.4}
& 17.5 {\color{orange}$\uparrow$5.0}
& 71.1 {\color{orange}$\uparrow$3.8}
& 45.1 {\color{orange}$\uparrow$6.8}
& 52.2 {\color{orange}$\uparrow$6.1}
& \textbf{Gemini-3-Pro-Coder}
& 66.1 {\color{orange}$\uparrow$3.2}
& 22.5 {\color{orange}$\uparrow$4.5}
& 74.1 {\color{orange}$\uparrow$2.5}
& 52.2 {\color{orange}$\uparrow$3.3}
& 58.2 {\color{orange}$\uparrow$3.5} \\
\hline\hline

\multicolumn{12}{c}{\textbf{Training-based}} \\
\hline

\multirow{2}{*}{\textbf{Setting}}
& \multicolumn{5}{c||}{\textbf{Qwen3-VL-8B-Instruct}} 
& \multirow{2}{*}{\textbf{Setting}}
& \multicolumn{5}{c}{\textbf{Qwen3-VL-2B-Instruct}} \\
\cline{2-6} \cline{8-12}

& \textbf{Single-Image} & \textbf{Paired Image} & \textbf{Polynomial} & \textbf{NP-Hard} & \textbf{Average}
& & \textbf{Single-Image} & \textbf{Paired Image} & \textbf{Polynomial} & \textbf{NP-Hard} & \textbf{Average} \\
\hline\hline

\textbf{Direct Inference}
& 17.1 & 0.0 & 30.5 & 7.8 & 14.0
& \textbf{Direct Inference}
& 3.1 & 1.5 & 3.1 & 2.7 & 2.8 \\
\hline
\textbf{SFT w/o GIE}
& 19.3 {\color{orange}$\uparrow$2.2}
& 0.7 {\color{orange}$\uparrow$0.7}
& 33.3 {\color{orange}$\uparrow$2.8}
& 9.4 {\color{orange}$\uparrow$1.6}
& 15.9 {\color{orange}$\uparrow$1.9}
& \textbf{SFT w/o GIE}
& 7.0 {\color{orange}$\uparrow$3.9}
& 2.0 {\color{orange}$\uparrow$0.5}
& 9.0 {\color{orange}$\uparrow$5.9}
& 5.0 {\color{orange}$\uparrow$2.3}
& 6.1 {\color{orange}$\uparrow$3.3} \\

\rowcolor[HTML]{D7F6FF}
\textbf{SFT}
& 20.6 {\color{orange}$\uparrow$3.5}
& 2.9 {\color{orange}$\uparrow$2.9}
& 36.0 {\color{orange}$\uparrow$5.5}
& 10.4 {\color{orange}$\uparrow$2.6}
& 17.4 {\color{orange}$\uparrow$3.4}
& \textbf{SFT}
& 8.4 {\color{orange}$\uparrow$5.3}
& 4.1 {\color{orange}$\uparrow$2.6}
& 8.2 {\color{orange}$\uparrow$5.1}
& 7.4 {\color{orange}$\uparrow$4.7}
& 7.6 {\color{orange}$\uparrow$4.8} \\

\textbf{RL w/o GIE}
& 33.5 {\color{orange}$\uparrow$16.4}
& 17.5 {\color{orange}$\uparrow$17.5}
& 35.9 {\color{orange}$\uparrow$5.4}
& 28.6 {\color{orange}$\uparrow$20.8}
& 30.6 {\color{orange}$\uparrow$16.6}
& \textbf{RL w/o GIE}
& 3.9 {\color{orange}$\uparrow$0.8}
& 2.2 {\color{orange}$\uparrow$0.7}
& 5.7 {\color{orange}$\uparrow$2.6}
& 2.8 {\color{orange}$\uparrow$0.1}
& 3.6 {\color{orange}$\uparrow$0.8} \\

\hline

\rowcolor[HTML]{D7F6FF}
\textbf{RL}
& 38.9 {\color{orange}$\uparrow$21.8}
& 16.1 {\color{orange}$\uparrow$16.1}
& 37.3 {\color{orange}$\uparrow$6.8}
& 33.8 {\color{orange}$\uparrow$26.0}
& 34.8 {\color{orange}$\uparrow$20.8}
& \textbf{RL}
& 7.1 {\color{orange}$\uparrow$4.0}
& 1.8 {\color{orange}$\uparrow$0.3}
& 7.3 {\color{orange}$\uparrow$4.2}
& 5.7 {\color{orange}$\uparrow$3.0}
& 6.1 {\color{orange}$\uparrow$3.3} \\
\hline\hline

\end{tabular}%
}

\label{tab:main_result_replicate}
\vspace{-0.5cm}
\end{table*}
\subsection{Improving MLLMs’ VGR Capabilities}
\label{sec:5.3}
The findings reveal several limitations of MLLMs on VGR. We therefore explore how to improve their VGR from two complementary directions: training-based and training-free methods. For training-free methods, we employ Program-of-Thoughts (PoT) prompting \cite{chen2022program} to enhance the procedural graph reasoning ability of MLLMs and improve their VGR performance. For training-based methods, we additionally construct 3,500 samples as the training set, with no overlap with the test set. We train  (details in §~\ref{sec:e}) and evaluate on these datasets to measure in-domain gains.

\noindent $\blacktriangleright$ \textbf{Multiple paths to better VGR.} 
As shown in Table~\ref{tab:main_result_replicate}, both PoT prompting and post-training consistently improve MLLMs' VGR performance. Post-training with GIE-augmented data yields larger gains than a matched no-GIE training set, demonstrating the value of GIE for in-domain VGR. Moreover, RL brings a 17.4\% gain over SFT on Qwen3-8B, but underperforms SFT by 1.5\% on the 2B model, suggesting that larger models can better unlock their VGR potential through RL, while smaller models benefit more from the stronger behavioral regularization and imitation-style supervision provided by SFT. Moreover, in §~\ref{sec:ablations_training}, we further demonstrate the effectiveness of GIE through a training-based ablation of direct visual access.

\begin{figure}[h]
    \centering
    \vspace{-0.3cm}
    \includegraphics[width=1\linewidth]{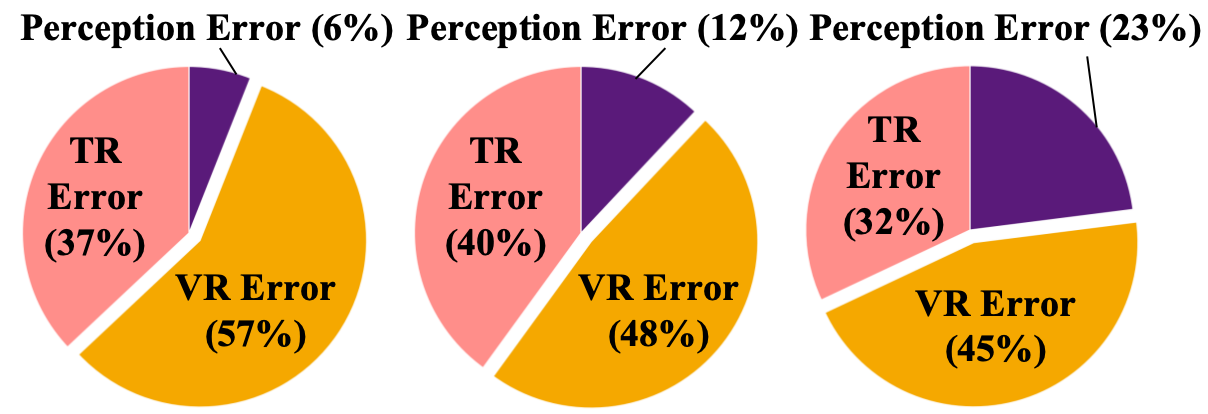}
    \caption{Pie Chart of Failure Mode Analysis for Representative MLLMs. TR and VR denote text-based and vision-based reasoning, respectively.}
    \label{fig:piechart}
    \vspace{-0.6cm}
\end{figure}

\subsection{In-Depth Failure Mode Analysis}
To better understand where MLLMs fall short in VGR, we manually categorize incorrect responses from three representative models into text-based reasoning (TR), vision-based reasoning (VR), or perception errors, as shown in Figure~\ref{fig:piechart}.

\noindent $\blacktriangleright$ \textbf{Vision-based reasoning is the dominant failure mode of MLLMs.} Across all three models, visual reasoning errors account for the largest share, with an average of 50\%. As model reasoning ability improves, from Kimi-K2.5 to Gemini-3-Pro, perception errors decrease while visual reasoning errors become increasingly dominant. This suggests that, for current frontier models, the main bottleneck lies in visual reasoning rather than perception, which further justifies our benchmark’s emphasis on vision-based reasoning.

\subsection{Transferability Analysis}
To examine whether GraphVerse improves transferable reasoning beyond VGR, we evaluate RL-trained Qwen3-VL-8B on the out-of-domain MathVista benchmark~(\citeyear{lu2023mathvista}). We follow the training setup in Section~\ref{sec:5.3}, with the reward design detailed in Section~\ref{sec:e}; results are reported in Table~\ref{tab:transfer}.

\begin{table}[h]
\centering
\scriptsize
\vspace{-0.3cm}
\caption{Category-wise accuracy (\%) comparison across different settings in MathVista. Overall denotes overall accuracy, MC denotes multiple-choice, and Math denotes math-targeted VQA. \myred{} arrow denotes the relative improvement. RL-Base refers to RL w/o GIE (Table~\ref{tab:main_result_replicate}).}
\vspace{-0.2cm}

\setlength{\tabcolsep}{2pt}
\renewcommand{\arraystretch}{1.0}
\begin{tabular}{l|cc|cc|c}
\toprule
\multirow{2}{*}{\textbf{Method}} 
& \multicolumn{2}{c|}{\textbf{Question Type}} 
& \multicolumn{2}{c|}{\textbf{VQA Category}} 
& \multirow{2}{*}{\textbf{Overall}} \\
\cline{2-5}
& \textbf{MC} & \textbf{Free Form} & \textbf{General} & \textbf{Math} & \\
\midrule
\textbf{DI}      
& 79.07 
& 65.65 
& 73.04 
& 72.78 
& 72.90 \\
\textbf{RL-Base} 
& 79.26 \myred{0.24} 
& 66.09 \myred{0.67} 
& 73.48 \myred{0.60} 
& 72.96 \myred{0.25} 
& 73.20 \myred{0.41} \\
\rowcolor[HTML]{D7F6FF}
\textbf{RL}      
& \textbf{80.56} \myred{1.88} 
& \textbf{66.30} \myred{0.99} 
& \textbf{74.13} \myred{1.49} 
& \textbf{73.89} \myred{1.53} 
& \textbf{74.00} \myred{1.51} \\
\bottomrule
\end{tabular}

\label{tab:transfer}
\vspace{-0.3cm}
\end{table}

\noindent $\blacktriangleright$ \textbf{What transfers from GraphVerse is not graph knowledge, but reasoning itself.} Compared with direct inference, both RL paradigms improve mathematical and general reasoning, with larger gains when trained on GIE-augmented data, achieving an overall relative improvement of 1.51\%, further demonstrating the effectiveness of our proposed GIE strategy. This nontrivial transfer beyond the original domain underscores the strength of GraphVerse as a source of broadly useful supervision.
\section{Conclusion}
We identify key limitations of existing VGR benchmarks and introduce GraphVerse to address them. With broader task types, GIE strategies that enforce genuine visual reasoning, diverse difficulty, realistic grounding, and fine-grained evaluation, GraphVerse enables a more faithful and comprehensive assessment of MLLM visual graph reasoning. Extensive experiments not only validate the effectiveness of GIE, but also reveal major weaknesses of current MLLMs, provide insights for improvement, and further show that VGR-acquired abilities can transfer beyond the original domain. We hope GraphVerse can serve as a strong benchmark for future visual graph reasoning research.

\section{Limitations}
Due to space limitations, we are unable to include more qualitative cases in the experimental section to further support our conclusions. Although these examples are provided in the appendix, presenting more of them in the main text could have made our findings clearer. Moreover, this paper does not extensively explore broader multi-image VGR settings, as multi-image reasoning is often treated as a separate research direction. Instead, we focus on first establishing a systematic VGR benchmark and an initial diagnostic probe for paired-image reasoning. We will continue maintaining GraphVerse and extend it to richer multi-image VGR scenarios in future releases, while encouraging further research on broader multi-image visual graph reasoning.

\section{Acknowledgments}
We sincerely thank the anonymous reviewers, AC and SAC for their valuable feedback. This research is partially supported by National Natural Science Foundation of China (62402320), Shanghai Science and Technology Program (24YF2731000), and the NYU Shanghai Boost Fund.

\bibliography{custom}

@inproceedings{TangZLCL25,
  author       = {Jianheng Tang and
                  Qifan Zhang and
                  Yuhan Li and
                  Nuo Chen and
                  Jia Li},
  title        = {GraphArena: Evaluating and Exploring Large Language Models on Graph
                  Computation},
  booktitle    = {The Thirteenth International Conference on Learning Representations,
                  {ICLR} 2025, Singapore, April 24-28, 2025},
  publisher    = {OpenReview.net},
  year         = {2025},
  url          = {https://openreview.net/forum?id=Y1r9yCMzeA},
  bibsource    = {dblp computer science bibliography, https://dblp.org}
}

@inproceedings{Ley02,
  author       = {Michael Ley},
  editor       = {Alberto H. F. Laender and
                  Arlindo L. Oliveira},
  title        = {The {DBLP} Computer Science Bibliography: Evolution, Research Issues,
                  Perspectives},
  booktitle    = {String Processing and Information Retrieval, 9th International Symposium,
                  {SPIRE} 2002, Lisbon, Portugal, September 11-13, 2002, Proceedings},
  series       = {Lecture Notes in Computer Science},
  volume       = {2476},
  pages        = {1--10},
  publisher    = {Springer},
  year         = {2002},
  url          = {https://doi.org/10.1007/3-540-45735-6\_1},
  doi          = {10.1007/3-540-45735-6\_1},
  bibsource    = {dblp computer science bibliography, https://dblp.org}
}

@inproceedings{RossiA15,
  author       = {Ryan A. Rossi and
                  Nesreen K. Ahmed},
  editor       = {Blai Bonet and
                  Sven Koenig},
  title        = {The Network Data Repository with Interactive Graph Analytics and Visualization},
  booktitle    = {Proceedings of the Twenty-Ninth {AAAI} Conference on Artificial Intelligence,
                  January 25-30, 2015, Austin, Texas, {USA}},
  pages        = {4292--4293},
  publisher    = {{AAAI} Press},
  year         = {2015},
  url          = {https://doi.org/10.1609/aaai.v29i1.9277},
  doi          = {10.1609/AAAI.V29I1.9277},
  bibsource    = {dblp computer science bibliography, https://dblp.org}
}

@article{BizerLKABCH09,
  author       = {Christian Bizer and
                  Jens Lehmann and
                  Georgi Kobilarov and
                  S{\"{o}}ren Auer and
                  Christian Becker and
                  Richard Cyganiak and
                  Sebastian Hellmann},
  title        = {DBpedia - {A} crystallization point for the Web of Data},
  journal      = {J. Web Semant.},
  volume       = {7},
  number       = {3},
  pages        = {154--165},
  year         = {2009},
  url          = {https://doi.org/10.1016/j.websem.2009.07.002},
  doi          = {10.1016/J.WEBSEM.2009.07.002},
  bibsource    = {dblp computer science bibliography, https://dblp.org}
}

@article{DBLP:journals/jcisd/NakataS17,
  author       = {Maho Nakata and
                  Tomomi Shimazaki},
  title        = {PubChemQC Project: {A} Large-Scale First-Principles Electronic Structure
                  Database for Data-Driven Chemistry},
  journal      = {J. Chem. Inf. Model.},
  volume       = {57},
  number       = {6},
  pages        = {1300--1308},
  year         = {2017},
  url          = {https://doi.org/10.1021/acs.jcim.7b00083},
  doi          = {10.1021/ACS.JCIM.7B00083},
  bibsource    = {dblp computer science bibliography, https://dblp.org}
}

@inproceedings{DBLP:conf/nips/HuFRNDL21,
  author       = {Weihua Hu and
                  Matthias Fey and
                  Hongyu Ren and
                  Maho Nakata and
                  Yuxiao Dong and
                  Jure Leskovec},
  editor       = {Joaquin Vanschoren and
                  Sai{-}Kit Yeung},
  title        = {{OGB-LSC:} {A} Large-Scale Challenge for Machine Learning on Graphs},
  booktitle    = {Proceedings of the Neural Information Processing Systems Track on
                  Datasets and Benchmarks 1, NeurIPS Datasets and Benchmarks 2021, December
                  2021, virtual},
  year         = {2021},
  url          = {https://datasets-benchmarks-proceedings.neurips.cc/paper/2021/hash/db8e1af0cb3aca1ae2d0018624204529-Abstract-round2.html},
  bibsource    = {dblp computer science bibliography, https://dblp.org}
}

@inproceedings{DBLP:conf/gd/EllsonGKNW00,
  author       = {John Ellson and
                  Emden R. Gansner and
                  Eleftherios Koutsofios and
                  Stephen C. North and
                  Gordon Woodhull},
  editor       = {Petra Mutzel and
                  Michael J{\"{u}}nger and
                  Sebastian Leipert},
  title        = {Graphviz - Open Source Graph Drawing Tools},
  booktitle    = {Graph Drawing, 9th International Symposium, {GD} 2001 Vienna, Austria,
                  September 23-26, 2001, Revised Papers},
  series       = {Lecture Notes in Computer Science},
  volume       = {2265},
  pages        = {483--484},
  publisher    = {Springer},
  year         = {2001},
  url          = {https://doi.org/10.1007/3-540-45848-4\_57},
  doi          = {10.1007/3-540-45848-4\_57},
  bibsource    = {dblp computer science bibliography, https://dblp.org}
}

@article{bai2025qwen2,
  title={Qwen2. 5-vl technical report},
  author={Bai, Shuai and Chen, Keqin and Liu, Xuejing and Wang, Jialin and Ge, Wenbin and Song, Sibo and Dang, Kai and Wang, Peng and Wang, Shijie and Tang, Jun and others},
  journal={arXiv preprint arXiv:2502.13923},
  year={2025}
}

@article{team2025kimi,
  title={Kimi-vl technical report},
  author={Team, Kimi and Du, Angang and Yin, Bohong and Xing, Bowei and Qu, Bowen and Wang, Bowen and Chen, Cheng and Zhang, Chenlin and Du, Chenzhuang and Wei, Chu and others},
  journal={arXiv preprint arXiv:2504.07491},
  year={2025}
}

@inproceedings{yue2024mmmu,
  title={Mmmu: A massive multi-discipline multimodal understanding and reasoning benchmark for expert agi},
  author={Yue, Xiang and Ni, Yuansheng and Zhang, Kai and Zheng, Tianyu and Liu, Ruoqi and Zhang, Ge and Stevens, Samuel and Jiang, Dongfu and Ren, Weiming and Sun, Yuxuan and others},
  booktitle={Proceedings of the IEEE/CVF Conference on Computer Vision and Pattern Recognition},
  pages={9556--9567},
  year={2024}
}

@article{yu2023mm,
  title={Mm-vet: Evaluating large multimodal models for integrated capabilities},
  author={Yu, Weihao and Yang, Zhengyuan and Li, Linjie and Wang, Jianfeng and Lin, Kevin and Liu, Zicheng and Wang, Xinchao and Wang, Lijuan},
  journal={arXiv preprint arXiv:2308.02490},
  year={2023}
}

@article{lu2023mathvista,
  title={Mathvista: Evaluating mathematical reasoning of foundation models in visual contexts},
  author={Lu, Pan and Bansal, Hritik and Xia, Tony and Liu, Jiacheng and Li, Chunyuan and Hajishirzi, Hannaneh and Cheng, Hao and Chang, Kai-Wei and Galley, Michel and Gao, Jianfeng},
  journal={arXiv preprint arXiv:2310.02255},
  year={2023}
}

@inproceedings{zhang2024mathverse,
  title={Mathverse: Does your multi-modal llm truly see the diagrams in visual math problems?},
  author={Zhang, Renrui and Jiang, Dongzhi and Zhang, Yichi and Lin, Haokun and Guo, Ziyu and Qiu, Pengshuo and Zhou, Aojun and Lu, Pan and Chang, Kai-Wei and Qiao, Yu and others},
  booktitle={European Conference on Computer Vision},
  pages={169--186},
  year={2024},
  organization={Springer}
}

@inproceedings{yue2025mmmu,
  title={Mmmu-pro: A more robust multi-discipline multimodal understanding benchmark},
  author={Yue, Xiang and Zheng, Tianyu and Ni, Yuansheng and Wang, Yubo and Zhang, Kai and Tong, Shengbang and Sun, Yuxuan and Yu, Botao and Zhang, Ge and Sun, Huan and others},
  booktitle={Proceedings of the 63rd Annual Meeting of the Association for Computational Linguistics (Volume 1: Long Papers)},
  pages={15134--15186},
  year={2025}
}

@article{shchur2018pitfalls,
  title={Pitfalls of graph neural network evaluation},
  author={Shchur, Oleksandr and Mumme, Maximilian and Bojchevski, Aleksandar and G{\"u}nnemann, Stephan},
  journal={arXiv preprint arXiv:1811.05868},
  year={2018}
}

@article{hu2020open,
  title={Open graph benchmark: Datasets for machine learning on graphs},
  author={Hu, Weihua and Fey, Matthias and Zitnik, Marinka and Dong, Yuxiao and Ren, Hongyu and Liu, Bowen and Catasta, Michele and Leskovec, Jure},
  journal={Advances in neural information processing systems},
  volume={33},
  pages={22118--22133},
  year={2020}
}

@article{hamilton2017inductive,
  title={Inductive representation learning on large graphs},
  author={Hamilton, Will and Ying, Zhitao and Leskovec, Jure},
  journal={Advances in neural information processing systems},
  volume={30},
  year={2017}
}

@article{li2024visiongraph,
  title={Visiongraph: Leveraging large multimodal models for graph theory problems in visual context},
  author={Li, Yunxin and Hu, Baotian and Shi, Haoyuan and Wang, Wei and Wang, Longyue and Zhang, Min},
  journal={arXiv preprint arXiv:2405.04950},
  year={2024}
}

@article{wei2024gita,
  title={Gita: Graph to visual and textual integration for vision-language graph reasoning},
  author={Wei, Yanbin and Fu, Shuai and Jiang, Weisen and Zhang, Zejian and Zeng, Zhixiong and Wu, Qi and Kwok, James and Zhang, Yu},
  journal={Advances in Neural Information Processing Systems},
  volume={37},
  pages={44--72},
  year={2024}
}

@inproceedings{zhu2025benchmarking,
  title={Benchmarking and improving large vision-language models for fundamental visual graph understanding and reasoning},
  author={Zhu, Yingjie and Bai, Xuefeng and Chen, Kehai and Xiang, Yang and Yu, Jun and Zhang, Min},
  booktitle={Proceedings of the 63rd Annual Meeting of the Association for Computational Linguistics (Volume 1: Long Papers)},
  pages={30678--30701},
  year={2025}
}

@article{babaiee2025visual,
  title={Visual Graph Arena: Evaluating Visual Conceptualization of Vision and Multimodal Large Language Models},
  author={Babaiee, Zahra and Kiasari, Peyman M and Rus, Daniela and Grosu, Radu},
  journal={arXiv preprint arXiv:2506.06242},
  year={2025}
}

@article{xu2025visulogic,
  title={Visulogic: A benchmark for evaluating visual reasoning in multi-modal large language models},
  author={Xu, Weiye and Wang, Jiahao and Wang, Weiyun and Chen, Zhe and Zhou, Wengang and Yang, Aijun and Lu, Lewei and Li, Houqiang and Wang, Xiaohua and Zhu, Xizhou and others},
  journal={arXiv preprint arXiv:2504.15279},
  year={2025}
}

@article{yuan2025mme,
  title={MME-Reasoning: A Comprehensive Benchmark for Logical Reasoning in MLLMs},
  author={Yuan, Jiakang and Peng, Tianshuo and Jiang, Yilei and Lu, Yiting and Zhang, Renrui and Feng, Kaituo and Fu, Chaoyou and Chen, Tao and Bai, Lei and Zhang, Bo and others},
  journal={arXiv preprint arXiv:2505.21327},
  year={2025}
}

@inproceedings{johnson2017clevr,
  title={Clevr: A diagnostic dataset for compositional language and elementary visual reasoning},
  author={Johnson, Justin and Hariharan, Bharath and Van Der Maaten, Laurens and Fei-Fei, Li and Lawrence Zitnick, C and Girshick, Ross},
  booktitle={Proceedings of the IEEE conference on computer vision and pattern recognition},
  pages={2901--2910},
  year={2017}
}

@inproceedings{hudson2019gqa,
  title={Gqa: A new dataset for real-world visual reasoning and compositional question answering},
  author={Hudson, Drew A and Manning, Christopher D},
  booktitle={Proceedings of the IEEE/CVF conference on computer vision and pattern recognition},
  pages={6700--6709},
  year={2019}
}

@article{mao2022clevrer,
  title={Clevrer-humans: Describing physical and causal events the human way},
  author={Mao, Jiayuan and Yang, Xuelin and Zhang, Xikun and Goodman, Noah and Wu, Jiajun},
  journal={Advances in Neural Information Processing Systems},
  volume={35},
  pages={7755--7768},
  year={2022}
}

@inproceedings{huang2021seeing,
  title={Seeing out of the box: End-to-end pre-training for vision-language representation learning},
  author={Huang, Zhicheng and Zeng, Zhaoyang and Huang, Yupan and Liu, Bei and Fu, Dongmei and Fu, Jianlong},
  booktitle={Proceedings of the IEEE/CVF conference on computer vision and pattern recognition},
  pages={12976--12985},
  year={2021}
}

@inproceedings{zhang2019raven,
  title={Raven: A dataset for relational and analogical visual reasoning},
  author={Zhang, Chi and Gao, Feng and Jia, Baoxiong and Zhu, Yixin and Zhu, Song-Chun},
  booktitle={Proceedings of the IEEE/CVF conference on computer vision and pattern recognition},
  pages={5317--5327},
  year={2019}
}

@article{fatemi2023talk,
  title={Talk like a graph: Encoding graphs for large language models},
  author={Fatemi, Bahare and Halcrow, Jonathan and Perozzi, Bryan},
  journal={arXiv preprint arXiv:2310.04560},
  year={2023}
}

@article{wang2023can,
  title={Can language models solve graph problems in natural language?},
  author={Wang, Heng and Feng, Shangbin and He, Tianxing and Tan, Zhaoxuan and Han, Xiaochuang and Tsvetkov, Yulia},
  journal={Advances in Neural Information Processing Systems},
  volume={36},
  pages={30840--30861},
  year={2023}
}

@inproceedings{wu2025grapheval36k,
  title={GraphEval36K: Benchmarking coding and reasoning capabilities of large language models on graph datasets},
  author={Wu, Qiming and Chen, Zichen and Corcoran, Will and Sra, Misha and Singh, Ambuj},
  booktitle={Findings of the Association for Computational Linguistics: NAACL 2025},
  pages={8095--8117},
  year={2025}
}

@inproceedings{shen2025assessing,
  title={Assessing visually-continuous corruption robustness of neural networks relative to human performance},
  author={Shen, Huakun and Hu, Boyue Caroline and Czarnecki, Krzysztof and Marsso, Lina and Chechik, Marsha},
  booktitle={2025 IEEE/CVF Winter Conference on Applications of Computer Vision (WACV)},
  pages={6300--6310},
  year={2025},
  organization={IEEE}
}

@inproceedings{anderson2018bottom,
  title={Bottom-up and top-down attention for image captioning and visual question answering},
  author={Anderson, Peter and He, Xiaodong and Buehler, Chris and Teney, Damien and Johnson, Mark and Gould, Stephen and Zhang, Lei},
  booktitle={Proceedings of the IEEE conference on computer vision and pattern recognition},
  pages={6077--6086},
  year={2018}
}

@article{liu2023visual,
  title={Visual spatial reasoning},
  author={Liu, Fangyu and Emerson, Guy and Collier, Nigel},
  journal={Transactions of the Association for Computational Linguistics},
  volume={11},
  pages={635--651},
  year={2023},
  publisher={MIT Press One Broadway, 12th Floor, Cambridge, Massachusetts 02142, USA~…}
}

@article{team2026kimi,
  title={Kimi K2. 5: Visual Agentic Intelligence},
  author={Team, Kimi and Bai, Tongtong and Bai, Yifan and Bao, Yiping and Cai, SH and Cao, Yuan and Charles, Y and Che, HS and Chen, Cheng and Chen, Guanduo and others},
  journal={arXiv preprint arXiv:2602.02276},
  year={2026}
}

@article{zeng2025glm,
  title={Glm-4.5: Agentic, reasoning, and coding (arc) foundation models},
  author={Zeng, Aohan and Lv, Xin and Zheng, Qinkai and Hou, Zhenyu and Chen, Bin and Xie, Chengxing and Wang, Cunxiang and Yin, Da and Zeng, Hao and Zhang, Jiajie and others},
  journal={arXiv preprint arXiv:2508.06471},
  year={2025}
}

@misc{abdin2024phi3,
      title={Phi-3 Technical Report: A Highly Capable Language Model Locally on Your Phone}, 
      author={Marah Abdin and Jyoti Aneja and Hany Awadalla and Ahmed Awadallah and Ammar Ahmad Awan and Nguyen Bach and Amit Bahree and Arash Bakhtiari and Jianmin Bao and Harkirat Behl and Alon Benhaim and Misha Bilenko and Johan Bjorck and Sébastien Bubeck and Martin Cai and Qin Cai and Vishrav Chaudhary and Dong Chen and Dongdong Chen and Weizhu Chen and Yen-Chun Chen and Yi-Ling Chen and Hao Cheng and Parul Chopra and Xiyang Dai and Matthew Dixon and Ronen Eldan and Victor Fragoso and Jianfeng Gao and Mei Gao and Min Gao and Amit Garg and Allie Del Giorno and Abhishek Goswami and Suriya Gunasekar and Emman Haider and Junheng Hao and Russell J. Hewett and Wenxiang Hu and Jamie Huynh and Dan Iter and Sam Ade Jacobs and Mojan Javaheripi and Xin Jin and Nikos Karampatziakis and Piero Kauffmann and Mahoud Khademi and Dongwoo Kim and Young Jin Kim and Lev Kurilenko and James R. Lee and Yin Tat Lee and Yuanzhi Li and Yunsheng Li and Chen Liang and Lars Liden and Xihui Lin and Zeqi Lin and Ce Liu and Liyuan Liu and Mengchen Liu and Weishung Liu and Xiaodong Liu and Chong Luo and Piyush Madan and Ali Mahmoudzadeh and David Majercak and Matt Mazzola and Caio César Teodoro Mendes and Arindam Mitra and Hardik Modi and Anh Nguyen and Brandon Norick and Barun Patra and Daniel Perez-Becker and Thomas Portet and Reid Pryzant and Heyang Qin and Marko Radmilac and Liliang Ren and Gustavo de Rosa and Corby Rosset and Sambudha Roy and Olatunji Ruwase and Olli Saarikivi and Amin Saied and Adil Salim and Michael Santacroce and Shital Shah and Ning Shang and Hiteshi Sharma and Yelong Shen and Swadheen Shukla and Xia Song and Masahiro Tanaka and Andrea Tupini and Praneetha Vaddamanu and Chunyu Wang and Guanhua Wang and Lijuan Wang and Shuohang Wang and Xin Wang and Yu Wang and Rachel Ward and Wen Wen and Philipp Witte and Haiping Wu and Xiaoxia Wu and Michael Wyatt and Bin Xiao and Can Xu and Jiahang Xu and Weijian Xu and Jilong Xue and Sonali Yadav and Fan Yang and Jianwei Yang and Yifan Yang and Ziyi Yang and Donghan Yu and Lu Yuan and Chenruidong Zhang and Cyril Zhang and Jianwen Zhang and Li Lyna Zhang and Yi Zhang and Yue Zhang and Yunan Zhang and Xiren Zhou},
      year={2024},
      eprint={2404.14219},
      archivePrefix={arXiv},
      primaryClass={cs.CL},
      url={https://arxiv.org/abs/2404.14219}, 
}

@article{wang2024qwen2,
  title={Qwen2-vl: Enhancing vision-language model's perception of the world at any resolution},
  author={Wang, Peng and Bai, Shuai and Tan, Sinan and Wang, Shijie and Fan, Zhihao and Bai, Jinze and Chen, Keqin and Liu, Xuejing and Wang, Jialin and Ge, Wenbin and others},
  journal={arXiv preprint arXiv:2409.12191},
  year={2024}
}

@article{chen2024expanding,
  title={Expanding performance boundaries of open-source multimodal models with model, data, and test-time scaling},
  author={Chen, Zhe and Wang, Weiyun and Cao, Yue and Liu, Yangzhou and Gao, Zhangwei and Cui, Erfei and Zhu, Jinguo and Ye, Shenglong and Tian, Hao and Liu, Zhaoyang and others},
  journal={arXiv preprint arXiv:2412.05271},
  year={2024}
}

@article{bai2025qwen3,
  title={Qwen3-vl technical report},
  author={Bai, Shuai and Cai, Yuxuan and Chen, Ruizhe and Chen, Keqin and Chen, Xionghui and Cheng, Zesen and Deng, Lianghao and Ding, Wei and Gao, Chang and Ge, Chunjiang and others},
  journal={arXiv preprint arXiv:2511.21631},
  year={2025}
}

@article{team2023gemini,
  title={Gemini: a family of highly capable multimodal models},
  author={Team, Gemini and Anil, Rohan and Borgeaud, Sebastian and Alayrac, Jean-Baptiste and Yu, Jiahui and Soricut, Radu and Schalkwyk, Johan and Dai, Andrew M and Hauth, Anja and Millican, Katie and others},
  journal={arXiv preprint arXiv:2312.11805},
  year={2023}
}

@article{achiam2023gpt,
  title={Gpt-4 technical report},
  author={Achiam, Josh and Adler, Steven and Agarwal, Sandhini and Ahmad, Lama and Akkaya, Ilge and Aleman, Florencia Leoni and Almeida, Diogo and Altenschmidt, Janko and Altman, Sam and Anadkat, Shyamal and others},
  journal={arXiv preprint arXiv:2303.08774},
  year={2023}
}

@article{chen2022program,
  title={Program of thoughts prompting: Disentangling computation from reasoning for numerical reasoning tasks},
  author={Chen, Wenhu and Ma, Xueguang and Wang, Xinyi and Cohen, William W},
  journal={arXiv preprint arXiv:2211.12588},
  year={2022}
}

@article{wang2024muirbench,
  title={Muirbench: A comprehensive benchmark for robust multi-image understanding},
  author={Wang, Fei and Fu, Xingyu and Huang, James Y and Li, Zekun and Liu, Qin and Liu, Xiaogeng and Ma, Mingyu Derek and Xu, Nan and Zhou, Wenxuan and Zhang, Kai and others},
  journal={arXiv preprint arXiv:2406.09411},
  year={2024}
}

@article{li2025mits,
  title={MITS: Enhanced Tree Search Reasoning for LLMs via Pointwise Mutual Information},
  author={Li, Jiaxi and Shi, Yucheng and Lu, Jin and Liu, Ninghao},
  journal={arXiv preprint arXiv:2510.03632},
  year={2025}
}

@misc{qwen3.5,
    title  = {{Qwen3.5}: Towards Native Multimodal Agents},
    author = {{Qwen Team}},
    month  = {February},
    year   = {2026},
    url    = {https://qwen.ai/blog?id=qwen3.5}
}

@article{liu2025mitigating,
  title={Mitigating hallucination through theory-consistent symmetric multimodal preference optimization},
  author={Liu, Wenqi and Song, Xuemeng and Li, Jiaxi and Wei, Yinwei and Zheng, Na and Yin, Jianhua and Nie, Liqiang},
  journal={arXiv preprint arXiv:2506.11712},
  year={2025}
}

@article{zhu2025internvl3,
  title={Internvl3: Exploring advanced training and test-time recipes for open-source multimodal models},
  author={Zhu, Jinguo and Wang, Weiyun and Chen, Zhe and Liu, Zhaoyang and Ye, Shenglong and Gu, Lixin and Tian, Hao and Duan, Yuchen and Su, Weijie and Shao, Jie and others},
  journal={arXiv preprint arXiv:2504.10479},
  year={2025}
}

@article{wang2025internvl3,
title={InternVL3. 5: Advancing Open-Source Multimodal Models in Versatility, Reasoning, and Efficiency},
author={Wang, Weiyun and Gao, Zhangwei and Gu, Lixin and Pu, Hengjun and Cui, Long and Wei, Xingguang and Liu, Zhaoyang and Jing, Linglin and Ye, Shenglong and Shao, Jie and others},
journal={arXiv preprint arXiv:2508.18265},
year={2025}
}

@misc{deepseekai2026deepseekv4,
      title={DeepSeek-V4: Towards Highly Efficient Million-Token Context Intelligence},
      author={DeepSeek-AI},
      year={2026},
}

@article{DBLP:qwen2.5,
  author       = {An Yang and
                  Baosong Yang and
                  Beichen Zhang and
                  Binyuan Hui and
                  Bo Zheng and
                  Bowen Yu and
                  Chengyuan Li and
                  Dayiheng Liu and
                  Fei Huang and
                  Haoran Wei and
                  Huan Lin and
                  Jian Yang and
                  Jianhong Tu and
                  Jianwei Zhang and
                  Jianxin Yang and
                  Jiaxi Yang and
                  Jingren Zhou and
                  Junyang Lin and
                  Kai Dang and
                  Keming Lu and
                  Keqin Bao and
                  Kexin Yang and
                  Le Yu and
                  Mei Li and
                  Mingfeng Xue and
                  Pei Zhang and
                  Qin Zhu and
                  Rui Men and
                  Runji Lin and
                  Tianhao Li and
                  Tingyu Xia and
                  Xingzhang Ren and
                  Xuancheng Ren and
                  Yang Fan and
                  Yang Su and
                  Yichang Zhang and
                  Yu Wan and
                  Yuqiong Liu and
                  Zeyu Cui and
                  Zhenru Zhang and
                  Zihan Qiu},
  title        = {Qwen2.5 Technical Report},
  journal      = {CoRR},
  volume       = {abs/2412.15115},
  year         = {2024},
  url          = {https://doi.org/10.48550/arXiv.2412.15115},
  doi          = {10.48550/ARXIV.2412.15115},
  eprinttype   = {arXiv},
  eprint       = {2412.15115},
  bibsource    = {dblp computer science bibliography, https://dblp.org}
}

@misc{qwq32b,
    title = {QwQ-32B: Embracing the Power of Reinforcement Learning},
    url = {https://qwenlm.github.io/blog/qwq-32b/},
    author = {Qwen Team},
    month = {March},
    year = {2025}
}

@inproceedings{sun2025graphicl,
  title={Graphicl: Unlocking graph learning potential in llms through structured prompt design},
  author={Sun, Yuanfu and Ma, Zhengnan and Fang, Yi and Ma, Jing and Tan, Qiaoyu},
  booktitle={Findings of the Association for Computational Linguistics: NAACL 2025},
  pages={2440--2459},
  year={2025}
}

@inproceedings{sun2026agentgl,
  title={Agentgl: Towards agentic graph learning with llms via reinforcement learning},
  author={Sun, Yuanfu and Li, Kang and Fan, Dongzhe and Liu, Jiajin and Tan, Qiaoyu},
  booktitle={Proceedings of the 64th Annual Meeting of the Association for Computational Linguistics (Volume 1: Long Papers)},
  pages={25313--25335},
  year={2026}
}

@article{sun2026mario,
  title={Mario: Multimodal graph reasoning with large language models},
  author={Sun, Yuanfu and Li, Kang and Guo, Pengkang and Liu, Jiajin and Tan, Qiaoyu},
  journal={arXiv preprint arXiv:2603.05181},
  year={2026}
}

@inproceedings{zhang2025trustglm,
  title={Trustglm: Evaluating the robustness of graphllms against prompt, text, and structure attacks},
  author={Zhang, Qihai and Sheng, Xinyue and Sun, Yuanfu and Tan, Qiaoyu},
  booktitle={Proceedings of the 31st ACM SIGKDD Conference on Knowledge Discovery and Data Mining V. 2},
  pages={5912--5923},
  year={2025}
}

@article{yang2026one,
  title={One Model, Many Graphs: Learning over Attributed Graphs across Heterogeneous Modalities with Vision-Language Models},
  author={Yang, Jiayi and Chen, Yifang and Sun, Yuanfu and Liu, Jiajin and Tan, Qiaoyu},
  journal={arXiv preprint arXiv:2607.19128},
  year={2026}
}

@inproceedings{chen2025medfact,
  title={Medfact: A large-scale chinese dataset for evidence-based medical fact-checking of llm responses},
  author={Chen, Tong and Wang, Zimu and Miao, Yiyi and Luo, Haoran and Yuanfei, Sun and Wang, Wei and Jiang, Zhengyong and Sen, Procheta and Su, Jionglong},
  booktitle={Proceedings of the 2025 Conference on Empirical Methods in Natural Language Processing},
  pages={32328--32341},
  year={2025}
}

@misc{fan2026mlaga,
      title={MLaGA: Multimodal Large Language and Graph Assistant}, 
      author={Dongzhe Fan and Yi Fang and Jiajin Liu and Djellel Difallah and Qiaoyu Tan},
      year={2026},
      eprint={2506.02568},
      archivePrefix={arXiv},
      primaryClass={cs.AI},
      url={https://arxiv.org/abs/2506.02568}, 
}

@misc{liu2026graphvlm,
      title={GraphVLM: Benchmarking Vision Language Models for Multimodal Graph Learning}, 
      author={Jiajin Liu and Dongzhe Fan and Chuanhao Ji and Daochen Zha and Qiaoyu Tan},
      year={2026},
      eprint={2603.13370},
      archivePrefix={arXiv},
      primaryClass={cs.CV},
      url={https://arxiv.org/abs/2603.13370}, 
}

\appendix
\clearpage
\part{Appendix}
\parttoc
\section{Dataset Details}
\label{sec:b}
\subsection{Statistics}
 In this section, we provide detailed statistics of GraphVerse. As shown in Table~\ref{tab:graphverse_stats}, we report the number of instances for each task under each graph-centric image editing (GIE) operation, offering a comprehensive view of the benchmark composition and data distribution. It can be seen that, during dataset construction, we carefully ensured that the data generated by different GIE operations are distributed as evenly as possible within each task, rather than being dominated by any single operation. At the same time, some task–operation pairs contain no data. These are highly exceptional cases where the corresponding GIE operation is simply not suitable for that task.

\subsection{Special Case}
For example, the coloring task does not include data generated by the Recolor operation. This is because constructing Recolor examples requires assigning initial colors to nodes in the original graph, whereas for graph coloring, such a setting with predefined initial colors is not a common or natural formulation of the problem. We therefore discard this setting.
Similarly, for the TSP task, we do not include the other three operations. This is because solving TSP requires starting from a complete graph and explicitly representing the pairwise distances between nodes in the image. Under such a setup, after applying Patch, Composition, or Attn operations, we would still need to ensure that the resulting graph remains complete. However, since all distance information is already provided in the image at the outset, these operations would allow models to exploit shortcuts: instead of reasoning over graph connectivity, a model could simply identify the nodes and read distances directly from the distance table to make predictions. This would deviate from our goal of emphasizing genuine visual reasoning. We therefore exclude these three operations. In contrast, the Recolor operation remains feasible for TSP, since one can determine whether correct visual reasoning has been performed by examining the colors of the nodes along the corresponding Hamiltonian tour.

\subsection{Testmini}
We note in the main text, for efficiency, we construct TestMini by sampling from the full test set during evaluation. This sampling procedure does not substantially alter the original data distribution. As shown in Table~\ref{tab:graphverse_stats_mini}, the task–GIE pairs remain evenly distributed, thereby avoiding biases caused by highly uneven or skewed distributions.

\begin{table}[h]
\centering
\small
\setlength{\tabcolsep}{4pt}
\renewcommand{\arraystretch}{1.2}
\setlength{\arrayrulewidth}{0.3mm}
\begin{tabular}{lcccc|c}
\hline
\rowcolor{CadetBlue!20}
\textbf{Task} 
& \textbf{Recolor} 
& \textbf{Patch} 
& \textbf{Composition} 
& \textbf{Attn} 
& \textbf{Total} \\
\hline
Cycle         & 250  & 250  & 250  & 250  & 1000 \\
Diameter      & 250  & 250  & 250  & 250  & 1000 \\
Distance      & 250  & 250  & 250  & 250  & 1000 \\
GED           & 250  & 250  & 250  & 250  & 1000 \\
Coloring      & --   & 333  & 333  & 334  & 1000 \\
MCP           & 250  & 250  & 250  & 250  & 1000 \\
MCS           & 250  & 250  & 250  & 250  & 1000 \\
MIS           & 250  & 250  & 250  & 250  & 1000 \\
MVC           & 250  & 250  & 250  & 250  & 1000 \\
Neighbor      & 250  & 250  & 250  & 250  & 1000 \\
TSP           & 1000 & --   & --   & --   & 1000 \\
\hline
\rowcolor{CadetBlue!10}
\textbf{Total} & \textbf{3250} & \textbf{2583} & \textbf{2583} & \textbf{2584} & \textbf{11000} \\
\hline
\end{tabular}
\caption{Statistics of GraphVerse.}
\label{tab:graphverse_stats}
\end{table}

\begin{table}[h]
\centering
\small
\setlength{\tabcolsep}{4pt}
\renewcommand{\arraystretch}{1.2}
\setlength{\arrayrulewidth}{0.3mm}
\begin{tabular}{lcccc|c}
\hline
\rowcolor{CadetBlue!20}
\textbf{Task} 
& \textbf{Recolor} 
& \textbf{Patch} 
& \textbf{Composition} 
& \textbf{Attn} 
& \textbf{Total} \\
\hline
Cycle         & 25    & 25    & 25    & 25    & 100 \\
Diameter      & 25    & 25    & 25    & 25    & 100 \\
Distance      & 25    & 25    & 25    & 25    & 100 \\
GED           & 25    & 25    & 25    & 25    & 100 \\
Coloring      & --    & 27    & 27    & 26    & 80 \\
MCP           & 25    & 25    & 25    & 25    & 100 \\
MCS           & 25    & 25    & 25    & 25    & 100 \\
MIS           & 25    & 25    & 25    & 25    & 100 \\
MVC           & 25    & 25    & 25    & 25    & 100 \\
Neighbor      & 25    & 25    & 25    & 25    & 100 \\
TSP           & 80    & --    & --    & --    & 80 \\
\hline
\rowcolor{CadetBlue!10}
\textbf{Total} & \textbf{305} & \textbf{252} & \textbf{252} & \textbf{251} & \textbf{1060} \\
\hline
\end{tabular}
\caption{Statistics of GraphVerse testmini.}
\label{tab:graphverse_stats_mini}
\end{table}

\subsection{Graph Sampling}
The node range used for graph sampling is determined based on the intrinsic characteristics of each task. In addition, these ranges are chosen with reference to commonly used settings in existing text-based graph reasoning benchmarks \cite{TangZLCL25}, ensuring that the sampled graphs remain both computationally tractable and comparable to prior evaluations.

\begin{table}[ht]
\centering
\small
\setlength{\tabcolsep}{6pt}
\renewcommand{\arraystretch}{1.15}
\setlength{\arrayrulewidth}{0.3mm}
\begin{tabular}{lcc}
\hline
\rowcolor{CadetBlue!20}
\textbf{Task} & \textbf{Min Nodes} & \textbf{Max Nodes} \\
\hline
TSP       & 7  & 9  \\
Cycle     & 6  & 10 \\
Coloring  & 8  & 15 \\
Neighbor  & 10 & 16 \\
Distance  & 8  & 15 \\
Diameter  & 8  & 15 \\
MIS       & 8  & 15 \\
MVC       & 8  & 15 \\
MCP       & 8  & 15 \\
GED       & 4  & 8  \\
MCS       & 8  & 15 \\
\hline
\end{tabular}
\caption{Node range for graph sampling.}
\label{tab:easy_node_range}
\end{table}

\begin{algorithm*}[t]
\caption{Graph Attentional Focusing}
\label{alg:attentional_focusing}
\begin{algorithmic}[1]
\Require Graph $G=(V,E)$, color set $\mathcal{C}$, target color $c^\star$
\Ensure Edited image $I'$, target subgraph $M=(V_M,E_M)$, edit metadata $\mathcal{M}_{\text{focus}}$
\State Sample a connected subgraph $M=(V_M,E_M)$ from $G$
\State Initialize a color assignment $col:V\rightarrow \mathcal{C}$
\ForAll{$v\in V_M$}
    \State $col(v)\gets c^\star$
\EndFor
\ForAll{$u\in V\setminus V_M$ such that $\exists v\in V_M,\ (u,v)\in E$}
    \State Sample $col(u)\in \mathcal{C}\setminus\{c^\star\}$
\EndFor
\ForAll{remaining nodes $u\in V\setminus V_M$ without assigned color}
    \State Sample $col(u)\in \mathcal{C}$
\EndFor
\State Let
\[
G[c^\star] = G\big[\{v\in V\mid col(v)=c^\star\}\big]
\]
be the induced subgraph on all nodes colored $c^\star$
\State Compute the connected components $\mathcal{CC}(G[c^\star])$
\State Re-sample non-target colors if necessary until
\[
M = \arg\max_{C\in\mathcal{CC}(G[c^\star])} |V(C)|
\]
\State Render the recolored graph image:
\[
I'=Render(G;col)
\]
\State $\mathcal{M}_{\text{focus}} \gets \{c^\star,\; V_M,\; E_M\}$
\State \Return $I', M, \mathcal{M}_{\text{focus}}$
\end{algorithmic}
\end{algorithm*}

\section{Further Analysis of Related Work}
\label{sec:c}

\paragraph{Multi-graph content does not imply multi-image reasoning.}
A representative line of prior work, such as Visual Graph Arena \cite{babaiee2025visual}, may appear to involve multiple graphs or richer cross-graph structural information. However, this should not be conflated with the paired-image setting studied in GraphVerse. In those benchmarks, multiple graph components are typically presented within a single visual canvas, so the model still operates over one jointly rendered image. This setting is closer in spirit to our \textit{Cross-Graph Composition} strategy, where distinct graph structures are composed into one image and reasoning remains grounded in a unified visual field. By contrast, our paired-image setting requires the model to reason across two separate images, each with its own spatial layout, visual grouping, and local context. Paired-image reasoning introduces an additional layer of visual context integration and requires more sophisticated reasoning than single-image settings \cite{wang2024muirbench} that cannot be reduced to simply placing multiple graphs in one figure. Therefore, although some existing visual graph reasoning benchmarks include multi-graph content, they do not explicitly evaluate genuine multi-image visual reasoning, which is a key capability targeted by GraphVerse. 

\section{Additional Method Details}
\label{sec:d}
\subsection{Algorithmic Workflow}
In this section, we provide the detailed algorithmic descriptions of the four graph-centric image editing (GIE) operations, namely Image-Graph Patch Perturbation (Algorithm~\ref{alg:patch_perturbation}), Cross-Graph Composition (Algorithm~\ref{alg:cross_graph_composition}), Graph Attentional Focusing (Algorithm~\ref{alg:attentional_focusing}), and Spatially-Conditioned Recoloring (Algorithm~\ref{alg:spatial_recoloring}), together with the evaluation procedure for computing VGR-Score (Algorithm~\ref{alg:vgr_score}).

\begin{algorithm*}[h]
\caption{Image-Graph Patch Perturbation}
\label{alg:patch_perturbation}
\begin{algorithmic}[1]
\Require Graph image $I$, grid size $r$, perturbation type $\tau \in \{\textsc{Flip}, \textsc{Swap}\}$
\Ensure Edited image $I'$, edit metadata $\mathcal{M}_{\text{patch}}$
\State Partition image into patches: $\Pi_r(I)=\{P_{ij}\}_{i,j=1}^{r}$
\State Construct the non-empty patch set
We define the set of non-empty patches as
\[
\Omega = \{(i,j)\mid P_{ij}\neq \varnothing\},
\]
where $P_{ij}\neq \varnothing$ indicates that patch $P_{ij}$ contains at least one graph-relevant visual element.
\If{$\tau=\textsc{Flip}$}
    \State Sample one patch $(i,j)\sim \mathrm{Unif}(\Omega)$
    \State $P_{ij}' \gets Flip(P_{ij})$
    \State $I' \gets I$ with $P_{ij}$ replaced by $P_{ij}'$
    \State $\mathcal{M}_{\text{patch}} \gets \{(\texttt{flip},(i,j))\}$
\ElsIf{$\tau=\textsc{Swap}$}
    \State Sample two distinct patches $(i,j),(k,\ell)\sim \mathrm{Unif}(\Omega)$
    \State Swap patches: $(P_{ij}',P_{k\ell}') \gets (P_{k\ell},P_{ij})$
    \State $I' \gets I$ with $P_{ij},P_{k\ell}$ replaced by $P_{ij}',P_{k\ell}'$
    \State $\mathcal{M}_{\text{patch}} \gets \{(\texttt{swap},(i,j),(k,\ell))\}$
\EndIf
\State \Return $I', \mathcal{M}_{\text{patch}}$
\end{algorithmic}
\end{algorithm*}

\begin{algorithm*}[ht]
\caption{Cross-Graph Composition}
\label{alg:cross_graph_composition}
\begin{algorithmic}[1]
\Require Source graph $G=(V,E)$
\Ensure Paired images $(I_1,I_2)$, cross-graph links $E^{(\times)}$, composed graph $\widetilde{G}$
\State Partition $V$ into two disjoint subsets $V^{(1)}$ and $V^{(2)}$ such that
\[
V^{(1)} \cap V^{(2)} = \emptyset,\qquad
V^{(1)} \cup V^{(2)} = V
\]
\State Induce two subgraphs
\[
G^{(1)}=(V^{(1)},E^{(1)}),\qquad
G^{(2)}=(V^{(2)},E^{(2)})
\]
where $E^{(1)}=E\cap (V^{(1)}\times V^{(1)})$ and $E^{(2)}=E\cap (V^{(2)}\times V^{(2)})$
\State Render two views:
\[
I_1=Render(G^{(1)}),\qquad I_2=Render(G^{(2)})
\]
\State Sample or instantiate a set of cross-graph links
\[
E^{(\times)} \subseteq V^{(1)}\times V^{(2)}
\]
\State Form the composed graph
\[
\widetilde{G} = \big(V^{(1)}\cup V^{(2)},\; E^{(1)}\cup E^{(2)}\cup E^{(\times)}\big)
\]
\State \Return $(I_1,I_2), E^{(\times)}, \widetilde{G}$
\end{algorithmic}
\end{algorithm*}

\begin{algorithm*}[t]
\caption{Spatially-Conditioned Recoloring}
\label{alg:spatial_recoloring}
\begin{algorithmic}[1]
\Require Graph $G=(V,E)$, rendered image coordinates $\{p(v)=(x_v,y_v)\}_{v\in V}$, target color $c^\dagger$
\Ensure Edited image $I'$, recolored node set $S_\delta(a)$, edit metadata $\mathcal{M}_{\text{spatial}}$
\State Sample an anchor node $a\in V$
\State Sample a direction $\delta \in \{\textsc{Left},\textsc{Right},\textsc{Above},\textsc{Below}\}$
\If{$\delta=\textsc{Left}$}
    \State $S_\delta(a)\gets \{v\in V\mid x_v < x_a\}$
\ElsIf{$\delta=\textsc{Right}$}
    \State $S_\delta(a)\gets \{v\in V\mid x_v > x_a\}$
\ElsIf{$\delta=\textsc{Above}$}
    \State $S_\delta(a)\gets \{v\in V\mid y_v < y_a\}$
\ElsIf{$\delta=\textsc{Below}$}
    \State $S_\delta(a)\gets \{v\in V\mid y_v > y_a\}$
\EndIf
\State Initialize or inherit the original color assignment $col$
\ForAll{$v\in S_\delta(a)$}
    \State $col(v)\gets c^\dagger$
\EndFor
\State Render the recolored image:
\[
I'=Render(G;col)
\]
\State $\mathcal{M}_{\text{spatial}} \gets \{a,\delta,S_\delta(a),c^\dagger\}$
\State \Return $I', S_\delta(a), \mathcal{M}_{\text{spatial}}$
\end{algorithmic}
\end{algorithm*}

\begin{algorithm*}[t]
\caption{VGR-Score Computation}
\label{alg:vgr_score}
\begin{algorithmic}[1]
\Require Response $R$, graph $G$, answer set $\mathcal{A}^\star$, edit metadata $\mathcal{M}$, coefficient $\lambda$
\Ensure $R_{\text{step}}, R_{\text{ans}}, \mathrm{VGR\mbox{-}Score}$
\State $E^\star \gets Verbalize(G)\cup Verbalize(\mathcal{M})$
\State $\mathcal{S}=\{s_1,\dots,s_m\}\gets ExtractSteps(R)$
\For{$i=1$ to $m$}
    \State $z_i \gets JudgeLLM(s_i, E^\star)$
\EndFor
\State $R_{\text{step}} \gets \frac{1}{m}\sum_{i=1}^m z_i$ if $m>0$, else $0$
\State $\hat{y}\gets ExtractAnswer(R)$
\State $R_{\text{ans}} \gets \mathbb{I}[\hat{y}\in \mathcal{A}^\star]$
\State $\mathrm{VGR\mbox{-}Score} \gets \lambda R_{\text{step}} + (1-\lambda)R_{\text{ans}}$
\State \Return $R_{\text{step}}, R_{\text{ans}}, \mathrm{VGR\mbox{-}Score}$
\end{algorithmic}
\end{algorithm*}

\begin{table*}[b]
\centering
\footnotesize
\caption{Performance comparison under training-free and training-based settings across four task categories. \myred{} indicates absolute improvements over direct inference. The table reports VGR-S only, while Acc results are listed in the main text. For Gemini-3-Pro, we apply PoT and denote the variant as Coder. For GPT-5.2, we use Codex with PoT.}
\label{tab:app_in_domain}
\vspace{-0.2cm}

\setlength{\tabcolsep}{5pt}
\renewcommand{\arraystretch}{1.35}
\arrayrulecolor{black}
\resizebox{\linewidth}{!}{%
\begin{tabular}{c||ccccc||c||ccccc}
\hline\hline

\multicolumn{12}{c}{\textbf{Training Free}} \\
\hline

\multirow{2}{*}{\textbf{Setting}}
& \multicolumn{5}{c||}{\textbf{GPT-5.2}} 
& \multirow{2}{*}{\textbf{Setting}}
& \multicolumn{5}{c}{\textbf{Gemini-3-Pro}} \\
\cline{2-6} \cline{8-12}

& \textbf{Single-Image} & \textbf{Paired Image} & \textbf{Polynomial} & \textbf{NP-Hard} & \textbf{Average}
& & \textbf{Single-Image} & \textbf{Paired Image} & \textbf{Polynomial} & \textbf{NP-Hard} & \textbf{Average} \\
\hline\hline

\textbf{Direct Inference} 
& 56.2 & 25.4 & 68.2 & 44.0 & 50.6
& \textbf{Direct Inference} 
& 63.5 & 34.7 & 65.1 & 55.7 & 58.3 \\

\hline
\rowcolor[HTML]{D7F6FF}
\textbf{GPT-5.2-Codex} 
& 63.7 {\color{orange}$\uparrow$7.5}
& 28.8 {\color{orange}$\uparrow$3.4}
& 71.9 {\color{orange}$\uparrow$3.7}
& 51.9 {\color{orange}$\uparrow$7.9}
& 57.4 {\color{orange}$\uparrow$6.8}
& \textbf{Gemini-3-Pro-Coder}
& 67.6 {\color{orange}$\uparrow$4.1}
& 47.7 {\color{orange}$\uparrow$13.0}
& 72.2 {\color{orange}$\uparrow$7.1}
& 60.9 {\color{orange}$\uparrow$5.2}
& 64.0 {\color{orange}$\uparrow$5.7} \\
\hline\hline

\multicolumn{12}{c}{\textbf{Training-based}} \\
\hline

\multirow{2}{*}{\textbf{Setting}}
& \multicolumn{5}{c||}{\textbf{Qwen3-VL-8B-Instruct}} 
& \multirow{2}{*}{\textbf{Setting}}
& \multicolumn{5}{c}{\textbf{Qwen3-VL-2B-Instruct}} \\
\cline{2-6} \cline{8-12}

& \textbf{Single-Image} & \textbf{Paired Image} & \textbf{Polynomial} & \textbf{NP-Hard} & \textbf{Average}
& & \textbf{Single-Image} & \textbf{Paired Image} & \textbf{Polynomial} & \textbf{NP-Hard} & \textbf{Average} \\
\hline\hline

\textbf{Direct Inference}
& 30.4 & 7.5 & 41.8 & 20.4 & 26.2
& \textbf{Direct Inference}
& 11.2 & 6.7 & 17.8 & 7.6 & 10.4 \\
\hline
\textbf{SFT w/o GIE}
& 39.4 {\color{orange}$\uparrow$9.0}
& 18.1 {\color{orange}$\uparrow$10.6}
& 43.6 {\color{orange}$\uparrow$1.8}
& 32.5 {\color{orange}$\uparrow$12.1}
& 35.5 {\color{orange}$\uparrow$9.3}
& \textbf{SFT w/o GIE}
& 15.3 {\color{orange}$\uparrow$4.1}
& 7.6 {\color{orange}$\uparrow$0.9}
& 24.3 {\color{orange}$\uparrow$6.5}
& 10.0 {\color{orange}$\uparrow$2.4}
& 13.9 {\color{orange}$\uparrow$3.5} \\

\rowcolor[HTML]{D7F6FF}
\textbf{SFT}
& 45.9 {\color{orange}$\uparrow$15.5}
& 20.1 {\color{orange}$\uparrow$12.6}
& 45.5 {\color{orange}$\uparrow$3.7}
& 39.6 {\color{orange}$\uparrow$19.2}
& 41.2 {\color{orange}$\uparrow$15.0}
& \textbf{SFT}
& 17.1 {\color{orange}$\uparrow$5.9}
& 11.8 {\color{orange}$\uparrow$5.1}
& 24.5 {\color{orange}$\uparrow$6.7}
& 13.0 {\color{orange}$\uparrow$5.4}
& 16.1 {\color{orange}$\uparrow$5.7} \\

\textbf{RL w/o GIE}
& 40.7 {\color{orange}$\uparrow$10.3}
& 27.6 {\color{orange}$\uparrow$20.1}
& 51.7 {\color{orange}$\uparrow$9.9}
& 33.3 {\color{orange}$\uparrow$12.9}
& 38.3 {\color{orange}$\uparrow$12.1}
& \textbf{RL w/o GIE}
& 16.5 {\color{orange}$\uparrow$5.3}
& 9.2 {\color{orange}$\uparrow$2.5}
& 23.9 {\color{orange}$\uparrow$6.1}
& 11.9 {\color{orange}$\uparrow$4.3}
& 15.2 {\color{orange}$\uparrow$4.8} \\

\hline

\rowcolor[HTML]{D7F6FF}
\textbf{RL}
& 48.2 {\color{orange}$\uparrow$17.8}
& 35.2 {\color{orange}$\uparrow$27.7}
& 50.2 {\color{orange}$\uparrow$8.4}
& 44.2 {\color{orange}$\uparrow$23.8}
& 45.8 {\color{orange}$\uparrow$19.6}
& \textbf{RL}
& 30.2 {\color{orange}$\uparrow$19.0}
& 21.0 {\color{orange}$\uparrow$14.3}
& 29.4 {\color{orange}$\uparrow$11.6}
& 28.2 {\color{orange}$\uparrow$20.6}
& 28.5 {\color{orange}$\uparrow$18.1} \\
\hline\hline

\end{tabular}%
}
\end{table*}

\clearpage

\noindent \textbf{Paired-Image VGR.}
Each GIE operation is defined at the single-image level and is therefore applied independently to one rendered graph image at a time. To extend GIE to the paired-image setting, we adopt two complementary configurations for data construction. For each GIE type, we apply the operation to \emph{both} images in half of the paired-image instances, while in the remaining half we apply it to \emph{only one} of the two images. This design increases the structural diversity of the paired-image data and prevents the benchmark from being restricted to a single intervention pattern.

\subsection{Design Rationale Addendum}
\label{sec:design_ration}

\paragraph{Graph Layout Design.}
GraphVerse adopts rule-based structured graph layouts to ensure controllability, scalability, and verifiable evaluation. In visual graph reasoning, real-world graph images paired with natural questions and exact labels are rarely available at scale, making large-scale benchmark construction challenging. Programmatic rendering therefore provides a practical and principled solution: it enables systematic sampling of graph structures, precise control over task difficulty, and exact ground-truth generation through graph algorithms. This design also follows common practice in visual graph computation benchmarks, where structured rendering is typically used to support scalable and reproducible evaluation. During pilot studies, we compared several rendering styles and found that many alternatives became cluttered at larger graph scales, caused severe node/edge overlaps, or failed to clearly show graph attributes. We therefore use GraphViz for clear, scalable, and consistent visualization of topology and attributes.

\paragraph{Visual Robustness Beyond Perception.}
GraphVerse does not primarily focus on generic visual corruptions such as occlusion, blurring, low resolution, viewpoint changes, or hand-drawn styles. These perturbations are valuable for testing visual perception robustness, but once the graph structure is recognized, the problem can often be reduced to textual graph reasoning. Instead, our Graph-Centric Image Editing operations are designed to evaluate visually grounded reasoning beyond initial perception. Models must identify edited visual evidence and further incorporate the updated visual attributes into relational or algorithmic graph reasoning. Thus, GraphVerse emphasizes whether MLLMs can reason with visual graph changes, rather than only perceive noisy graph images.

\paragraph{Novel Paired-Image Task Design.}
For paired-image visual graph reasoning, GraphVerse includes Graph Edit Distance and Maximum Common Subgraph, two canonical cross-graph comparison tasks. These tasks require models to align nodes or substructures across two graph images, compare relational patterns, and infer graph-level correspondences. Therefore, even with a focused task set, they already provide a representative probe of cross-image graph alignment and structural reasoning. Given that multi-image reasoning is often treated as a separate direction in VQA benchmarks, GraphVerse prioritizes building a broad single-image VGR suite while using GED and MCS to diagnose paired-image reasoning ability.

\paragraph{Paired-Image Editing Design.}
For paired-image GIE, GraphVerse considers two basic settings: editing one image or editing both images. Although simple in form, these settings interact with GED and MCS to create non-trivial cross-graph reasoning challenges, since the model must compare edited visual evidence across graph instances and reason about structural differences or common subgraphs. More complex settings, such as temporal graphs, cross-graph semantic relations, or reasoning over more than two graph images, are promising extensions. We view the current paired-image setup as an initial diagnostic step and leave broader multi-image visual graph reasoning to future versions of GraphVerse.

\subsection{Verbalization Details}

To make the evaluation pipeline fully transparent, we further clarify the verbalization process used for VGR-Score. Our verbalization is fully rule-based and does not involve any LLM, which avoids additional hallucination or judge-side noise. For the original graph structure, we verbalize each edge deterministically with node names and colors, e.g., \texttt{JJU(rose)--JNN(rose)} and \texttt{JJU(rose)--JNS(lemon)}.

For each Graph-Centric Image Editing (GIE) operation, we store and verbalize the affected evidence in a deterministic format.

\paragraph{Spatially-Conditioned Recoloring.}
For spatial recoloring, we store the direction, anchor entity, recolored entities, and new color. For example:
\begin{verbatim}
{
  "type": "color_swap",
  "direction": "right",
  "anchor": "Nepenthes macfarlanei",
  "recolored_entities": [
    "Zamia chigua",
    "Berthold Carl Seemann"
  ],
  "new_color": "peach"
}
\end{verbatim}

\paragraph{Graph Attentional Focusing.}
For attentional focusing, we store the highlight color, highlighted nodes, and the induced subgraph adjacency matrix. For example:
\begin{verbatim}
{
  "highlight_color": "mint",
  "highlight_nodes": [
    "Linlithgow Palace",
    "Martin Garzez"
  ],
  "subgraph": ...
}
\end{verbatim}

\paragraph{Image-Graph Patch Perturbation.}
For patch perturbation, we store the entities inside the perturbed patches. For example:
\begin{verbatim}
{
  "patch_entities": [
    "Alchemilla vulgaris",
    "Alchemilla"
  ]
}
\end{verbatim}
Since this operation does not change the graph topology or node colors, the scoring agent uses these perturbation signals together with the original graph-structure evidence.

\paragraph{Cross-Graph Composition.}
For cross-graph composition, we store the cross-graph connectors and the composed graph edges. For example:
\begin{verbatim}
{
  "connectors": [
    "Rosids --[order]-- Fabaceae"
  ],
  "composed_graph_edges": ...
}
\end{verbatim}

In the scoring prompt, we only add simple transition sentences before introducing each type of evidence, such as ``Graph edges with node colors (undirected; list each edge once).'' This ensures that all evidence provided to the scoring agent is derived from deterministic metadata rather than LLM-generated descriptions. For gold-label collection, solver correctness is ensured by applying standard task-specific graph algorithms to the known underlying graph before visualization. For each graph computational problem, we run a corresponding Python solver to obtain the ground-truth answer.

\subsection{Subgraph Sampling}

We ensure that each question is defined only on the rendered subgraph, not on the original full graph.
For molecular graphs, we do not further sample subgraphs, since this may break chemically meaningful molecular structures. For other graph types, we sample subgraphs with moderate sizes, as reported in Table~\ref{tab:easy_node_range}, avoiding overly small graphs with insufficient structural information. We also apply preprocessing including removing isolated nodes so that the resulting subgraphs preserve enough connectivity and task-relevant evidence.

\section{Robustness Analysis of VGR-Score}
\label{sec:vgr_robustness}

To further examine the reliability of VGR-Score, we conduct additional robustness analyses from two perspectives: its agreement with human annotations and its sensitivity to different LLM judges. These studies aim to verify whether VGR-Score provides a stable and human-aligned process-level evaluation beyond final-answer accuracy.

\subsection{Human Agreement Study}

We first evaluate the consistency between VGR-Score computed by the LLM judge and manual human annotations. Specifically, we recruit two human annotators (volunteers) and select five representative tasks: Diameter, Shortest Distance (SD), Cycle, Coloring, and Graph Edit Distance (GED). For each task, we randomly sample 25 examples and ask the annotators to score the model responses following the same VGR-Score criteria. We then compare the manually computed scores with those obtained from the LLM judge.
As shown in Table~\ref{tab:human_agreement}, VGR-Score exhibits strong agreement with human annotations across different tasks and models. For Gemini-3-Pro, GPT-5.2, and Kimi-K2.5, the differences between LLM-judge scores and manual annotations are consistently small, mostly within $\pm 0.3$, with the largest deviation being only $+0.4$ on Coloring. These results indicate that the LLM judge can closely approximate human process-level evaluation, supporting the reliability of VGR-Score as a human-aligned metric.

\begin{table*}[t]
\centering
\small
\caption{Human agreement study of VGR-Score across diverse MLLMs. Values in parentheses indicate the difference from the LLM-judge score.}
\label{tab:human_agreement}
\resizebox{\textwidth}{!}{
\begin{tabular}{lccccc}
\toprule
\textbf{Model} & \textbf{Diameter} & \textbf{SD} & \textbf{Cycle} & \textbf{Coloring} & \textbf{GED} \\
\midrule
Gemini-3-Pro & 63.4 & 55.7 & 89.7 & 72.9 & 35.1 \\
Gemini-3-Pro w/ Manual Annotation 
& 63.2 \diff{(-0.2)}
& 55.6 \diff{(-0.1)}
& 89.9 \diff{(+0.2)}
& 73.3 \diff{(+0.4)}
& 35.2 \diff{(+0.1)} \\
\midrule
GPT-5.2 & 60.6 & 55.9 & 91.8 & 43.5 & 30.5 \\
GPT-5.2 w/ Manual Annotation 
& 60.5 \diff{(-0.1)}
& 56.2 \diff{(+0.3)}
& 91.7 \diff{(-0.1)}
& 43.6 \diff{(+0.1)}
& 30.7 \diff{(+0.2)} \\
\midrule
Kimi-K2.5 & 48.1 & 52.5 & 67.7 & 62.1 & 34.5 \\
Kimi-K2.5 w/ Manual Annotation 
& 47.8 \diff{(-0.3)}
& 52.6 \diff{(+0.1)}
& 67.7 \diff{(+0.0)}
& 62.4 \diff{(+0.3)}
& 34.4 \diff{(-0.1)} \\
\bottomrule
\end{tabular}
}
\end{table*}

\subsection{Judge Robustness Analysis}

We further analyze whether VGR-Score is robust to the choice of LLM judge. In addition to the default GPT-5.1 judge, we recompute VGR-Score using Gemini-3-flash as an alternative and competitive non-GPT-family judge. We also repeat the evaluation with GPT-5.1 in a separate run to examine run-to-run stability. This comparison is conducted on responses generated by Gemini-3-Pro across the same five representative tasks.
Table~\ref{tab:judge_robustness} shows that VGR-Score remains highly stable under different judging settings. Replacing GPT-5.1 with Gemini-3-flash changes the scores by only $-0.4$ to $+0.7$, while the second GPT-5.1 run introduces variations within $-0.5$ to $+0.2$. These fluctuations are minor compared with the overall score scale, suggesting that VGR-Score is not overly sensitive to either the specific LLM judge or random run-level variation. Overall, these results demonstrate that VGR-Score is a stable and judge-robust process-level metric.

\begin{table*}[t]
\centering
\small
\caption{Judge robustness and run-to-run stability analysis of VGR-Score. We recompute VGR-Score on Gemini-3-Pro responses using different judge settings. Values in parentheses indicate the difference from the default GPT-5.1 judge.}
\label{tab:judge_robustness}
\resizebox{\textwidth}{!}{
\begin{tabular}{lccccc}
\toprule
\textbf{Judge Setting} & \textbf{Diameter} & \textbf{SD} & \textbf{Cycle} & \textbf{Coloring} & \textbf{GED} \\
\midrule
GPT-5.1-as-Judge 
& 63.4 & 55.7 & 89.7 & 72.9 & 35.1 \\
Gemini-3-flash-as-Judge 
& 63.0 \diff{(-0.4)}
& 56.2 \diff{(+0.5)}
& 89.6 \diff{(-0.1)}
& 73.6 \diff{(+0.7)}
& 34.8 \diff{(-0.3)} \\
GPT-5.1-as-Judge Run 2.0 
& 63.6 \diff{(+0.2)}
& 55.3 \diff{(-0.4)}
& 89.4 \diff{(-0.3)}
& 72.4 \diff{(-0.5)}
& 34.9 \diff{(-0.2)} \\
\bottomrule
\end{tabular}
}
\end{table*}

\section{Supplementary Experiments}
\label{sec:e}

\subsection{VGR Results for Table~\ref{tab:main_result_replicate}.}
As shown in Table~\ref{tab:app_in_domain}, VGR-Score, similar to Acc, also achieves consistent improvements after applying the enhancement methods listed in the table.

\begin{table}[ht]
\centering
\caption{Further ablation of individual graph-centric image editing strategies under training-based settings on Qwen3-VL-8B-Instruct.}

\setlength{\tabcolsep}{8pt}
\renewcommand{\arraystretch}{1.35}
\arrayrulecolor{black}
\resizebox{\linewidth}{!}{%
\begin{tabular}{c||cc}
\hline\hline

\multicolumn{3}{c}{\textbf{Training-based}} \\
\hline

\multirow{2}{*}{\textbf{Setting}}
& \multicolumn{2}{c}{\textbf{Qwen3-VL-8B-Instruct}} \\
\cline{2-3}

& \textbf{Acc} & \textbf{VGR-S} \\
\hline\hline

\textbf{RL w/o GIE}
& 30.6 & 38.3 \\
\hline

\textbf{RL w Image-Graph Patch Perturbation}
& 32.8 {\color{orange}$\uparrow$7.2\%}
& 43.4 {\color{orange}$\uparrow$13.3\%} \\
\hline

\textbf{RL w Cross-Graph Composition}
& 32.6 {\color{orange}$\uparrow$6.5\%}
& 42.6 {\color{orange}$\uparrow$11.2\%} \\
\hline

\textbf{RL w Graph Attentional Focusing}
& 32.9 {\color{orange}$\uparrow$7.5\%}
& 43.9 {\color{orange}$\uparrow$14.6\%} \\
\hline

\textbf{RL w Spatially-Conditioned Recoloring}
& 33.4 {\color{orange}$\uparrow$9.2\%}
& 44.5 {\color{orange}$\uparrow$16.2\%} \\
\hline\hline

\end{tabular}%
}

\label{tab:appen_ablation}
\end{table}

\subsection{Further Ablation on Individual GIE Strategies}
To further examine whether each proposed graph-centric image editing (GIE) strategy is individually effective, we conduct an additional ablation in which Qwen3-VL-8B-Instruct is trained with RL using only one type of GIE data at a time. As shown in Table~\ref{tab:appen_ablation}, every single strategy consistently outperforms the RL w/o GIE baseline on both Accuracy and VGR-S, demonstrating that none of the proposed operations is redundant. Specifically, compared with RL w/o GIE (30.6 Acc / 38.3 VGR-S), training with a single GIE strategy improves Accuracy to 32.6--33.4 and VGR-S to 42.6--44.5. The relative gains range from 6.5\% to 9.2\% in Accuracy and from 11.2\% to 16.2\% in VGR-S. Among the four strategies, Spatially-Conditioned Recoloring achieves the largest improvement, while Graph Attentional Focusing also yields strong gains. These results suggest that each GIE strategy provides useful and complementary supervision for visual graph reasoning by strengthening the model's ability to reason over visually grounded structures, salient regions, and relational cues.
\begin{table*}[t]
\centering
\footnotesize
\caption{Performance comparison across three settings on MathVista.}
\setlength{\tabcolsep}{5pt}
\renewcommand{\arraystretch}{1.5}
\resizebox{\linewidth}{!}{%
\begin{tabular}{l|l|ccc}
\hline
\textbf{Dimension} & \textbf{Group} & \textbf{Direct Inference} & \textbf{RL w/o GIE (RL-base)} & \textbf{RL} \\
\hline\hline

\textbf{Overall} & Accuracy & 72.90 & 73.20 & \textbf{74.00} \\
\hline

\multirow{2}{*}{\textbf{Question Type}}
& Multi-choice & 79.07 & 79.26 & 80.56 \\
& Free-form    & 65.65 & 66.09 & 66.30 \\
\hline

\multirow{2}{*}{\textbf{Category}}
& General-VQA        & 73.04 & 73.48 & 74.13 \\
& Math-targeted-VQA  & 72.78 & 72.96 & 73.89 \\
\hline

\multirow{5}{*}{\textbf{Task}}
& Math word problem           & 80.11 & 84.95 & 81.18 \\
& Geometry problem solving    & 78.37 & 75.00 & 77.88 \\
& Textbook question answering & 74.68 & 73.42 & 75.32 \\
& Figure question answering   & 73.61 & 72.12 & 76.95 \\
& Visual question answering   & 56.42 & 60.34 & 56.42 \\
\hline

\multirow{4}{*}{\textbf{Grade}}
& High school       & 82.68 & 79.74 & 82.35 \\
& Elementary school & 68.16 & 72.14 & 71.14 \\
& Daily life        & 69.55 & 70.87 & 70.87 \\
& College           & 66.07 & 65.18 & 66.96 \\
\hline

\multirow{7}{*}{\textbf{Skills}}
& Statistical reasoning & 84.72 & 84.05 & 86.71 \\
& Algebraic reasoning   & 76.87 & 74.73 & 77.22 \\
& Geometry reasoning    & 77.82 & 74.48 & 76.99 \\
& Arithmetic reasoning  & 67.99 & 71.95 & 69.12 \\
& Scientific reasoning  & 73.77 & 69.67 & 72.13 \\
& Numeric commonsense   & 39.58 & 43.75 & 40.28 \\
& Logical reasoning     & 27.03 & 24.32 & 37.84 \\
\hline
\end{tabular}%
}
\label{tab:mathverse_three_runs}
\end{table*}

\subsection{Training-Based Further Ablation of Direct Visual Access}

\label{sec:ablations_training}
To further isolate the role of direct visual access, we introduce a text-only baseline, denoted as \textit{MLLM Description $\rightarrow$ LLM}. In this setting, an MLLM first converts the edited graph image into a textual description using GPT-5.2, and a text-only LLM then solves the task based only on the generated description. We evaluate this transcription-based pipeline in both in-domain evaluation and transfer evaluation on MathVista.

As shown in Table~\ref{tab:text_only_indomain}, the original visual setting consistently outperforms the MLLM Description baseline across both Qwen3-VL-8B-Instruct and Qwen3-VL-2B-Instruct backbones. For Qwen3-VL-8B-Instruct, SFT drops from 17.4/41.2 to 10.6/32.3 in Acc/VGR-S, while RL drops from 34.8/45.8 to 22.6/30.7. The degradation is even larger for Qwen3-VL-2B-Instruct, where RL decreases from 6.1/28.5 to 0.4/10.5.

The transfer results in Table~\ref{tab:text_only_transfer} show a similar trend. The original visual setting achieves an overall score of 74.00 on MathVista, outperforming the transcription-based pipeline by 5\%. These results suggest that GIE cannot be fully reduced to textual transcription. Removing direct visual access consistently hurts both in-domain reasoning and out-of-domain transfer, supporting our claim that GIE requires visually grounded graph reasoning.

\begin{table*}[t]
\centering
\small
\caption{In-domain evaluation under Qwen3-VL-8B-Instruct and Qwen3-VL-2B-Instruct backbones. Values in parentheses indicate the improvement over the MLLM Description baseline.}
\label{tab:text_only_indomain}
\resizebox{\textwidth}{!}{
\begin{tabular}{lcccc}
\toprule
\textbf{Setting} & \textbf{Acc (8B)} & \textbf{VGR-S (8B)} & \textbf{Acc (2B)} & \textbf{VGR-S (2B)} \\
\midrule
SFT (Original Setting) 
& 17.4 \diff{(+6.8)} 
& 41.2 \diff{(+8.9)} 
& 7.6 \diff{(+6.2)} 
& 16.1 \diff{(+6.9)} \\
SFT (MLLM Description Baseline) 
& 10.6 
& 32.3 
& 1.4 
& 9.2 \\
\midrule
RL (Original Setting) 
& 34.8 \diff{(+12.2)} 
& 45.8 \diff{(+15.1)} 
& 6.1 \diff{(+5.7)} 
& 28.5 \diff{(+18.0)} \\
RL (MLLM Description Baseline) 
& 22.6 
& 30.7 
& 0.4 
& 10.5 \\
\bottomrule
\end{tabular}
}
\end{table*}

\begin{table*}[t]
\centering
\small
\caption{Transfer evaluation on MathVista. Values in parentheses indicate the improvement over the MLLM Description baseline.}
\label{tab:text_only_transfer}
\resizebox{\textwidth}{!}{
\begin{tabular}{lccccc}
\toprule
\textbf{Setting} & \textbf{MC} & \textbf{Free Form} & \textbf{General} & \textbf{Math} & \textbf{Overall} \\
\midrule
RL (Original Setting) 
& 80.56 
& 66.30 
& 74.13 
& 73.89 
& 74.00 \diff{(+5.0)} \\
RL (MLLM Description $\rightarrow$ LLM) 
& 76.24 
& 60.00 
& 68.67 
& 69.28 
& 69.00 \\
\bottomrule
\end{tabular}
}
\end{table*}

\subsection{RL Training Details}
We use a lightweight rule-based reward combined with the GRPO algorithm to encourage faithful reasoning, valid formatting, and accurate final answers. The overall reward consists of three main parts: a \textit{thinking reward}, a \textit{format reward}, and a \textit{content reward}. First, the thinking reward encourages the model to produce a non-trivial reasoning process by assigning a higher score to longer thought content, with rewards increasing from 0 to 0.25 based on length, and a small 0.05 penalty applied when the reasoning becomes excessively verbose. Second, the format reward checks whether the output follows the task-specific answer format; valid formatting receives a fixed reward of 0.20, while invalid formatting leads to a partial deduction of the thinking reward and skips subsequent content evaluation. Third, the content reward serves as the core supervision signal and measures answer correctness through either exact set matching against candidate gold answers or keyword / numerical matching when only reference descriptions are available, with the main reward typically ranging from 0.30 to 0.55 depending on the matching quality. Finally, the total reward is clipped to the range of $[0,1]$ and rounded to two decimal places. This design provides a simple yet effective supervision signal that jointly promotes reasoning quality, format compliance, and answer accuracy. All training experiments are finished using 8 NVIDIA RTX PRO 6000 GPUs 96GB.

\subsection{Detailed Transfer Results on MathVista}
Table~\ref{tab:mathverse_three_runs} provides a more fine-grained view of the transfer results on MathVista. Overall, standard RL training on our data leads to consistent gains over direct inference, improving the overall accuracy from 72.90 to 74.00. More importantly, the improvement is not limited to a single subset, but is reflected across multiple dimensions, including question type, category, task, grade level, and reasoning skill. For example, RL improves both multi-choice and free-form questions, and yields gains on both General-VQA and Math-targeted-VQA. At the task level, it achieves the strongest improvements on textbook question answering and figure question answering, while maintaining competitive performance on the remaining tasks. Across grade levels, RL remains beneficial for elementary school, daily life, and college problems, and stays comparable to direct inference on high-school questions. At the skill level, the gains are especially notable for statistical reasoning and logical reasoning, with logical reasoning improving substantially from 27.03 to 37.84. These results suggest that the benefit of our RL training is broad rather than narrow: it enhances not only overall answer accuracy, but also the model's general reasoning ability across diverse mathematical and visual reasoning dimensions. This further supports that the supervision induced by our training data is transferable beyond in-domain visual graph reasoning and can improve more general multimodal reasoning performance.

\subsection{Human Reference}
The human reference results were completed by two annotators holding Ph.D. degrees in Computer Science. All instructions provided to annotators are fair, impartial, and consistent with common benchmark practices.

\subsection{AI Usage}
We used AI tools to assist with proofreading and typo correction.

\begin{figure*}[t]
    \centering
    \setlength{\tabcolsep}{3pt}
    
    {\bfseries Spatially-Conditioned Recoloring}\par\vspace{3pt}
    \begin{tabular}{ccccc}
        \subcaptionbox{5}{\includegraphics[width=0.18\textwidth]{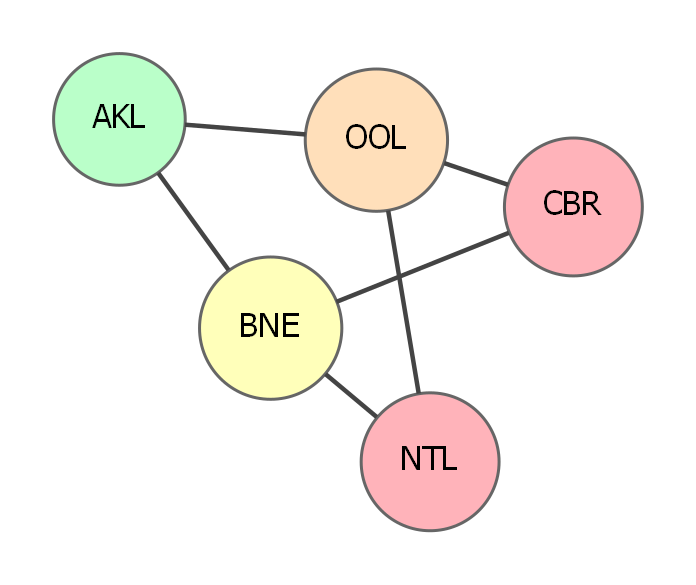}} &
        \subcaptionbox{10}{\includegraphics[width=0.18\textwidth]{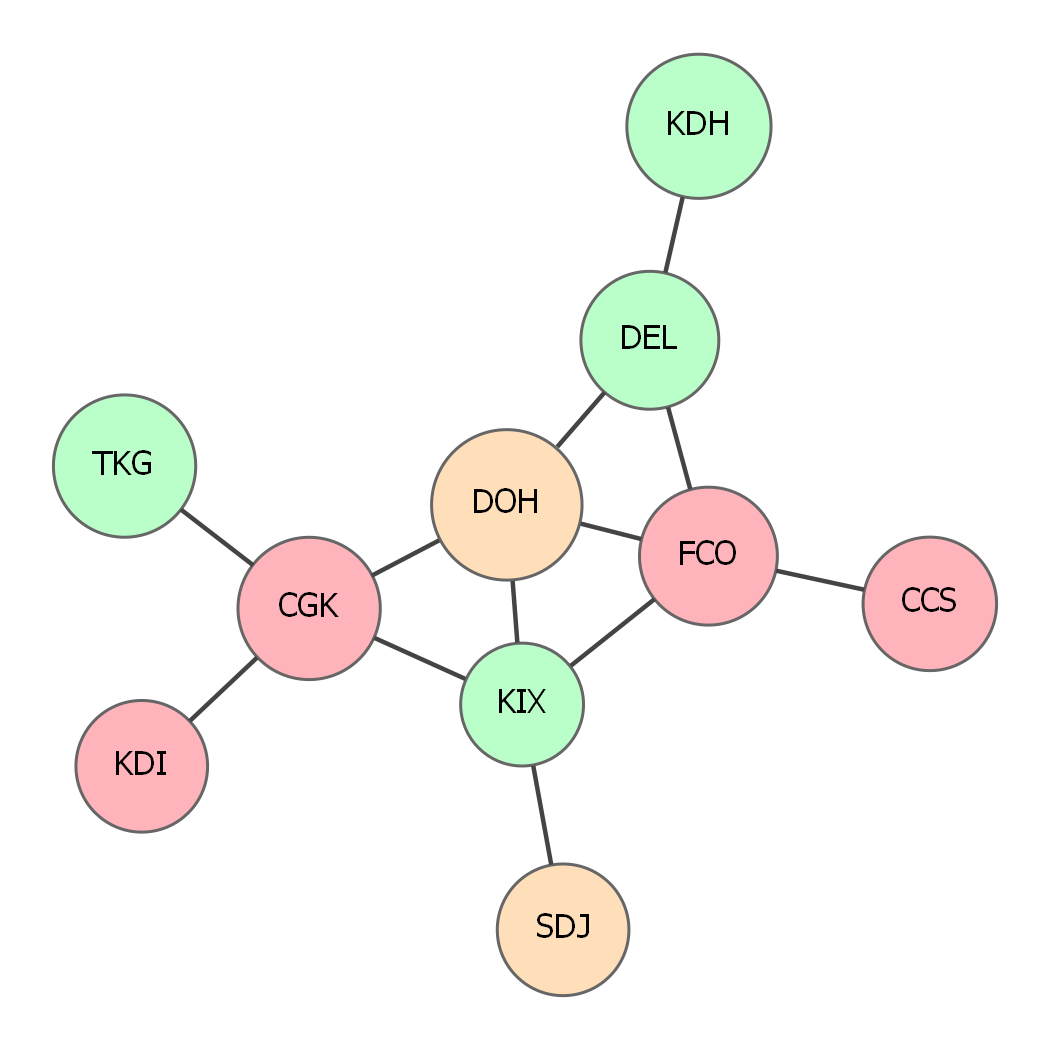}} &
        \subcaptionbox{15}{\includegraphics[width=0.18\textwidth]{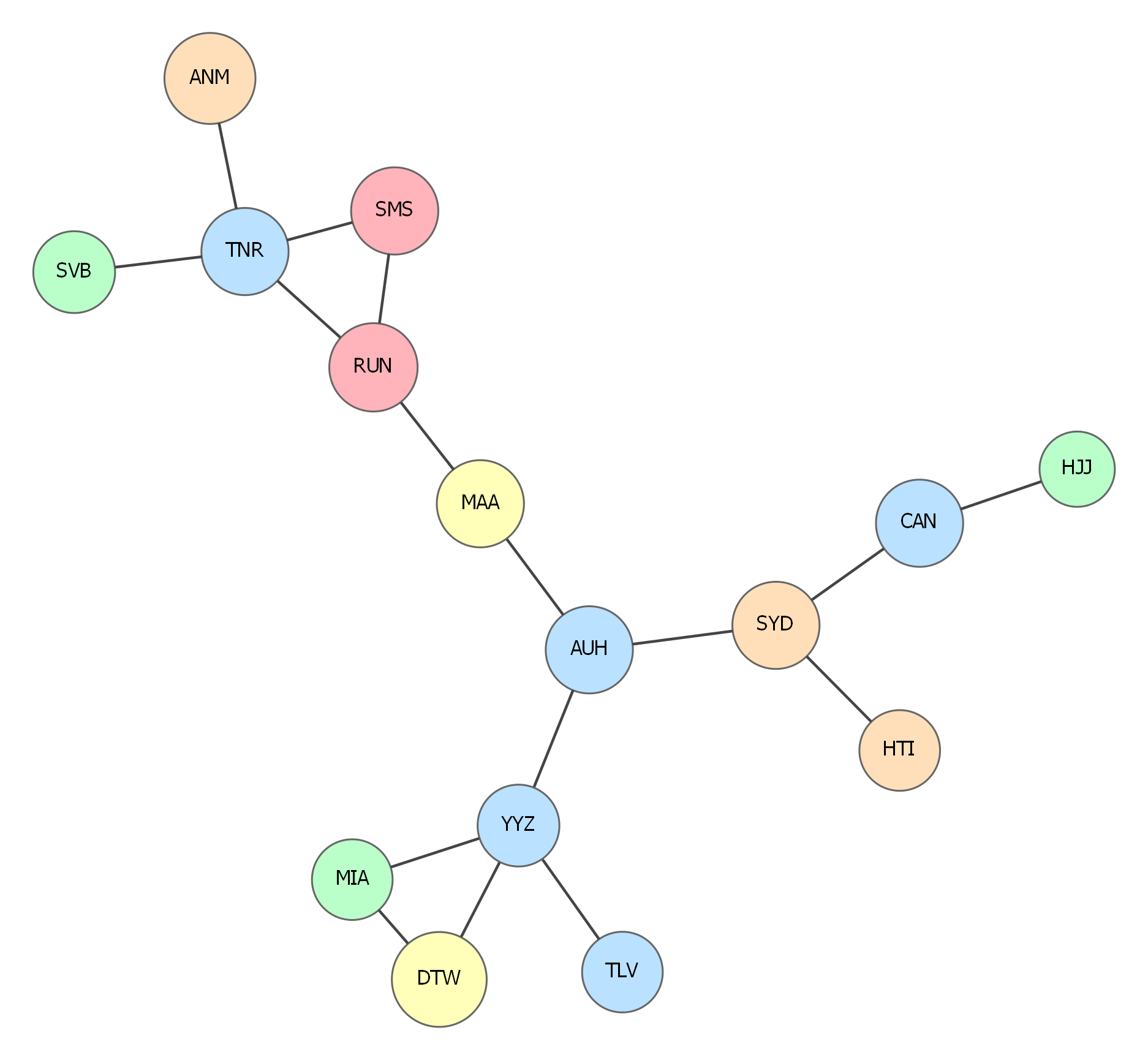}} &
        \subcaptionbox{20}{\includegraphics[width=0.18\textwidth]{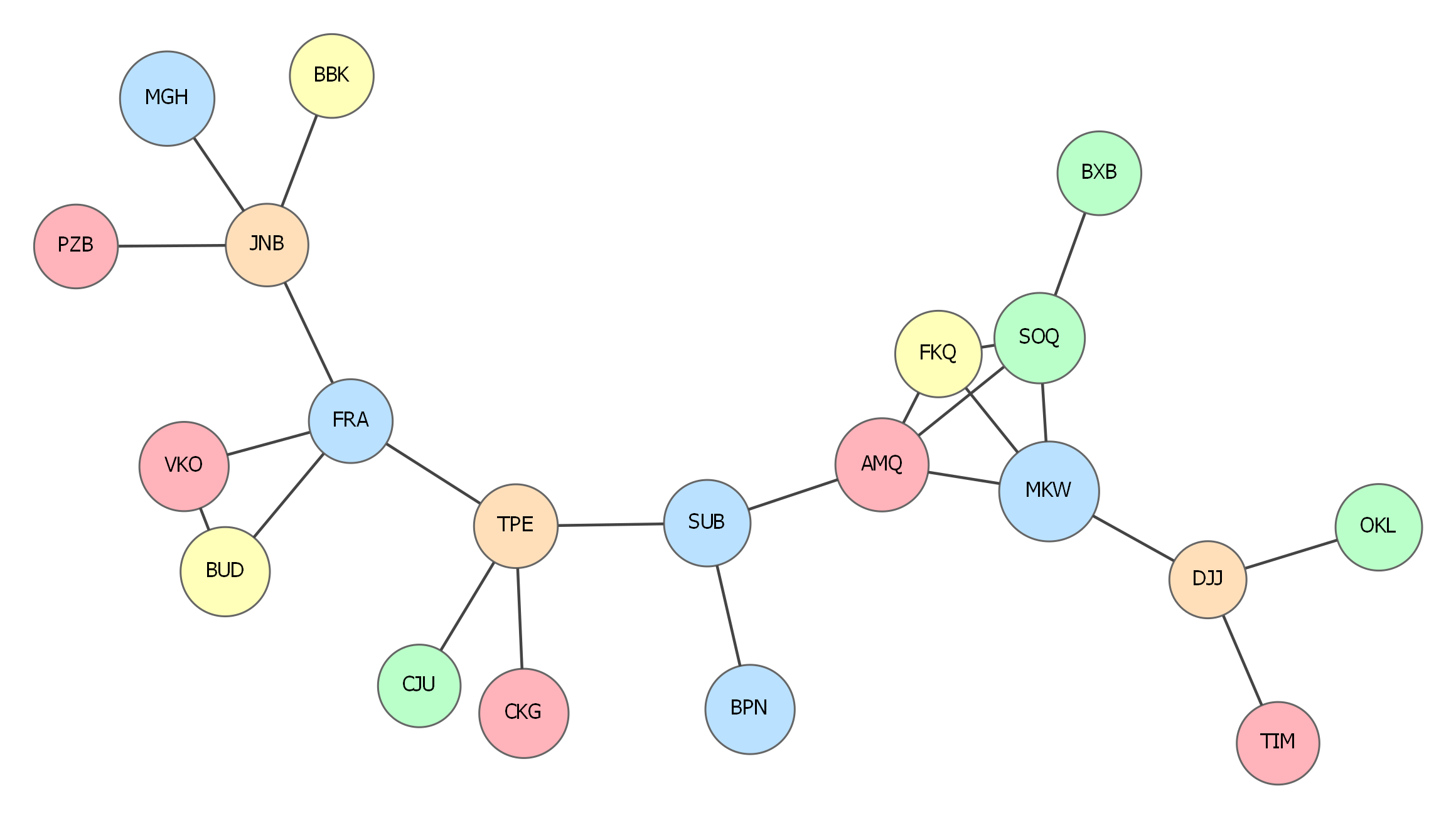}} &
        \subcaptionbox{25}{\includegraphics[width=0.18\textwidth]{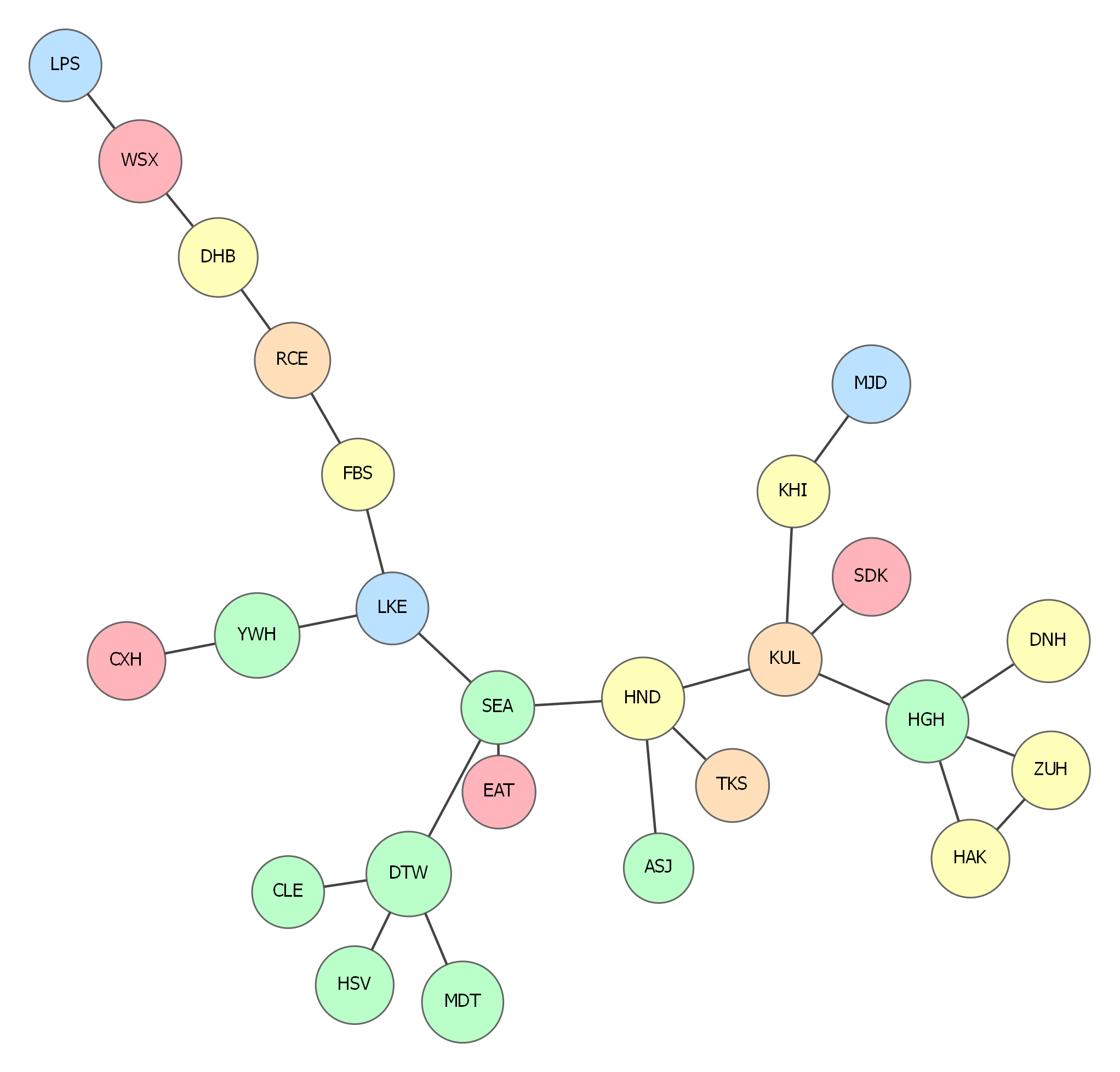}}
    \end{tabular}

    \vspace{14pt}

    {\bfseries Image-Graph Patch Perturbation}\par\vspace{3pt}
    \begin{tabular}{ccccc}
        \subcaptionbox{5}{\includegraphics[width=0.18\textwidth]{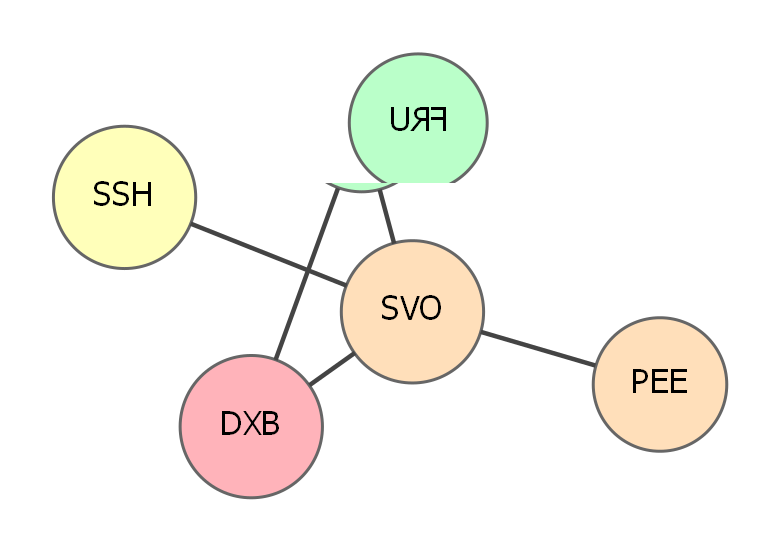}} &
        \subcaptionbox{10}{\includegraphics[width=0.18\textwidth]{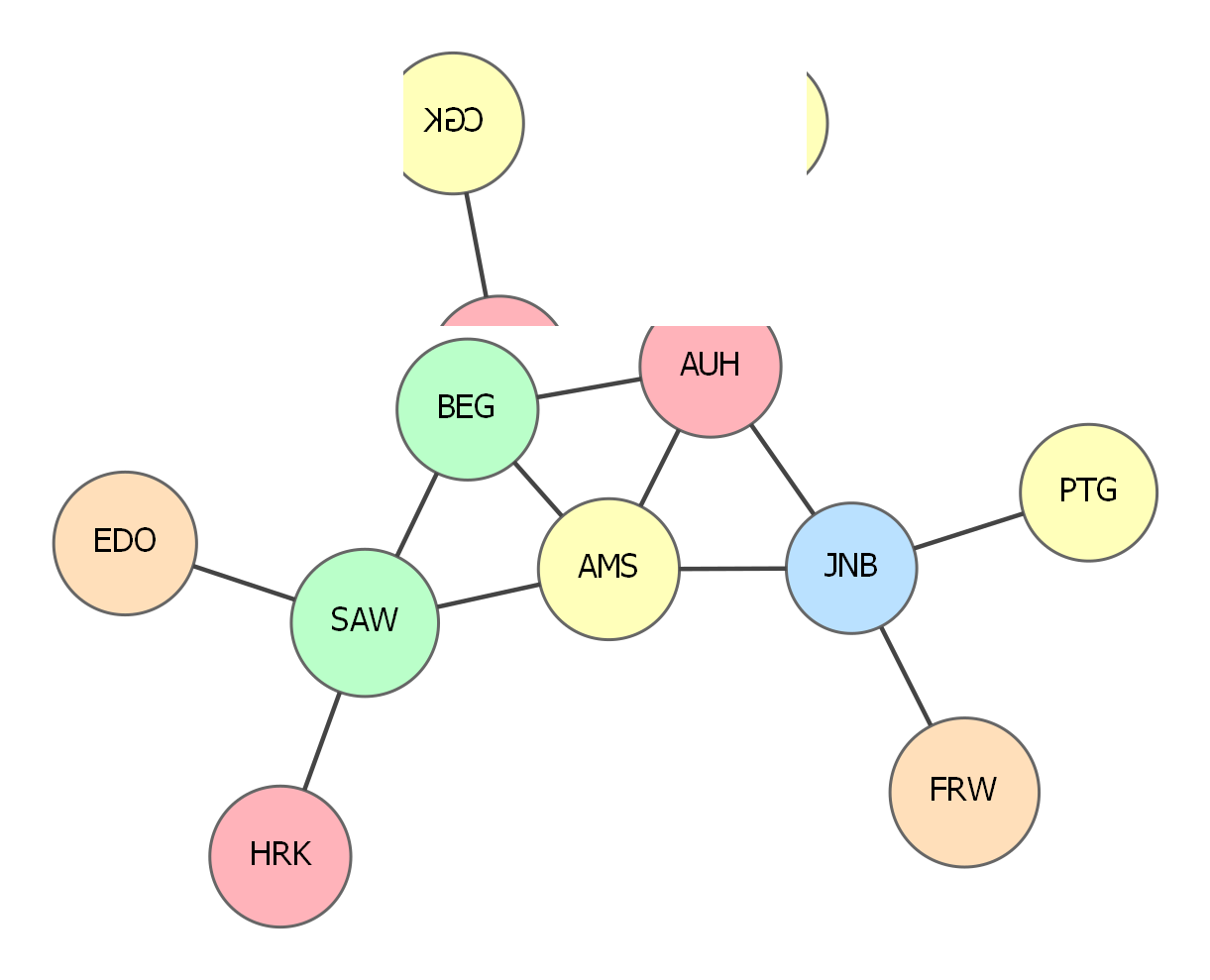}} &
        \subcaptionbox{15}{\includegraphics[width=0.18\textwidth]{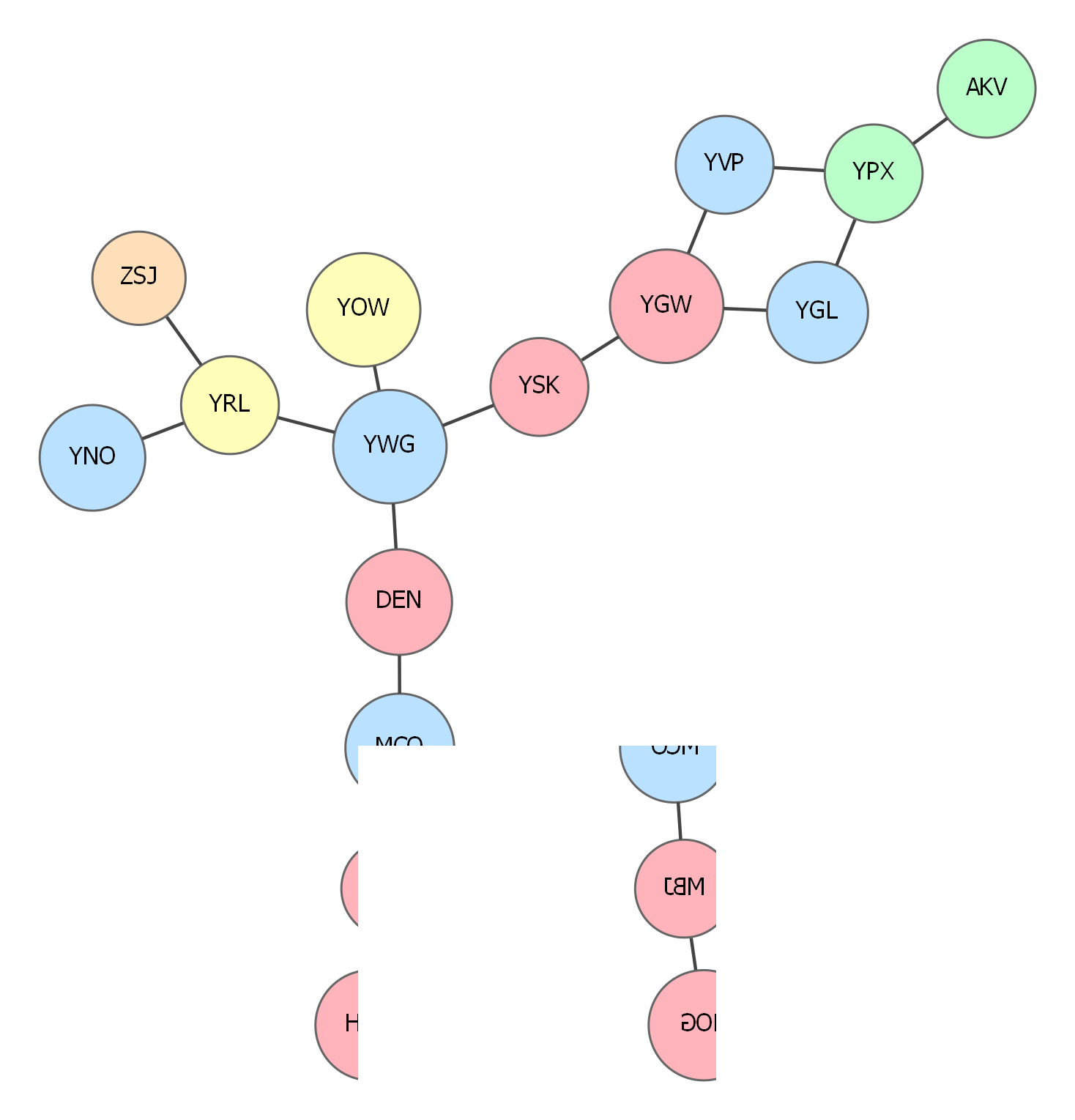}} &
        \subcaptionbox{20}{\includegraphics[width=0.18\textwidth]{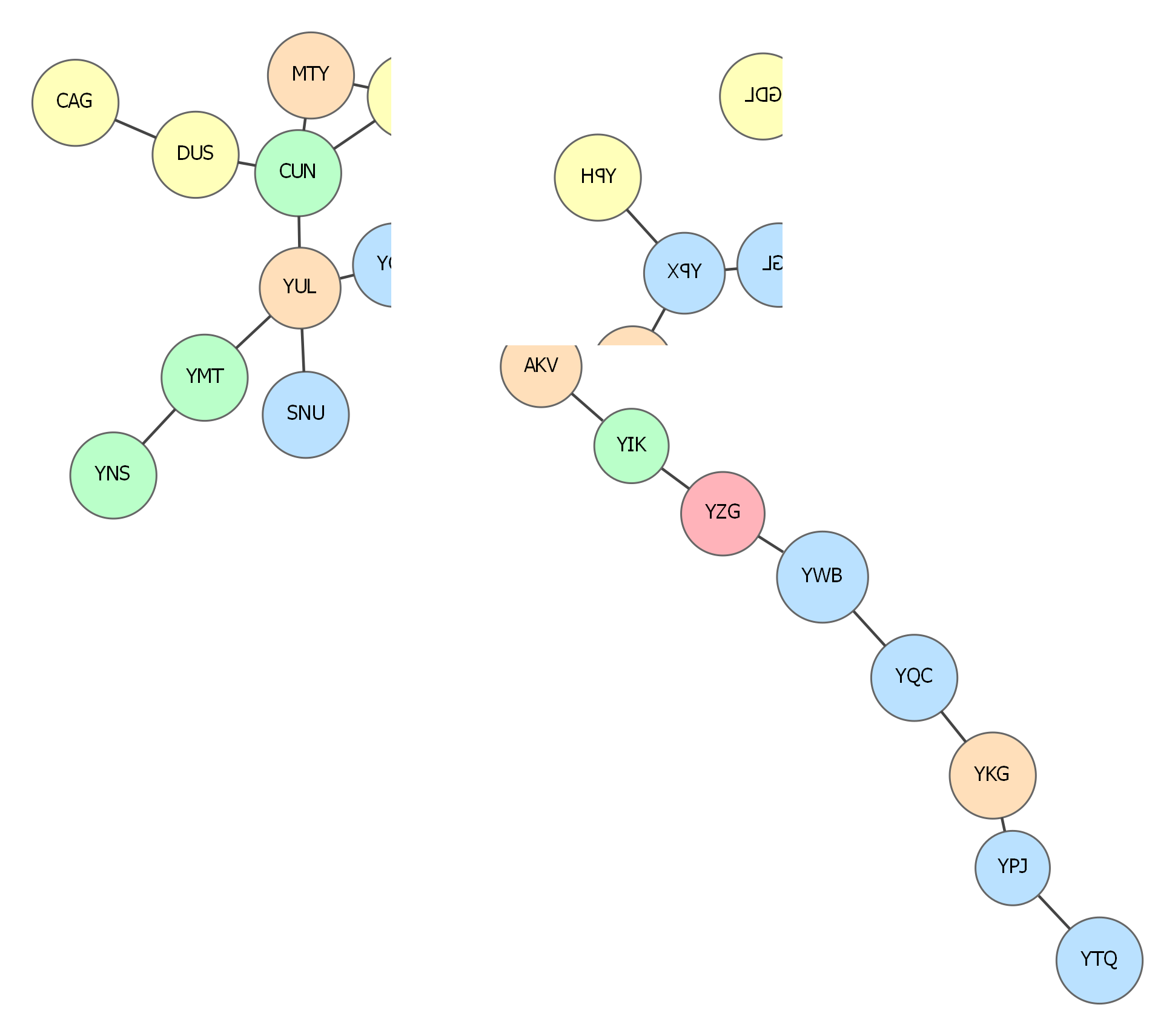}} &
        \subcaptionbox{25}{\includegraphics[width=0.18\textwidth]{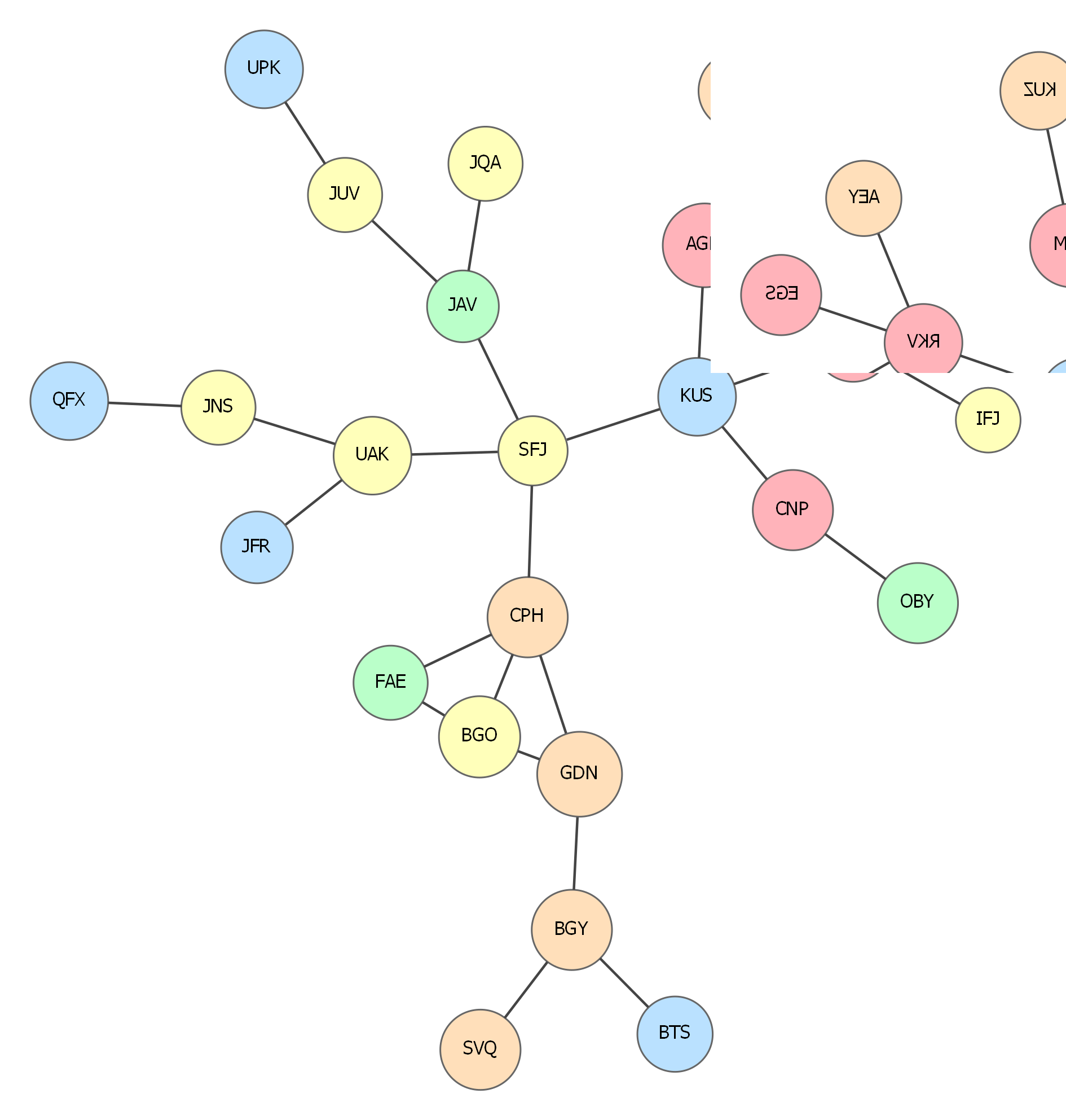}}
    \end{tabular}

    \vspace{14pt}

    {\bfseries Cross-Graph Composition}\par\vspace{3pt}
    \begin{tabular}{ccccc}
        \subcaptionbox{5}{\includegraphics[width=0.18\textwidth]{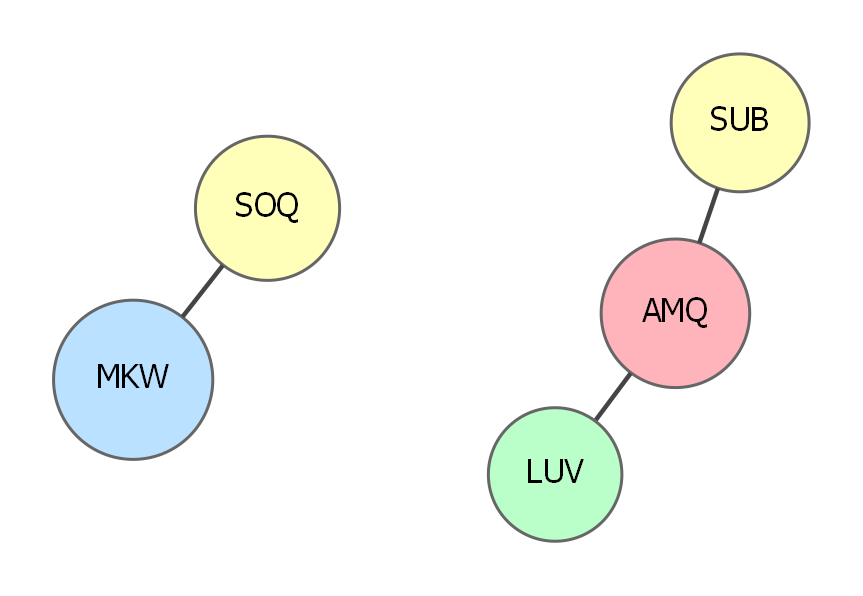}} &
        \subcaptionbox{10}{\includegraphics[width=0.18\textwidth]{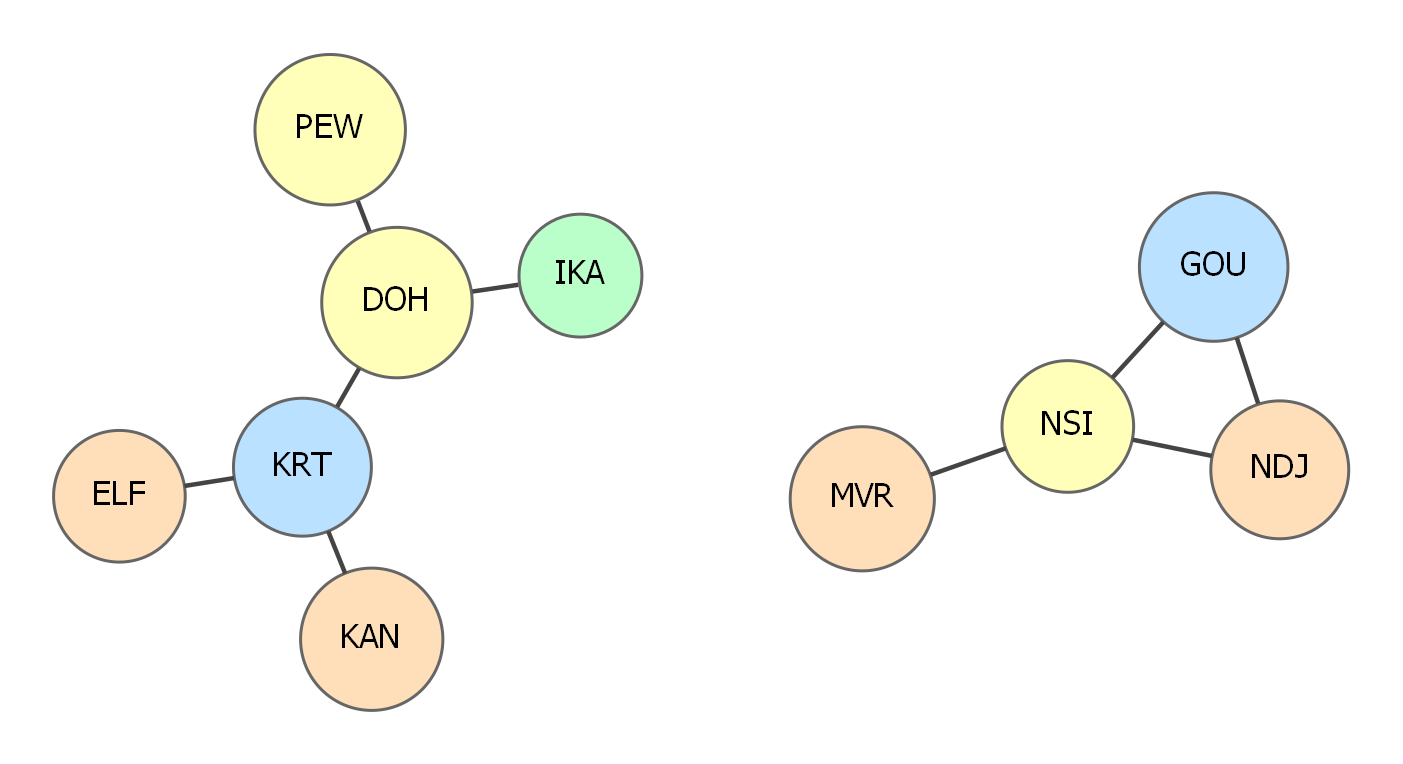}} &
        \subcaptionbox{15}{\includegraphics[width=0.18\textwidth]{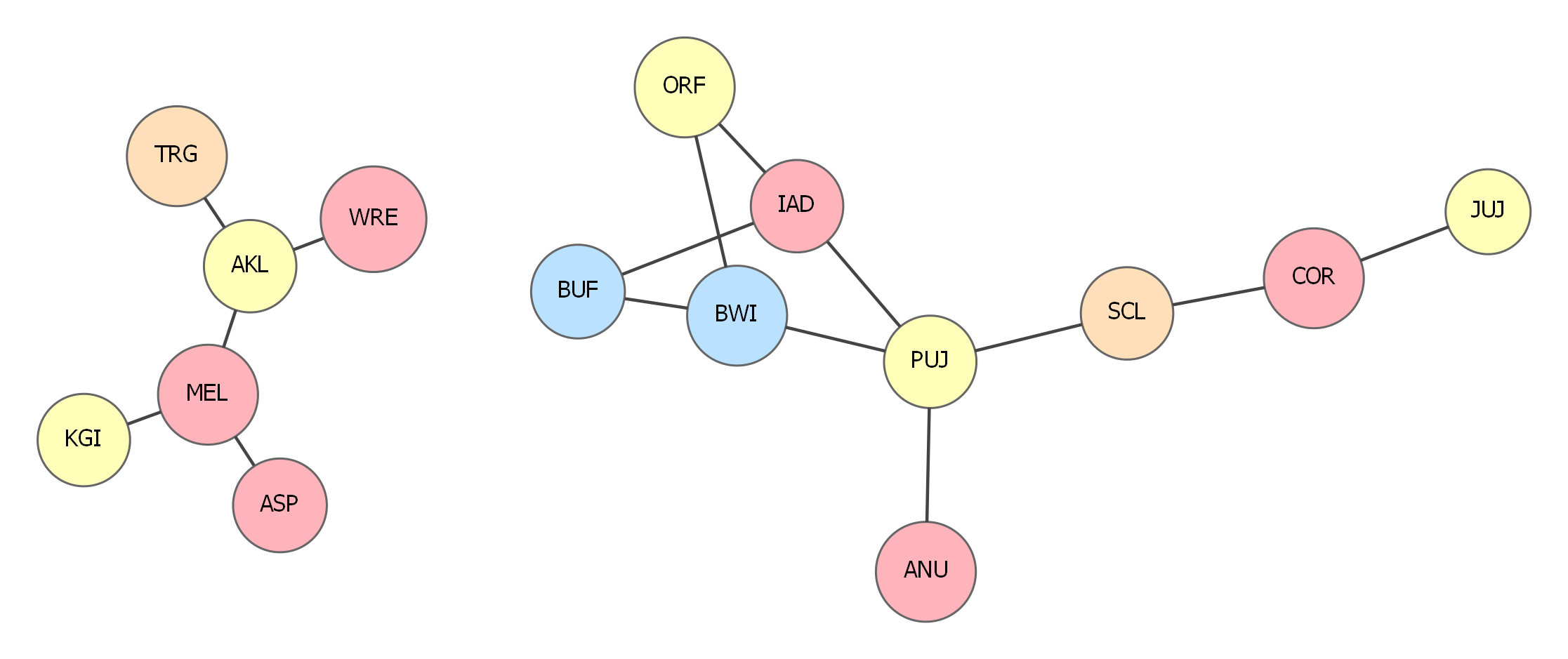}} &
        \subcaptionbox{20}{\includegraphics[width=0.18\textwidth]{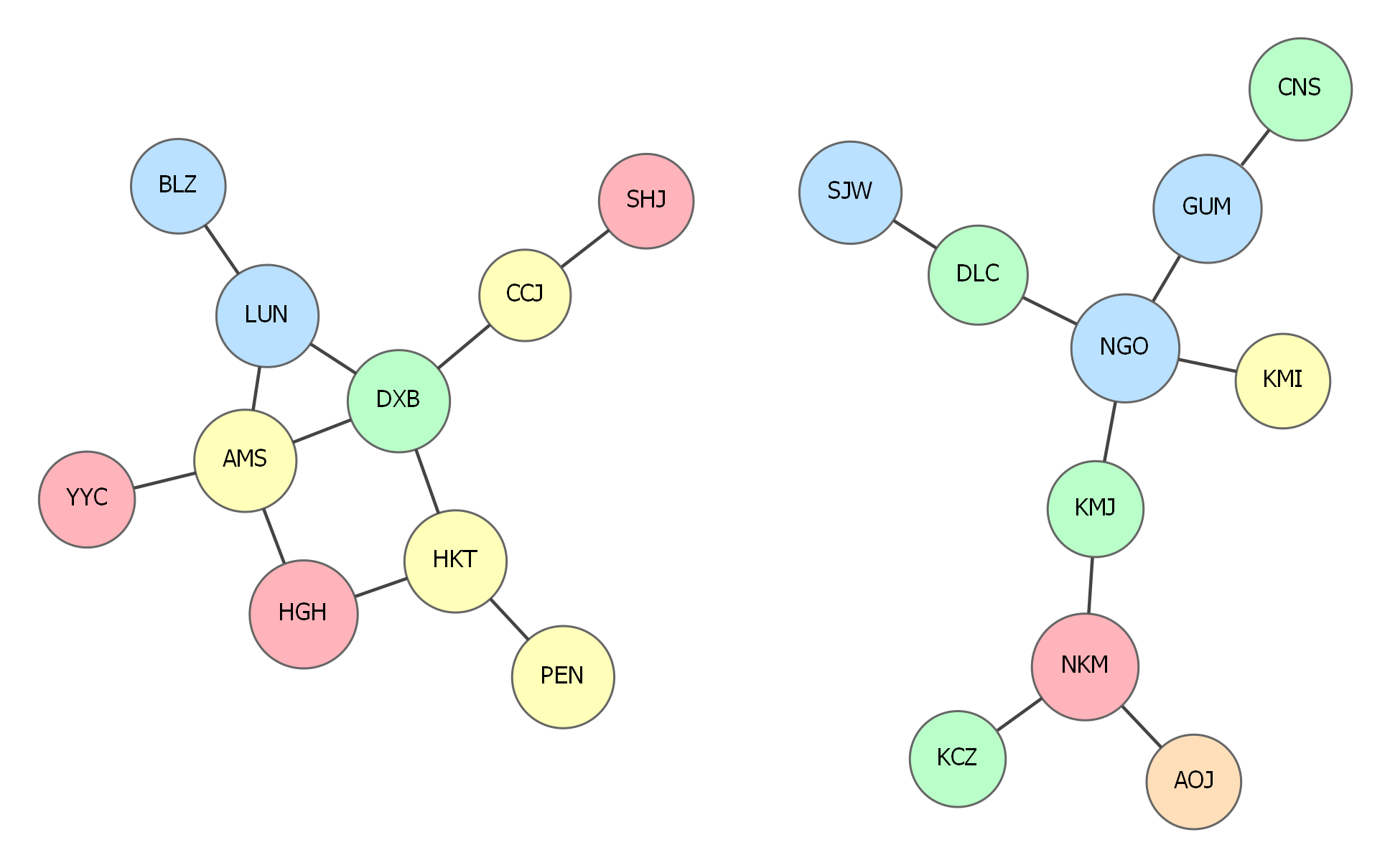}} &
        \subcaptionbox{25}{\includegraphics[width=0.18\textwidth]{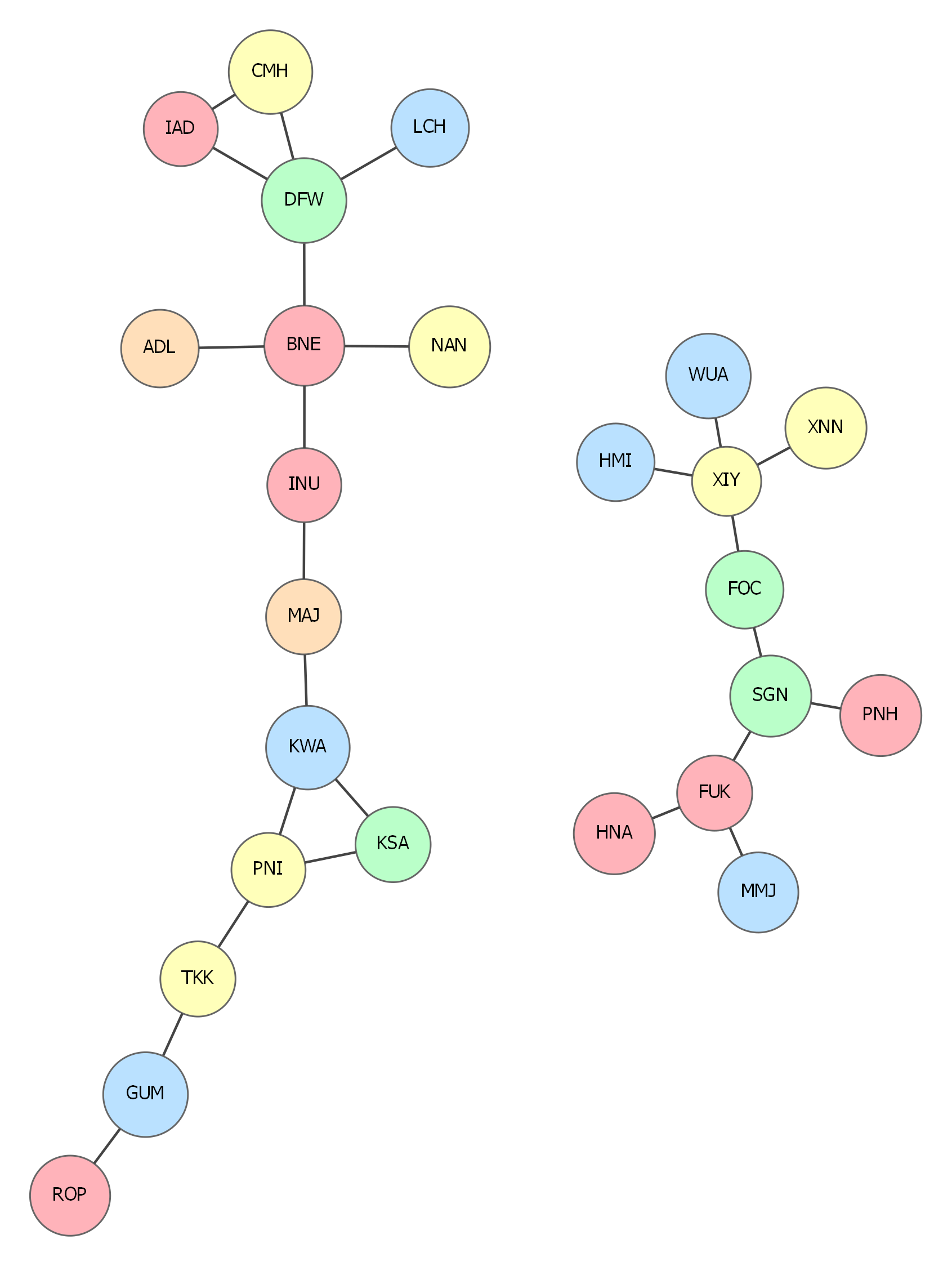}}
    \end{tabular}

    \vspace{14pt}

    {\bfseries Graph Attentional Focusing}\par\vspace{3pt}
    \begin{tabular}{ccccc}
        \subcaptionbox{5}{\includegraphics[width=0.18\textwidth]{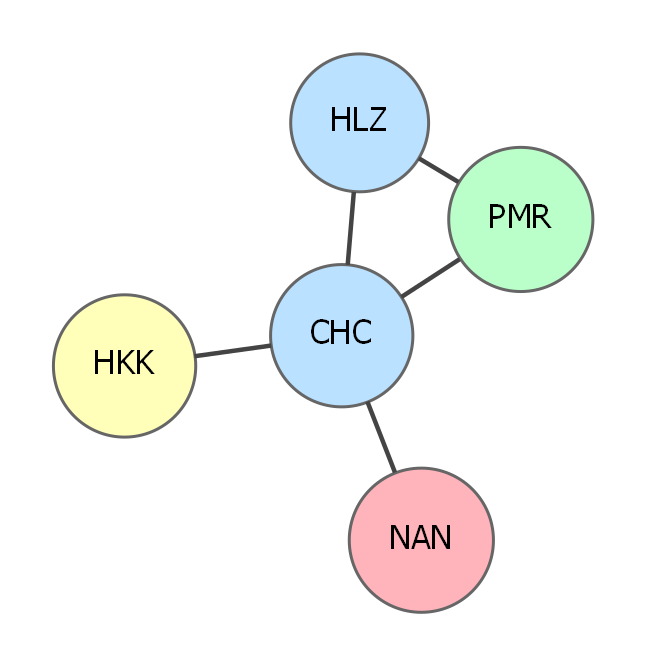}} &
        \subcaptionbox{10}{\includegraphics[width=0.18\textwidth]{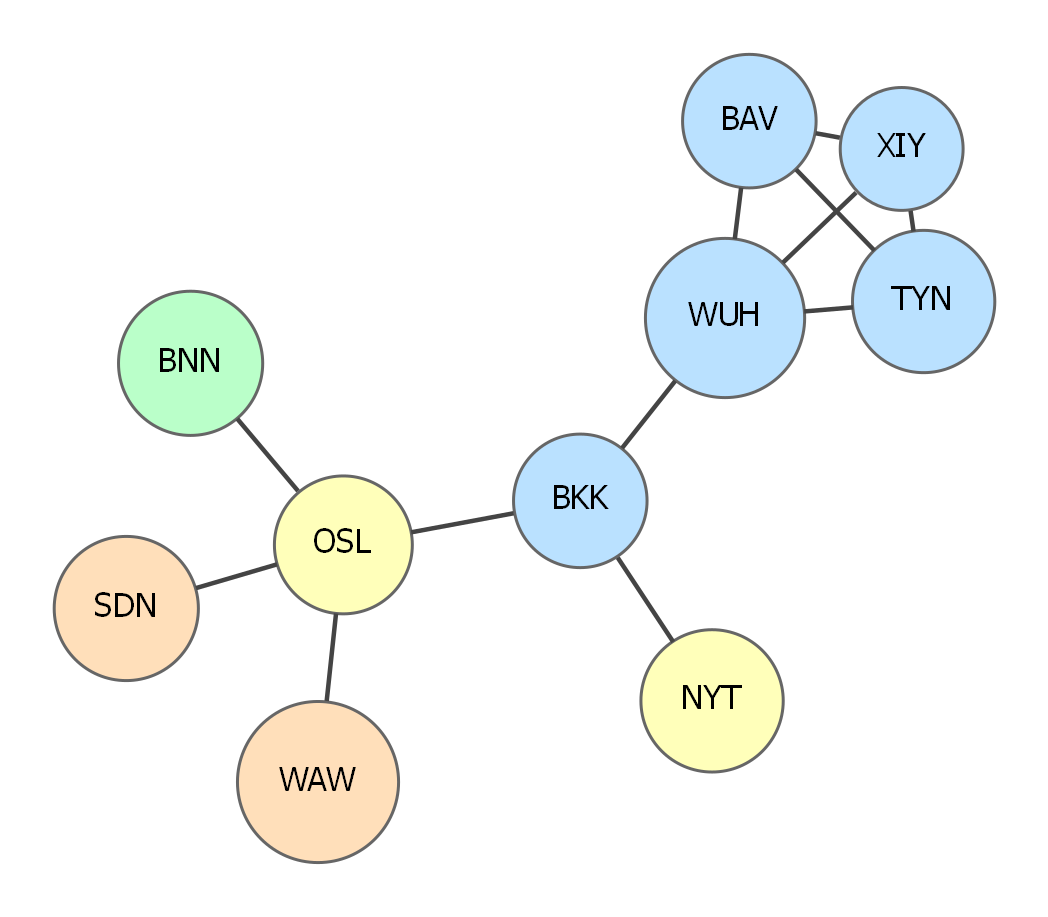}} &
        \subcaptionbox{15}{\includegraphics[width=0.18\textwidth]{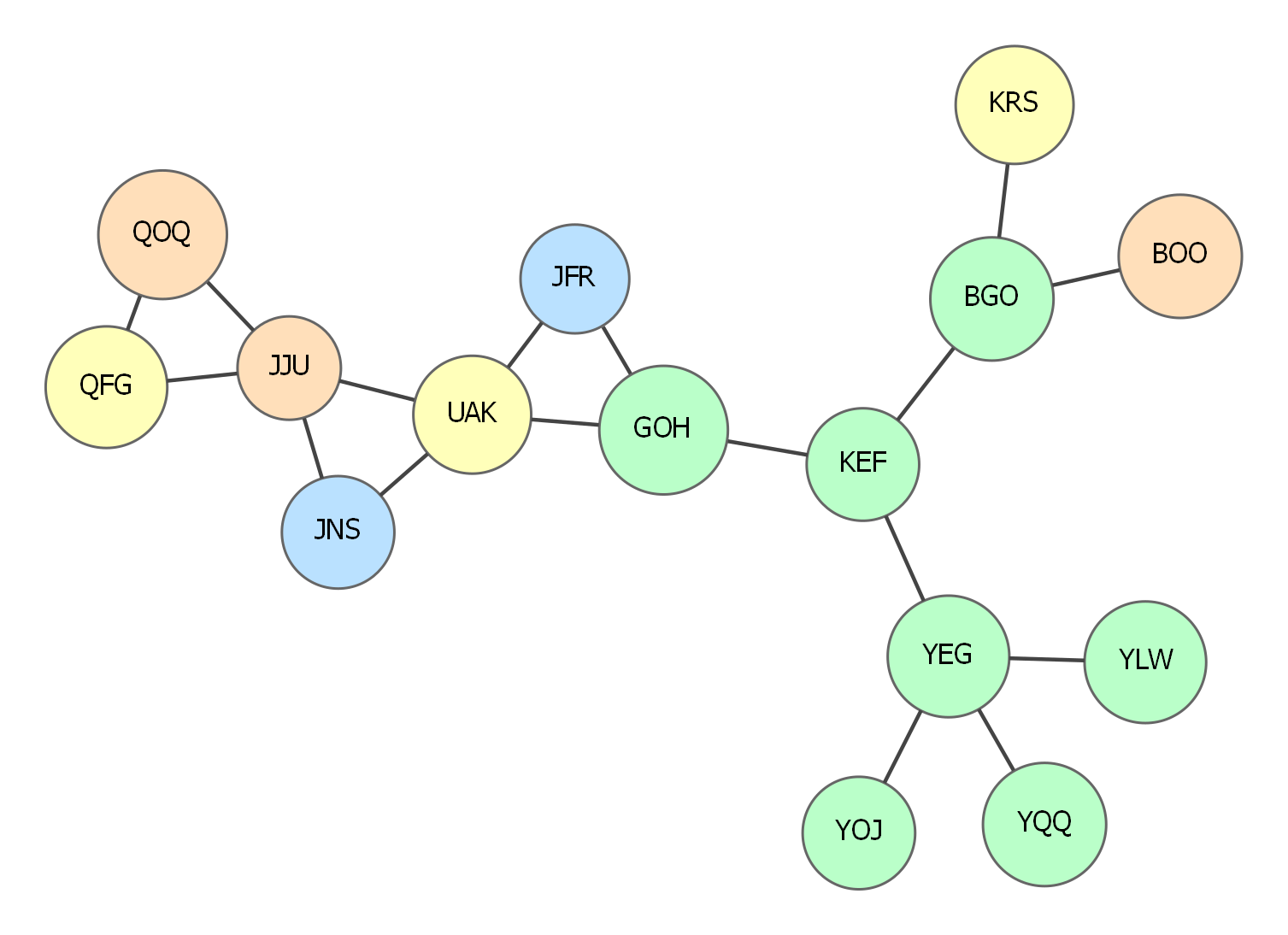}} &
        \subcaptionbox{20}{\includegraphics[width=0.18\textwidth]{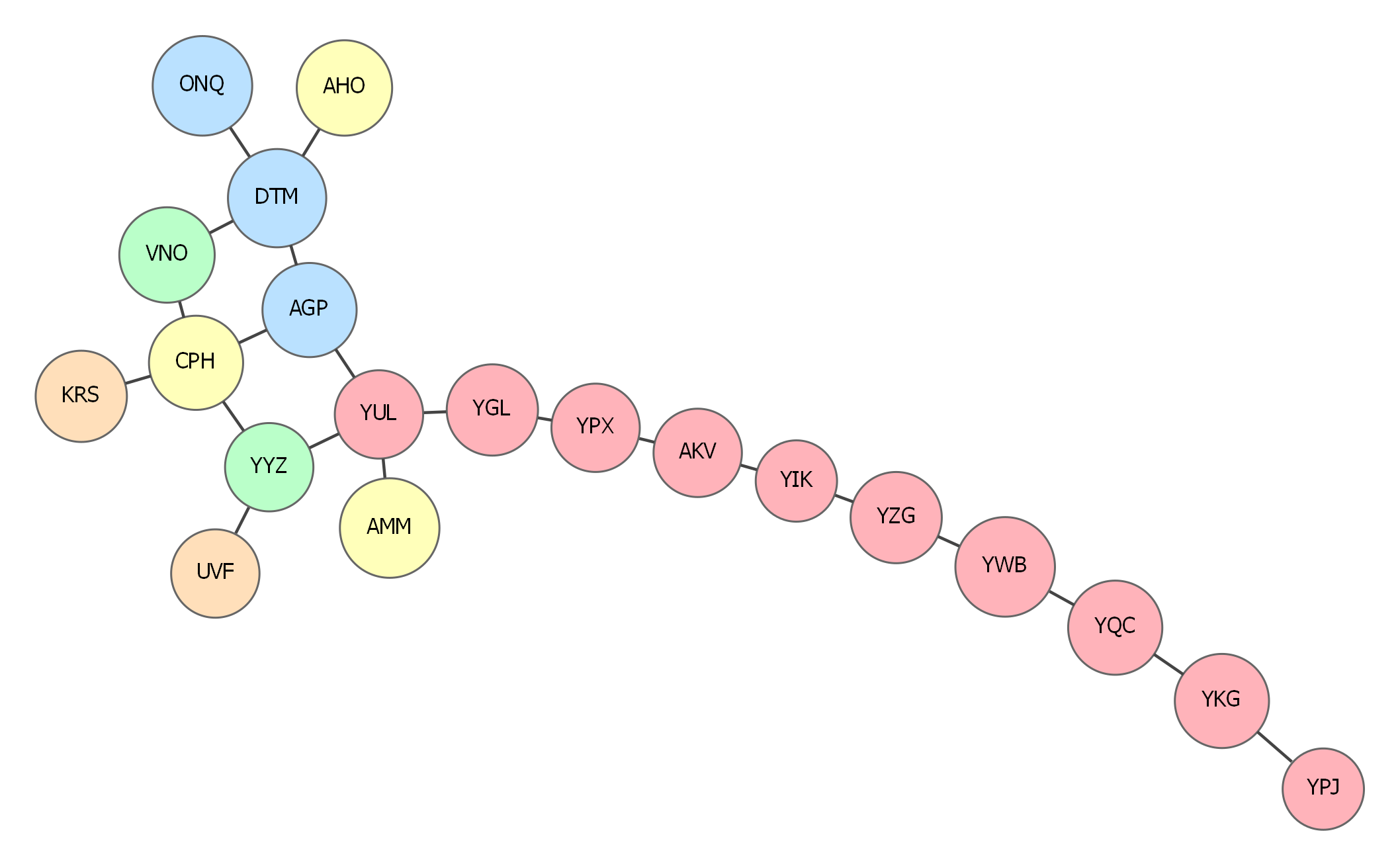}} &
        \subcaptionbox{25}{\includegraphics[width=0.18\textwidth]{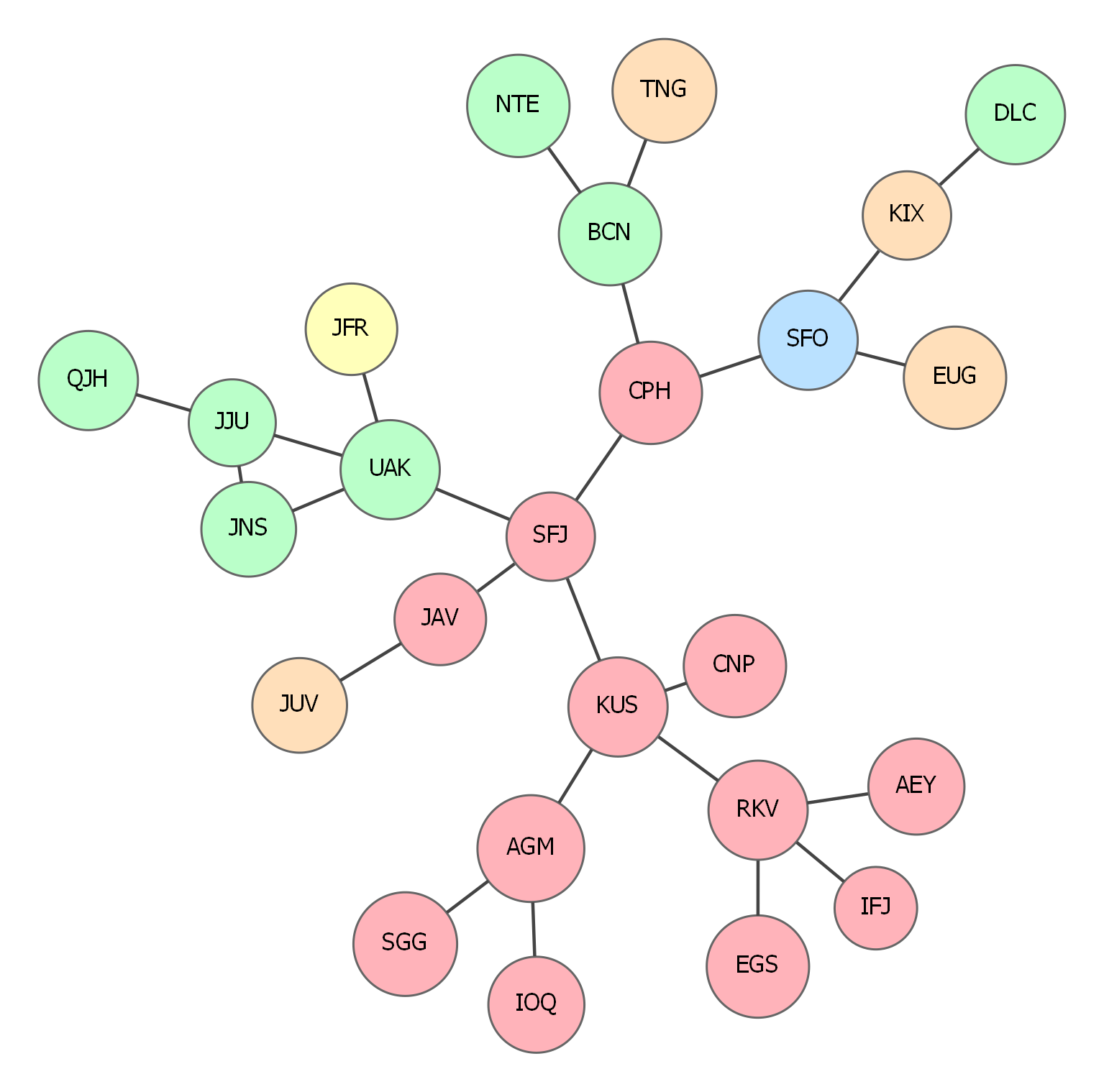}}
    \end{tabular}

    \caption{Visualization examples of four graph-centric image editing operations across different graph sizes. Within each row, subcaptions indicate the number of nodes.}
    \label{fig:gie-examples}
\end{figure*}

\clearpage

\section{The Ease of VGR Difficulty Scaling}
\label{sec:f}
Compared with traditional math-oriented benchmarks, GraphVerse is substantially easier to scale, especially in terms of difficulty. A key advantage of VGR is that task difficulty can be adjusted in a simple and controllable manner through the underlying graph structure. In particular, for most graph problems, the difficulty is closely tied to the size and connectivity pattern of the graph, such as the number of nodes, the density of edges, and the complexity of structural relations. As the graph becomes larger or structurally more intricate, the corresponding reasoning problem naturally becomes harder. This makes it possible to increase or decrease difficulty in a principled way by directly controlling graph generation or sampling.

In contrast, the difficulty scaling of traditional multimodal mathematical reasoning benchmarks is often much less straightforward. Constructing harder math problems usually requires searching for new problem instances, manually curating more challenging questions, or relying on costly expert annotation. As a result, difficulty control is typically indirect, less fine-grained, and harder to standardize. By comparison, GraphVerse supports a more efficient and flexible scaling process: one can generate large amounts of new data while maintaining explicit control over difficulty through graph size and structure. This property makes our benchmark not only easier to extend, but also more suitable for systematic evaluation across different reasoning levels.

More importantly, such scalability applies to both the algorithmic and visual dimensions of the benchmark. On the one hand, increasing the graph size or structural complexity directly raises the difficulty of the underlying graph reasoning task. On the other hand, the visual difficulty can also be adjusted through graph-centric image editing strategies, which further increase the demands on perception and visual reasoning. Therefore, GraphVerse provides a convenient testbed for constructing evaluation settings with progressively increasing difficulty, which is much harder to achieve in conventional math benchmarks. Specifically, as illustrated in Figure~\ref{fig:gie-examples}, as the number of nodes increases, the difficulty of visual graph reasoning clearly becomes higher, since the substructures involved in each operation grow more complex, which in turn increases the reasoning difficulty.

\section{Case Study}
\label{sec:g}
\begin{figure*}[t]
    \centering
    \begin{subfigure}[t]{0.24\linewidth}
        \centering
        \includegraphics[width=\linewidth]{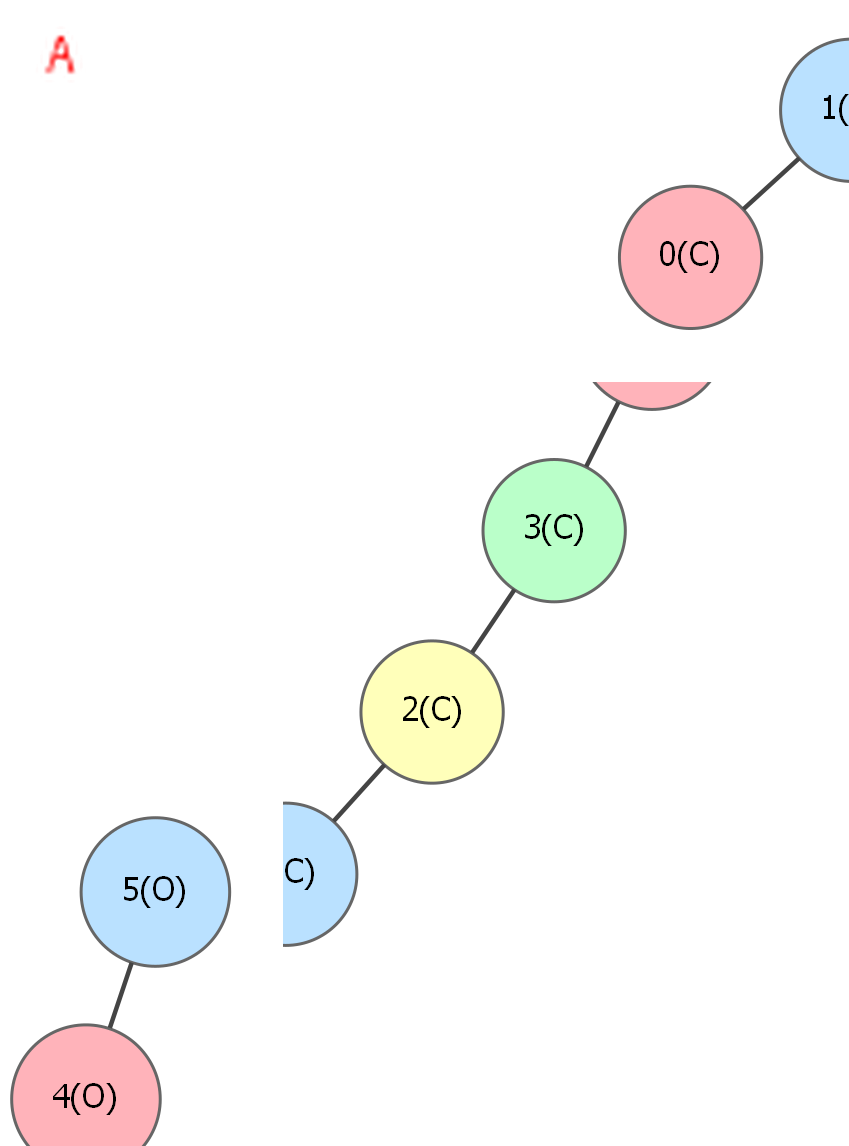}
        \caption{Graph A}
        \label{fig:image1}
    \end{subfigure}
    \hfill
    \begin{subfigure}[t]{0.24\linewidth}
        \centering
        \includegraphics[width=\linewidth]{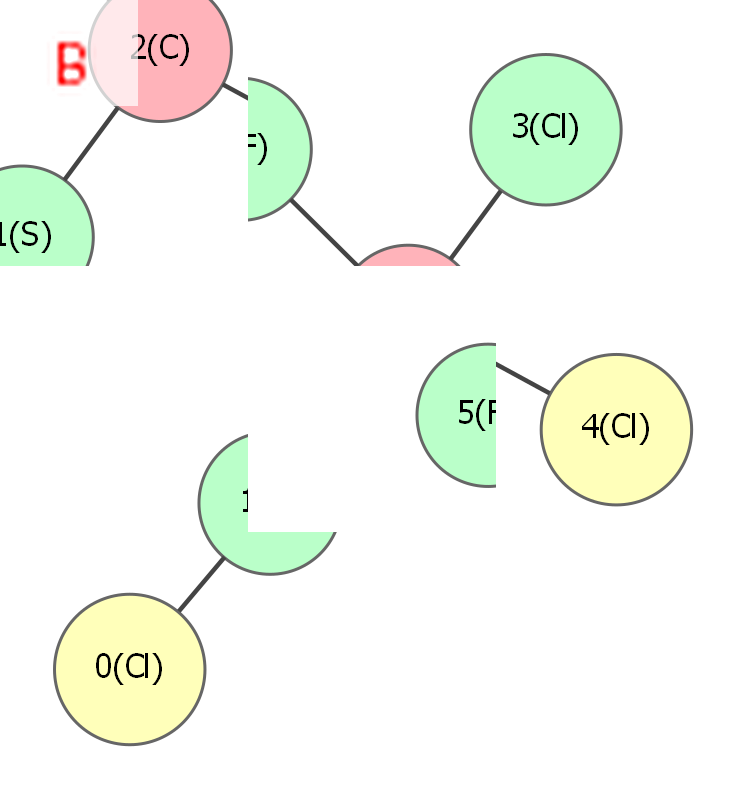}
        \caption{Graph B}
        \label{fig:image2}
    \end{subfigure}
    \caption{Graph Edit Evidence Image-Pair.}
    \label{fig:two_images}
\end{figure*}

In this section, we present several representative examples of model responses on visual graph reasoning tasks. Due to space limitations, we will include additional examples in the open-source code repository. All examples shown here are taken from the responses of Gemini-3-Pro. Figure~\ref{fig:case4} presents a Single-Image VGR example under the Spatially-Conditioned Recoloring setting, while Figure~\ref{fig:grapg1} shows the corresponding rendered graph image. Figures~\ref{fig:case5},~\ref{fig:case6},~\ref{fig:ged_prompt}, and~\ref{fig:two_images} illustrate a Paired-Image VGR example, including the model response, the Scoring Agent’s evaluation process, the original question, and the rendered images. Finally, we present the prompts of some representative tasks in our experiments.

\begin{table*}[htbp]  
\centering
\small
\setlength{\tabcolsep}{4pt}
\renewcommand{\arraystretch}{1.2}
\setlength{\arrayrulewidth}{0.3mm}
\begin{tabular}{l|l|l}
\hline
\rowcolor{CadetBlue!20}
\textbf{Task Name} & \textbf{Node Description} & \textbf{Edge Description} \\
\hline
TSP (Traveling Salesman Problem) & Airport (by code/name) & Nonstop flight routes (great-circle distance) \\
\hline
Cycle (Longest Simple Cycle) & Airport (by code/name) & Nonstop flight routes \\
\hline
Graph Coloring (DBLP) & Author (researcher) & Co-authorship (previous joint publication) \\
\hline
Diameter (DBLP) & Entity (concept entity) & Relationship (conceptual connection) \\
\hline
Neighbor (DBLP) & Author (researcher) & Co-authorship (joint publication between authors) \\
\hline
GED (Graph Edit Distance) & Atom (labeled by element) & Bond (covalent bond between atoms) \\
\hline
MCS (Maximum Common Subgraph) & Atom (labeled by element) & Bond (covalent bond between atoms) \\
\hline
MIS (Maximum Independent Set) & User (community member) & Friendship (social connection between users) \\
\hline
MVC (Minimum Vertex Cover) & Liaison (regional liaison) & Monitored connection (liaison connection) \\
\hline
MCP (Maximum Clique Problem) & Author (researcher) & Co-authorship (joint publication between authors) \\
\hline
Distance (DBLP) & Entity (concept entity) & Relationship (conceptual connection) \\
\hline
\end{tabular}
\caption{Graph Task Node and Edge Descriptions}
\label{tab:graph_task_desc}
\end{table*}

\begin{figure*}[t]
\centering
\caption{A visual graph reasoning case under spatially-conditioned recoloring, where the model identifies a representative longest simple cycle after applying the recoloring rule.}
\label{fig:case4}
\begin{caseboxpurple}{\scriptsize Case Study: Longest Cycle with Spatially-Conditioned Recoloring}
\scriptsize

{\scriptsize
\textbf{Reasoning Highlights:}
\vp{Visual Perception} \quad
\vr{Vision-based Reasoning} \quad
\tgr{Text-based Graph Reasoning}
}

\vspace{2pt}
\textbf{Question Prompt:}\\[-1pt]
\fontsize{7}{8}\selectfont
You are an airline network reliability engineer auditing the redundant loops that keep regional operations resilient during diversions.
Analyze the undirected flight network to find the single longest simple cycle (a closed loop with no repeated intermediate airports). If several loops tie for that maximum length, you may brief dispatch with any one representative loop.
Each node shows an airport's name or code, and each edge is an active nonstop corridor; distance annotations are intentionally omitted.
Color legend (operational priority tiers): rose=critical insight, peach=high-priority concept, lemon=standard fact, mint=supporting detail, sky=background reference.

Cardinal directions are determined by comparing circle centers (a node is ``above'' another if its center is higher, and ``left of'' another if its center is further to the left).

Scenario: operations retagged priority tiers. Current palette: lemon, mint, peach, rose, sky. As part of the redundancy tagging update, mentally recolor every airport located above anchor EMK (mint) to lemon while keeping other airports as rendered, then identify the longest loop (report airport names only).

Task: After applying the recoloring instructions, provide one representative longest cycle (airport names only). Also, examine the color of each node within the longest cycle and its length. List the airports in travel order and repeat the starting airport at the end to close the loop.

Answer format: \texttt{representative longest cycle: [Airport (color) -> ... -> Airport (color)]; length: <value> airports}

Example: \texttt{representative longest cycle: [DEN (blue) -> ABQ (green) -> PHX (yellow) -> DEN (blue)]; length: 3 airports}
\scriptsize

\vspace{2pt}
\textbf{Model Response:}\\[-1pt]
To identify the longest simple cycle in the flight network, we first analyze the graph structure and then apply the recoloring rule.

\vspace{2pt}
\textbf{\vp{Graph Structure.}}\\[-1pt]
\vp{The network contains airports EMK, KOT, WBB, UNK, ANC, SCC, FAI, BTI, AUK, SXP, and KSM, with nonstop corridors $(\mathrm{EMK},\mathrm{KOT})$, $(\mathrm{EMK},\mathrm{AUK})$, $(\mathrm{KOT},\mathrm{WBB})$, $(\mathrm{KOT},\mathrm{KSM})$, $(\mathrm{WBB},\mathrm{UNK})$, $(\mathrm{UNK},\mathrm{ANC})$, $(\mathrm{AUK},\mathrm{SXP})$, $(\mathrm{SXP},\mathrm{KSM})$, $(\mathrm{KSM},\mathrm{ANC})$, $(\mathrm{ANC},\mathrm{FAI})$, $(\mathrm{ANC},\mathrm{SCC})$, $(\mathrm{SCC},\mathrm{FAI})$, and $(\mathrm{SCC},\mathrm{BTI})$. The graph contains a large left component with an outer loop and an internal chord $(\mathrm{KOT},\mathrm{KSM})$, as well as a small triangle formed by ANC, SCC, and FAI.}

\vspace{2pt}
\textbf{\vr{Recoloring.}}\\[-1pt]
\vr{The rule states: ``recolor every airport located above anchor EMK (mint) to lemon.'' From the spatial layout, WBB, KOT, and UNK are above EMK and are therefore recolored to lemon, while the remaining airports keep their rendered colors. Thus, along the final cycle, the updated colors are EMK (mint), AUK (lemon), SXP (rose), KSM (peach), ANC (sky), UNK (lemon), WBB (lemon), and KOT (lemon).}

\vspace{2pt}
\textbf{\tgr{Cycle Identification.}}\\[-1pt]
\tgr{The graph contains a triangle $\mathrm{ANC}\!-\!\mathrm{SCC}\!-\!\mathrm{FAI}\!-\!\mathrm{ANC}$ of length 3, as well as two smaller 5-node loops induced by the chord $(\mathrm{KOT},\mathrm{KSM})$. The representative longest simple cycle is the outer boundary of the left component:}
\[
\tgr{\mathrm{EMK}\!\rightarrow\!\mathrm{AUK}\!\rightarrow\!\mathrm{SXP}\!\rightarrow\!\mathrm{KSM}\!\rightarrow\!\mathrm{ANC}\!\rightarrow\!\mathrm{UNK}\!\rightarrow\!\mathrm{WBB}\!\rightarrow\!\mathrm{KOT}\!\rightarrow\!\mathrm{EMK}}
\]
\tgr{Its length is 8 airports.}

\vspace{2pt}
\textbf{Final Answer:}\\[-1pt]
\texttt{representative longest cycle: [EMK (mint) -> AUK (lemon) -> SXP (rose) -> KSM (peach) -> ANC (sky) -> UNK (lemon) -> WBB (lemon) -> KOT (lemon) -> EMK (mint)]; length: 8 airports}

\end{caseboxpurple}
\end{figure*}

\begin{figure*}
    \centering
    \includegraphics[width=0.45\linewidth]{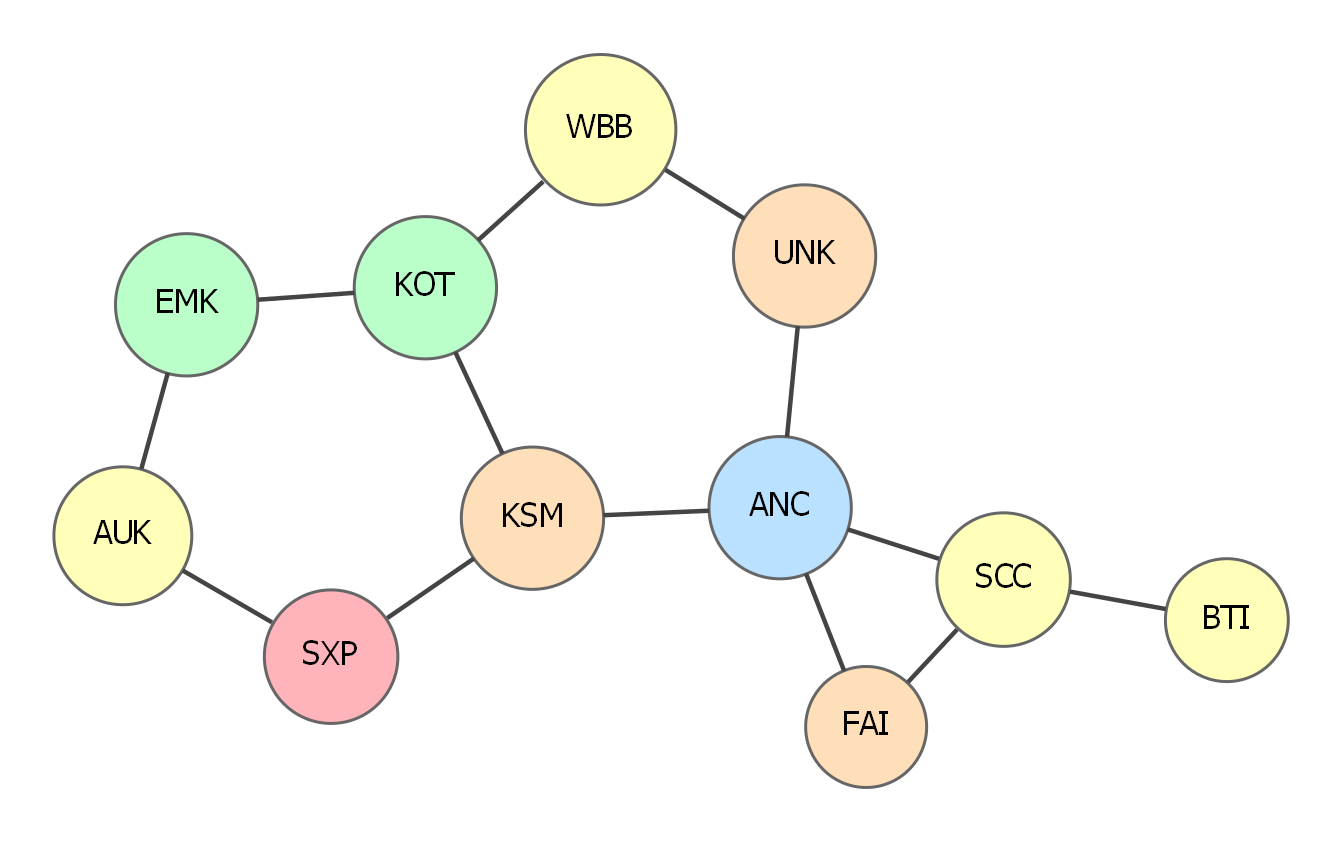}
    \caption{Rendered Graph of Figure~\ref{fig:case4}.}
    \label{fig:grapg1}
\end{figure*}

\begin{figure*}[t]
\centering
\caption{A case of visual graph reasoning under patch-swap perturbation, where the model reconstructs two molecular graphs from fragmented images and identifies the atom-level mapping that minimizes graph edit distance.}
\label{fig:case5}
\begin{caseboxpurple}{\fontsize{6.3}{7.0}\selectfont Case Study: Graph Edit Distance with Patch-Swap Perturbation}
\fontsize{5.9}{6.5}\selectfont

{\scriptsize
\textbf{Reasoning Highlights:}
\vp{Visual Perception} \quad
\vr{Vision-based Reasoning} \quad
\tgr{Text-based Graph Reasoning}
}

\textbf{Model Response:}\\[-2pt]
To find the atom-level mapping from Molecule A to Molecule B that minimizes the Graph Edit Distance (GED), we first need to determine the correct graph structures for both molecules from the fragmented images.

\vspace{1pt}
\textbf{1. Structural Analysis.}\\[-2pt]
\textbf{Molecule A (Image 1):}
\vp{Based on the visual cues and ``patch-swap'' description, the image consists of patches containing nodes labeled \texttt{0(C)}, \texttt{1(C)}, \texttt{2(C)}, \texttt{3(C)}, \texttt{4(O)}, and \texttt{5(O)}.}
\vr{For connectivity tracing, in the top-right region, \texttt{1(C)} (blue) is connected to \texttt{0(C)} (pink). In the middle region, \texttt{0(C)} connects to \texttt{3(C)} (green), and \texttt{3(C)} connects to \texttt{2(C)} (yellow). In the bottom region, the tile with \texttt{2(C)} shows a bond going to the bottom-left, while the bottom-left tile contains \texttt{5(O)} (blue) connected to \texttt{4(O)} (pink), so visual alignment suggests that \texttt{2(C)} connects to \texttt{5(O)}. Therefore, Structure A is a linear chain: \textbf{1(C) -- 0(C) -- 3(C) -- 2(C) -- 5(O) -- 4(O)}. The attributes are: 0: C, Pink; 1: C, Blue; 2: C, Yellow; 3: C, Green; 4: O, Pink; 5: O, Blue.}

\textbf{Molecule B (Image 2):}
\vp{This image is also tiled, and we identify nodes \texttt{0(Cl)}, \texttt{1(S)}, \texttt{2(C)}, \texttt{3(Cl)}, \texttt{4(Cl)}, and \texttt{5(F)}.}
\vr{For connectivity tracing, in the top-left/center region, \texttt{1(S)} (green) is connected to \texttt{2(C)} (pink). On the left/bottom, \texttt{0(Cl)} (yellow) is connected to \texttt{1(S)}. On the right/branching side, \texttt{2(C)} acts as a central hub: it connects to \texttt{1(S)}, to \texttt{3(Cl)} (green), and also to \texttt{5(F)} (green). At the tail, \texttt{5(F)} is connected to \texttt{4(Cl)} (yellow). Therefore, Structure B is a branched structure centered at \texttt{2(C)}: Branch 1 is \textbf{0(Cl) -- 1(S) -- 2(C)}, Branch 2 is \textbf{3(Cl)} attached to \textbf{2(C)}, and Branch 3 is \textbf{4(Cl) -- 5(F) -- 2(C)}. The attributes are: 0: Cl, Yellow; 1: S, Green; 2: C, Pink; 3: Cl, Green; 4: Cl, Yellow; 5: F, Green.}

\vspace{1pt}
\textbf{2. Graph Edit Distance Calculation.}\\[-2pt]
\tgr{We assume a cost of 1 for any node substitution if the element or color differs, and a cost of 1 for any edge insertion or deletion. For node mapping, the only perfect match is \texttt{A0} (C, Pink) with \texttt{B2} (C, Pink), so mapping \texttt{A0 -> B2} has cost 0. All other mappings involve a mismatch in element, color, or both, so \texttt{A1}, \texttt{A2}, \texttt{A3}, \texttt{A4}, and \texttt{A5} each incur cost 1, giving a total node cost of 5.}

\tgr{For edge mapping optimization, we align the linear chain of A (\texttt{1-0-3-2...}) onto the branches of B centered at \texttt{B2}. Under the anchor mapping \texttt{A0 -> B2}, the neighbors of A0 are A1 and A3, while the neighbors of B2 are B1, B3, and B5. The best topological fit is to map \texttt{A1 -> B3}, so edge \texttt{(A1,A0)} maps to \texttt{(B3,B2)} and is preserved; map \texttt{A3 -> B1}, so edge \texttt{(A0,A3)} maps to \texttt{(B2,B1)} and is preserved; then continue the chain by mapping \texttt{A2 -> B0}, so edge \texttt{(A3,A2)} maps to \texttt{(B1,B0)} and is also preserved. For the tail, the rest of A is that \texttt{A2} connects to \texttt{A5}, which connects to \texttt{A4}, while the rest of B is that \texttt{B2} connects to \texttt{B5}, which connects to \texttt{B4}. We map \texttt{A5 -> B5} and \texttt{A4 -> B4}, so edge \texttt{(A5,A4)} maps to \texttt{(B5,B4)} and is preserved. However, there are still two mismatches: in A, \texttt{A5} is connected to \texttt{A2}, but under the mapping this would require edge \texttt{(B5,B0)}, which does not exist, leading to one edge deletion; in B, \texttt{B5} is connected to \texttt{B2}, but the corresponding nodes \texttt{A5} and \texttt{A0} are not connected in A, leading to one edge insertion. Thus, the total edge cost is 2. Therefore, the total edit distance is \textbf{5 (nodes) + 2 (edges) = 7}.}

\textbf{\tgr{3. Python Solution.}}\\[-2pt]
{\color{purple!80!black}\fontsize{4.8}{5.2}\selectfont
\begin{multicols}{2}
\begin{lstlisting}[
language=Python,
basicstyle=\ttfamily\color{purple!80!black}\fontsize{4.8}{5.2}\selectfont,
keywordstyle=\color{purple!80!black},
commentstyle=\color{purple!80!black},
stringstyle=\color{purple!80!black},
breaklines=true
]
import networkx as nx
import itertools

def solve_ged():
    nodes_A = {
        0: {'elem': 'C', 'color': 'Pink'},
        1: {'elem': 'C', 'color': 'Blue'},
        2: {'elem': 'C', 'color': 'Yellow'},
        3: {'elem': 'C', 'color': 'Green'},
        4: {'elem': 'O', 'color': 'Pink'},
        5: {'elem': 'O', 'color': 'Blue'}
    }
    edges_A = [(1,0), (0,3), (3,2), (2,5), (5,4)]

    nodes_B = {
        0: {'elem': 'Cl', 'color': 'Yellow'},
        1: {'elem': 'S',  'color': 'Green'},
        2: {'elem': 'C',  'color': 'Pink'},
        3: {'elem': 'Cl', 'color': 'Green'},
        4: {'elem': 'Cl', 'color': 'Yellow'},
        5: {'elem': 'F',  'color': 'Green'}
    }
    edges_B = [(0,1), (1,2), (2,3), (2,5), (5,4)]

    adj_B = {i: set() for i in range(6)}
    for u, v in edges_B:
        adj_B[u].add(v)
        adj_B[v].add(u)

    min_cost = float('inf')
    best_mapping = []

    for perm in itertools.permutations(range(6)):
        node_cost = 0
        for a_idx, b_idx in enumerate(perm):
            attr_a = nodes_A[a_idx]
            attr_b = nodes_B[b_idx]
            if attr_a['elem'] != attr_b['elem'] or attr_a['color'] != attr_b['color']:
                node_cost += 1

        mapped_edges_A = 0
        for u, v in edges_A:
            b_u, b_v = perm[u], perm[v]
            if b_v in adj_B[b_u]:
                mapped_edges_A += 1

        edge_matches = mapped_edges_A
        edge_deletions = len(edges_A) - edge_matches
        edge_insertions = len(edges_B) - edge_matches
        total_cost = node_cost + edge_deletions + edge_insertions

        if total_cost < min_cost:
            min_cost = total_cost
            best_mapping = list(perm)

    print(f"edit_distance: {min_cost}; mapping: {best_mapping}")

solve_ged()
\end{lstlisting}
\end{multicols}
}

\vspace{1pt}
\textbf{4. Result.}\\[-2pt]
Running the analysis confirms the minimum edit distance is 7. The optimal mapping aligns the single matching Carbon atom (\texttt{A0 -> B2}) and maps the rest to minimize structural disruption.

\vspace{1pt}
\textbf{Final Answer:}\\[-2pt]
\texttt{edit\_distance: 7; mapping: [2, 3, 0, 1, 4, 5]}

\end{caseboxpurple}
\end{figure*}

\begin{figure*}
\caption{A visual graph reasoning case under spatially-conditioned recoloring, where the model identifies a representative longest simple cycle after applying the recoloring rule. (Output of Scoring Agent).}
\label{fig:case6}
\begin{caseboxpurple}{\scriptsize Judge Feedback}
\fontsize{5.8}{6.4}\selectfont

\textbf{1) Key Operations:}\\[-2pt]
(1) Claimed Molecule A node labels/elements: nodes 0--5 are 0:C, 1:C, 2:C, 3:C, 4:O, 5:O.\\
(2) Claimed Molecule A connectivity: chain 1--0--3--2--5--4 (edges \texttt{(1,0)}, \texttt{(0,3)}, \texttt{(3,2)}, \texttt{(2,5)}, \texttt{(5,4)}).\\
(3) Stated Structure A is the linear chain 1--0--3--2--5--4.\\
(4) Listed Molecule A node attributes (colors) for nodes 0--5.\\
(5) Claimed Molecule B node labels/elements: 0:Cl, 1:S, 2:C, 3:Cl, 4:Cl, 5:F.\\
(6) Claimed Molecule B connectivity: branched centered at node 2 with branches 0--1--2, 2--3, and 2--5--4 (implying edge \texttt{(5,4)}).\\
(7) Stated Structure B as described in (6).\\
(8) Asserted only A0 matches B2 in element/color and thus should be mapped A0$\rightarrow$B2.\\
(9) Asserted all other node substitutions cost 1 each, giving node substitution cost = 5.\\
(10) Proposed the detailed mapping: A0$\rightarrow$B2, A1$\rightarrow$B3, A3$\rightarrow$B1, A2$\rightarrow$B0, A5$\rightarrow$B5, A4$\rightarrow$B4 (mapping vector \texttt{[2,3,0,1,4,5]}) and claimed many corresponding edges match under this mapping.\\
(11) Computed edge insertions/deletions under that mapping, giving edge cost = 2.\\
(12) Summed node + edge costs to give total edit distance = 7 and reported final answer \texttt{edit\_distance: 7; mapping: [2,3,0,1,4,5]}.\\[2pt]

\textbf{2) Judge Analysis:}\\[-2pt]
The model's node labels/elements and colors for both graphs are invented and do not match the provided Graph A and Graph B descriptions, which use the color roles mint, lemon, sky, rose, peach, etc., together with specific edge lists. Therefore, steps (1), (4), and (5) are unsupported.

The true Graph A is a 6-node path 0--1--2--3--4--5, with edges \texttt{(0,1)}, \texttt{(1,2)}, \texttt{(2,3)}, \texttt{(3,4)}, and \texttt{(4,5)}. The model's claimed Molecule A connectivity in steps (2)--(3) is different and incorrect.

The true Graph B has edges \texttt{(0,1)}, \texttt{(1,2)}, \texttt{(2,3)}, \texttt{(2,4)}, and \texttt{(2,5)}. The model instead asserted an edge \texttt{(5,4)} and a different branch structure in steps (6)--(7), which is incorrect.

Because the model's node attributes and topologies are incorrect, all subsequent mapping reasoning in steps (8)--(12) is unsupported. In particular, several claimed edge matches under the proposed mapping are inconsistent with the actual Graph B edge set. For example, the model claims that edge \texttt{(A5,A4)} maps to \texttt{(B5,B4)}, but edge \texttt{(5,4)} does not exist in the true Graph B.

Therefore, every listed reasoning step contains incorrect or unsupported claims relative to the provided graphs, and the final mapping/edit distance are not validated and are incorrect.\\[2pt]

\textbf{3) Final Judge Result:}\\[-2pt]
\texttt{Reasoning Steps: [0,0,0,0,0,0,0,0,0,0,0,0]; Final Answer: 0/1}

\end{caseboxpurple}
\end{figure*}

\begin{figure*}[t]
\caption{A graph edit distance prompt for molecular graph alignment under restored visualization layouts (Question).}
\label{fig:ged_prompt}
\begin{caseboxpurple}{\scriptsize Judge Feedback}
\fontsize{5.8}{6.4}\selectfont

\textbf{Question Prompt:}\\[-2pt]
You are comparing two molecular graphs to brief medicinal chemists on structural alignment.

To quantify how similar these molecules are, solve a Graph Edit Distance-style matching problem by proposing an atom-level mapping from Molecule A to Molecule B that minimizes the total number of edit operations (insertions, deletions, or substitutions), assuming unit cost per edit.

Each node is an atom labeled with its element, edges are bonds, and both molecules are rendered separately with their atom labels shown on the diagram.

Color legend (hypothetical pharmacophore roles): rose=H-bond donor site, peach=H-bond acceptor site, lemon=aromatic ring atom, mint=protonatable/positive center, sky=deprotonatable/negative center (assumed annotations).

Scenario: the visualization panels were tampered during capture. Molecule A: had a 3$\times$3 tile patch-swap applied. Molecule B: had a 3$\times$3 tile patch-swap applied. For this task, assume the affected panels have been fully restored to their correct layouts before solving for the optimal mapping.

Task: Provide the minimum-edit-cost atom mapping from Molecule A to Molecule B as a vector: the entry at position $i$ corresponds to atom $i$ in Molecule A and contains either the ID of its matched atom in Molecule B or \texttt{null} if it is unmatched.

If multiple optimal mappings with the same minimum edit cost exist, returning any one valid mapping is acceptable. For each atom $i$ in Molecule A, if it has no counterpart in Molecule B under the optimal edit sequence, output \texttt{null} at position $i$ in the mapping vector to indicate the corresponding insertion/deletion event.

First, explicitly state the numerical value of the minimum edit distance, then provide the atom mapping vector.

Answer format: \texttt{edit\_distance: <numerical\_value>; mapping: [b0\_or\_null, b1\_or\_null, b2\_or\_null, ...]}

Example: \texttt{edit\_distance: 5; mapping: [2, 1, 0, 4, 3, 5, 6]}

\end{caseboxpurple}
\end{figure*}

\begin{figure*}[t]
\centering
\caption{Prompt: Longest Cycle Detection-Airline Network}
\begin{casebox}{Prompt: Longest Cycle Detection-Airline Network}

\scriptsize
\textbf{Base prompt:}\\[2pt]
You are an airline network reliability engineer auditing the redundant loops that keep regional operations resilient during diversions.\\
Analyze the undirected flight network to find the single longest simple cycle (a closed loop with no repeated intermediate airports). If several loops tie for that maximum length, you may brief dispatch with any one representative loop.\\
Each node shows an airport's name or code, and each edge is an active nonstop corridor; distance annotations are intentionally omitted.\\
Color legend (operational priority tiers): rose=critical insight, peach=high-priority concept, lemon=standard fact, mint=supporting detail, sky=background reference.\\
List the airports in travel order and repeat the starting airport at the end to close the loop.

\medskip

\textbf{Patch transformation:}\\[2pt]
Scenario: a 3x3 tile patch-flip was accidentally applied while exporting the board. Reconstruct the original layout before searching for loops.\\
Task: After mentally restoring the shuffled tiles, identify the longest simple cycle and report one qualifying loop with its length.\\
Answer format: representative longest cycle: [Airport $\rightarrow$ ... $\rightarrow$ Airport]; length: $<$value$>$ airports\\
Example: representative longest cycle: [PDX $\rightarrow$ SFO $\rightarrow$ LAS $\rightarrow$ PDX]; length: 3 airports

\medskip

\textbf{Subgraph transformation:}\\[2pt]
Scenario: only the highlighted airports belong to the maintenance drill you are auditing. Treat other airports as off-limits for this query.\\
Task: Focus only on the airports painted lemon, then report one representative longest cycle and its length within that induced subgraph.\\
Answer format: representative longest cycle: [Airport $\rightarrow$ ... $\rightarrow$ Airport]; length: $<$value$>$ airports\\
Example: representative longest cycle: [BOS $\rightarrow$ JFK $\rightarrow$ DCA $\rightarrow$ BOS]; length: 3 airports

\medskip

\textbf{Split transformation:}\\[2pt]
Scenario: operations split the network into two panes. Before solving, mentally add the listed connectors (ATL --[route]-- MIA; DEN --[route]-- PHX) to reconstitute the live routes.\\
Task: Treat the provided connectors as active routes linking the two panels, then report a representative longest cycle and its length.\\
Answer format: representative longest cycle: [Airport $\rightarrow$ ... $\rightarrow$ Airport]; length: $<$value$>$ airports\\
Example: representative longest cycle: [MIA $\rightarrow$ ATL $\rightarrow$ CLT $\rightarrow$ MIA]; length: 3 airports

\medskip

\textbf{Color-swap transformation:}\\[2pt]
Scenario: operations retagged priority tiers. Current palette: rose, peach, mint. Recolor every airport below anchor Phoenix Hub to rose while keeping all other airports unchanged.\\
Task: After applying the recoloring instructions, provide one representative longest cycle (airport names only) and its length.\\
Answer format: representative longest cycle: [Airport $\rightarrow$ ... $\rightarrow$ Airport]; length: $<$value$>$ airports\\
Example: representative longest cycle: [DEN $\rightarrow$ ABQ $\rightarrow$ PHX $\rightarrow$ DEN]; length: 3 airports

\end{casebox}
\end{figure*}

\begin{figure*}[t]
\centering
\caption{Prompt: Longest Relationship Chain - DBPedia Knowledge Graph}
\begin{casebox}{Prompt: Longest Relationship Chain - DBPedia Knowledge Graph}

\scriptsize
\textbf{Base prompt:}\\[2pt]
You are reviewing a DBPedia knowledge graph to find the longest relationship chain in the graph.\\
The longest relationship chain on this knowledge graph can be found by solving a graph diameter problem, where the diameter equals the maximum number of edges among all shortest paths between every pair of nodes.\\
Each node is a concept entity and each edge shows the relationship label directly in the diagram.\\
Color legend (importance tier): rose=critical insight, peach=high-priority concept, lemon=standard fact, mint=supporting detail, sky=background reference.\\
If multiple longest shortest paths exist, returning any one of them is acceptable.

\medskip

\textbf{Patch transformation:}\\[2pt]
Scenario: the exported screenshot was tampered during transit. A 3×3 tile was patch-fliped, so mentally undo that corruption before tracing the longest relationship chain.\\
Task: Identify the longest relationship chain (graph diameter) across the entire knowledge graph and list the entities along that chain.\\
Answer format: diameter: $<$int$>$; path: [Entity1, Entity2, ..., EntityN]\\
Example: diameter: 4; path: [EntityA, EntityB, EntityC, EntityD]

\medskip

\textbf{Subgraph transformation:}\\[2pt]
Scenario: incident response only trusts concepts painted rose, so restrict reasoning to that induced subgraph while preserving their original connections and still report the longest relationship chain inside it.\\
Task: Within the trusted subgraph, identify the longest relationship chain and list the entities along that chain.\\
Answer format: diameter: $<$int$>$; path: [Entity1, Entity2, ..., EntityN]\\
Example: diameter: 3; path: [EntityX, EntityY, EntityZ, EntityW]

\medskip

\textbf{Split transformation:}\\[2pt]
Scenario: during a streaming knowledge update, analysts discovered that certain entity pairs from two separate graph panels formed newly recorded relationships (EntityA --[parent]-- EntityB; EntityC --[influenced]-- EntityD). Treat each listed connector as a fresh bridge between the panels, merge everything into one graph, and then report its longest relationship chain.\\
Task: After reconstructing the up-to-date knowledge graph, identify the longest relationship chain across all entities.\\
Answer format: diameter: $<$int$>$; path: [Entity1, Entity2, ..., EntityN]\\
Example: diameter: 4; path: [EntityA, EntityB, EntityC, EntityD]

\medskip

\textbf{Color-swap transformation:}\\[2pt]
Scenario: knowledge architects updated the importance tiers, and cardinal directions are determined by comparing the circle centers (a node is 'above' another if its center is higher). The visualization currently uses these colors: rose, peach, mint. Recolor every node above anchor Bratislava Region to mint while keeping other nodes unchanged. When returning the diameter, append each entity's color name after applying this recoloring in the format Entity(ColorName).\\
Task: After mentally applying the recoloring instructions, identify the longest relationship chain and list each entity together with its concept-tier color.\\
Answer format: diameter: $<$int$>$; path with color: [Entity1(ColorName), Entity2(ColorName), ...]\\
Example: diameter: 4; path with color: [EntityA(Rose), EntityB(Sky), EntityC(Mint), EntityD(Peach)]

\end{casebox}
\end{figure*}

\begin{figure*}[t]
\centering
\caption{Prompt: Shortest Relationship Chain in Knowledge Graph}
\begin{casebox}{Prompt: Shortest Relationship Chain in Knowledge Graph}
\scriptsize
\textbf{Base:}\\
You are reviewing a DBPedia knowledge graph to brief analysts on how entities connect through relationship chains.\\
To answer connection requests, compute the fewest-hop path (graph shortest path) between the specified source and target entities.\\
Each node is a DBPedia entity and each edge is labeled with the relationship shown directly on the diagram.\\
Color legend (importance tier): rose=critical insight, peach=high-priority concept, lemon=standard fact, mint=supporting detail, sky=background reference.\\
If multiple shortest paths tie for the minimum hops, returning any one of them is acceptable.

\medskip

\textbf{Patch:}\\
Scenario: the exported capture was tampered with, and a 3×3 tile was patch-fliped. Mentally restore the layout before tracing the shortest chain from Entity Alpha to Entity Omega.\\
Task: Determine the shortest relationship chain that connects Entity Alpha to Entity Omega and list every entity along that path.\\
Answer format: distance: <int>; path: [Entity1, Entity2, ..., EntityN] \\
Example: distance: 3; path: [EntityA, EntityB, EntityC, EntityD]

\medskip

\textbf{Subgraph:}\\
Scenario: incident review currently trusts only the entities painted lemon, so restrict reasoning to that induced subgraph before solving for the fewest-hop chain between Entity Alpha and Entity Omega.\\
Task: Within the trusted panel, list the shortest relation chain that links Entity Alpha to Entity Omega.\\
Answer format: distance: <int>; path: [Entity1, Entity2, ..., EntityN] \\
Example: distance: 4; path: [EntityA, EntityB, EntityC, EntityD, EntityE]

\medskip

\textbf{Split:}\\
Scenario: during a streaming knowledge update, engineering displayed the legacy graph and the incoming panel separately. Alpha --[ally]-- Beta; Gamma --[historicalLink]-- Omega. Use those connector rules to combine both panels before solving.\\
Task: After recombining the two panels, report the shortest relationship chain between Entity Alpha and Entity Omega.\\
Answer format: distance: <int>; path: [Entity1, Entity2, ..., EntityN] \\
Example: distance: 3; path: [EntityA, EntityB, EntityC, EntityD]

\medskip

\textbf{Color Swap:}\\
Scenario: knowledge architects retiered the relationship cues, and cardinal directions are determined by comparing circle centers (a node is 'above' another if its center is higher). The visualization currently uses these colors: rose, peach, mint. Recolor every entity below anchor Compass Node to rose while keeping the rest as displayed. When responding, format each entity as Entity(ColorName).\\
Task: After mentally applying the recoloring instructions, identify the shortest path between Entity Alpha and Entity Omega and append each entity's color name.\\
Answer format: distance: <int>; path with color: [Entity1(ColorName), Entity2(ColorName), ...] \\
Example: distance: 3; path with color: [EntityA(Rose), EntityB(Sky), EntityC(Mint), EntityD(Peach)]
\end{casebox}
\end{figure*}

\begin{figure*}[t]
\centering
\caption{Prompt: Maximum Independent Set for Social Influence Network}
\begin{casebox}{Prompt: Maximum Independent Set for Social Influence Network}
\scriptsize
\textbf{Base:}\\
You are reviewing a social influence network to coordinate non-overlapping outreach squads.\\
To avoid conflicting contacts, solve the Maximum Independent Set problem: find the largest set of users that are pairwise nonadjacent (no friendship edge between any two of them).\\
Each node is a community user and each edge is a friendship connection shown directly on the diagram.\\
Color legend (persona focus): rose=Community Programs, peach=Product Strategy, lemon=Growth Marketing, mint=Sales Enablement, sky=Customer Support.\\
If multiple maximum independent sets exist in the graph, returning any one of them is acceptable.

\medskip

\textbf{Patch:}\\
Scenario: the exported capture was tampered and a 3×3 tile underwent a patch-flip transformation. For this task, assume the layout has been fully restored to its correct arrangement before selecting the conflict-free delegate set.\\
Task: Identify the largest group of mutually nonadjacent users (the maximum independent set) and list them all.\\
Answer format: independent set: [User1, User2, ...] \\
Example: independent set: [UserA, UserB, UserC]

\medskip

\textbf{Subgraph:}\\
Scenario: incident review only trusts users painted rose. Restrict your reasoning to the subgraph induced by those rose nodes, keeping all friendship edges between them that exist in the original graph.\\
Task: Within this trusted subgraph, list the maximum independent set of approved users.\\
Answer format: independent set: [User1, User2, ...] \\
Example: independent set: [UserA, UserB, UserC]

\medskip

\textbf{Split:}\\
Scenario: community operations synced two newly collected panels separately, and the following cross-community friendships were just confirmed: UserA --[friend]-- UserB; UserC --[ally]-- UserD. Treat these as new friendship edges, combine the panels into one graph, and then extract the maximum independent set.\\
Task: After merging the two panels into a single graph, report the largest set of mutually nonadjacent users (the maximum independent set).\\
Answer format: independent set: [User1, User2, ...] \\
Example: independent set: [UserA, UserB, UserC, UserD]

\medskip

\textbf{Color Swap:}\\
Scenario: persona leads re-tiered the outreach focus, and cardinal directions are determined by comparing circle centers (a node is 'above' another if its center is higher, and 'left of' another if its center is further to the left). The visualization currently uses these colors: rose, peach, mint. Recolor every user above anchor Outreach Hub to mint while keeping others unchanged. Because colors encode persona focus (rose=Community Programs, peach=Product Strategy, lemon=Growth Marketing, mint=Sales Enablement, sky=Customer Support), report each user as User(ColorName), with ColorName capitalized (e.g., Rose, Peach, Mint).\\
Task: After applying the recoloring instructions, list the maximum independent set and append each user's persona color.\\
Answer format: independent set with color: [User1(ColorName), User2(ColorName), ...] \\
Example: independent set with color: [UserA(Rose), UserB(Mint), UserC(Sky)]
\end{casebox}
\end{figure*}

\begin{figure*}[t]
\centering
\caption{Prompt: Common Neighbors in Co-authorship Network}
\begin{casebox}{Prompt: Common Neighbors in Co-authorship Network}
\scriptsize
\textbf{Base:}\\
You are reviewing a DBLP co-authorship network to advise research leads on overlapping collaboration circles.\\
To identify shared collaborators for two leads, compute the set of common neighbors: authors who are directly connected to both leads by a co-authorship edge.\\
Each node is an author and each edge marks at least one prior joint publication between those authors.\\
Color legend (research focus): rose=NLP, peach=Computer Vision, lemon=Data Mining, mint=Information Retrieval, sky=Knowledge Graphs.\\
The set of common neighbors is unique, but any ordering of the shared collaborators in your answer is acceptable.

\medskip

\textbf{Patch:}\\
Scenario: the exported slide was corrupted in transit and a 3×3 tile underwent a patch-swap transformation. For this task, assume the slide has been fully restored to its correct layout before listing the authors shared between Author Alice and Author Bob.\\
Task: Identify every author who directly collaborates with both Author Alice and Author Bob, and list them all.\\
Answer format: common neighbors: [Author1, Author2, ...] \\
Example: common neighbors: [AuthorX, AuthorY, AuthorZ]

\medskip

\textbf{Subgraph:}\\
Scenario: compliance has only cleared authors painted mint. Restrict your reasoning to the subgraph induced by those mint nodes, and within that subgraph report all authors who are directly connected to both Author Alice and Author Bob.\\
Task: Within this trusted subgraph, identify the authors who are common neighbors of Author Alice and Author Bob.\\
Answer format: common neighbors: [Author1, Author2, ...] \\
Example: common neighbors: [AuthorX, AuthorY]

\medskip

\textbf{Split:}\\
Scenario: bibliometric analysts surfaced two newly collected collaboration panels, each covering different venues. The following cross-panel connections were specified: Alice --[coauthor]-- Carol; Bob --[coauthor]-- Dave. Treat these as new co-authorship edges, merge the panels into one graph, and then identify all shared collaborators of Author Alice and Author Bob.\\
Task: After recombining the two panels into a single network, list every author who is a direct collaborator of both Author Alice and Author Bob.\\
Answer format: common neighbors: [Author1, Author2, ...] \\
Example: common neighbors: [AuthorX, AuthorY, AuthorZ]

\medskip

\textbf{Color Swap:}\\
Scenario: the research office reclassified focus areas, and cardinal directions are determined by comparing circle centers (a node is 'above' another if its center is higher, and 'left of' another if its center is further to the left). The visualization currently uses these colors: rose, peach, mint. Recolor every author right of anchor Central Hub to peach while keeping the remainder as shown. Because colors map to research interests (rose=NLP, peach=Computer Vision, lemon=Data Mining, mint=Information Retrieval, sky=Knowledge Graphs), report each author as Author(ColorName), with ColorName capitalized (e.g., Rose, Peach, Mint).\\
Task: After applying the recoloring instructions, report every common neighbor of Author Alice and Author Bob, appending each author's updated research-focus color.\\
Answer format: common neighbors with color: [Author1(ColorName), Author2(ColorName), ...] \\
Example: common neighbors with color: [AuthorX(Rose), AuthorY(Mint)]
\end{casebox}
\end{figure*}

\begin{figure*}[t]
\centering
\caption{Prompt: Airline Route Optimization (TSP)}
\begin{casebox}{Prompt: Airline Route Optimization (TSP)}
\scriptsize
\textbf{Base:}\\
You are an airline route optimization specialist tasked with briefing dispatchers on the most efficient inspection tour.\\
Solve a Traveling Salesman Problem (TSP) on the provided flight-distance graph to find the minimum-length cycle that visits every airport exactly once and returns to the start.\\
Each node represents an airport (labeled by code/name) and each edge shows the great-circle distance in kilometers; the graph may include derived edges created from shortest-path distances.\\
Color legend (operational priority tiers): rose=critical insight, peach=high-priority concept, lemon=standard fact, mint=supporting detail, sky=background reference.\\
If multiple tours tie for optimal distance, returning any one of them is acceptable as long as you report the total distance in kilometers.

\medskip

\textbf{Patch:}\\
Scenario: the exported diagram had a 3x3 tile patch-flip applied during capture. Mentally restore the affected layout before solving the TSP.\\
Task: Report the optimal tour that visits each airport once and returns to the start, along with the total distance.\\
Answer format: optimal tour: [Airport1 -> Airport2 -> ... -> Airport1]; distance: <value> km\\
Example: optimal tour: [ATL -> DEN -> ORD -> ATL]; distance: 5280 km

\medskip

\textbf{Subgraph:}\\
Scenario: only airports painted rose are certified for this inspection sweep. Treat all other airports as temporarily unavailable and solve the TSP on the remaining network.\\
Task: Restrict attention to the highlighted airports only, then report the optimal tour and total distance within that induced subgraph.\\
Answer format: optimal tour: [Airport1 -> Airport2 -> ... -> Airport1]; distance: <value> km\\
Example: optimal tour: [SFO -> LAS -> PHX -> SFO]; distance: 3120 km

\medskip

\textbf{Split:}\\
Scenario: two regional panels were separated for layout purposes. Treat the listed connectors as active routes bridging the panels (JFK --[route]-- ORD; SFO --[route]-- LAS) before solving the TSP on the merged graph.\\
Task: After applying the newly provided connectors, report the optimal tour and its total distance.\\
Answer format: optimal tour: [Airport1 -> Airport2 -> ... -> Airport1]; distance: <value> km\\
Example: optimal tour: [BOS -> YYZ -> DCA -> BOS]; distance: 1890 km

\medskip

\textbf{Color Swap:}\\
Scenario: network planners retagged airport priorities. The diagram currently uses these colors: rose, peach, mint. Recolor every airport above anchor Denver Hub to mint while keeping all other airports unchanged.\\
Task: After mentally applying the recoloring instructions, report the optimal tour, total distance, and list the color for each airport along that tour.\\
Answer format: optimal tour with colors: [Airport1(Color), Airport2(Color), ... , Airport1(Color)]; distance: <value> km\\
Example: optimal tour with colors: [MIA(rose) -> IAH(mint) -> DEN(peach) -> MIA(rose)]; distance: 4210 km
\end{casebox}
\end{figure*}

\end{document}